\documentclass{article}

\usepackage{arxiv}

\usepackage[utf8]{inputenc} 
\usepackage[T1]{fontenc}    
\usepackage{hyperref}       
\usepackage{url}            
\usepackage{booktabs}       
\usepackage{amsfonts}       
\usepackage{nicefrac}       
\usepackage{microtype}      
\usepackage{lipsum}
\usepackage{graphicx}
\graphicspath{ {./images/} }
\usepackage{relsize}
\usepackage{amsmath} 
\usepackage{float}
\usepackage{xcolor}
\usepackage{soul}
\usepackage[most]{tcolorbox}
\usepackage{flushend}
\usepackage{framed}

\title{Enhancing Medical Task Performance in GPT-4V: A Comprehensive Study on Prompt Engineering Strategies}

\author{
 Pengcheng Chen \\
  Shanghai AI Laboratory\\
   \texttt{chenpengcheng@pjlab.org.cn} \\
   \And
 Ziyan Huang\\
  Shanghai AI Laboratory\\
     \texttt{huangziyan@pjlab.org.cn} \\
  \And  
 Zhongying Deng \\
  Shanghai AI Laboratory\\
     \texttt{dengzhongying@pjlab.org.cn} \\
  \And
 Tianbin Li \\
  Shanghai AI Laboratory\\
     \texttt{litianbin@pjlab.org.cn} \\
  \And
 Yanzhou Su \\
  Shanghai AI Laboratory\\
     \texttt{suyanzhou@pjlab.org.cn} \\
  \And  
 Haoyu Wang \\
  Shanghai AI Laboratory\\
     \texttt{wanghaoyu@pjlab.org.cn} \\
  \And
 Jin Ye \\
  Shanghai AI Laboratory\\
     \texttt{yejin@pjlab.org.cn} \\
  \And 
 Yu Qiao \\
  Shanghai AI Laboratory\\
     \texttt{qiaoyu@pjlab.org.cn} \\
  \And 
 Junjun He \\
  Shanghai AI Laboratory\\
     \texttt{hejunjun@pjlab.org.cn} \\
  \And 
}

\begin{document}
\maketitle
\begin{abstract}
OpenAI's latest large vision-language model (LVLM), GPT-4V(ision), has piqued considerable interest for its potential in medical applications. Despite its promise, recent studies and internal reviews highlight its underperformance in specialized medical tasks. This paper explores the boundary of GPT-4V's capabilities in medicine, particularly in processing complex imaging data from endoscopies, CT scans, and MRIs etc. Leveraging open-source datasets, we assessed its foundational competencies, identifying substantial areas for enhancement. Our research emphasizes prompt engineering, an often-underutilized strategy for improving AI responsiveness. Through iterative testing, we refined the model's prompts, significantly improving its interpretative accuracy and relevance in medical imaging. From our comprehensive evaluations, we distilled 10 effective prompt engineering techniques, each fortifying GPT-4V's medical acumen. These methodical enhancements facilitate more reliable, precise, and clinically valuable insights from GPT-4V, advancing its operability in critical healthcare environments. Our findings are pivotal for those employing AI in medicine, providing clear, actionable guidance on harnessing GPT-4V's full diagnostic potential.
\end{abstract}



\section{Introduction}

The burgeoning field of multimodal medical large language models offers promising applications for the future of healthcare and artificial intelligence (AI) research. One of the leading AI models in this domain is OpenAI's GPT-4\cite{openai2023gpt4}, which has recently been enhanced with a multimodal variant, GPT-4V\cite{yang2023dawn}, incorporating image input capabilities. This model extension draws from the formidable linguistic foundation of GPT-4 and has garnered significant attention due to its potential in processing and interpreting medical imagery.

Prior evaluations, such as those conducted by Chaoyi Wu et al, have provided a detailed examination of GPT-4V's medical performance, probing the limits of its image processing abilities\cite{wu2023gpt4vision}, also the performance of GPT-4V in medical examination has been tested\cite{yang2023performance,buckley2023accuracy,han2023comparative} and reveals the potential of the usage of GPT-4V in the medical area. In the realm of MLLMs, performance is contingent on both image comprehension and textual understanding capabilities. Merely inputting an image along with basic, unrefined prompts does not fully utilize the model's language understanding potential and only partially engages its image interpretation capabilities. To delineate the boundaries of GPT-4V's abilities, research focused on prompt optimization is crucial. Building upon previous studies, this work concentrates on exploring the optimal combination of prompts to enable GPT-4V to operate at its peak efficiency, thereby offering novel insights for subsequent medical research utilizing GPT-4V.

In our investigation, we selected diverse modalities of data and experimented with various combinations of textual and visual prompts. Ultimately, we synthesized ten tips that significantly enhance GPT-4V's performance in medical applications. Given that the focus of this study is on prompt engineering rather than an exhaustive test of medical capabilities, we did not examine an extensive array of cases. Instead, this paper will showcase results from a selected set of tests, emphasizing the impact of strategic prompt design on enhancing the model's effectiveness in medical image analysis and interpretation.

\section{Methods}







\subsection{Overview}
In this study, we utilize an image and text prompt as the input, followed by a manual assessment of the output's quality. Our methodology involves presenting the same image with a variety of textual prompts. Each prompt's output is then subject to a thorough human evaluation. When a particular textual prompt yields exceptionally positive results, it undergoes more in-depth testing and further evaluation in our test sets. If this prompt strategy consistently demonstrates superior performance across all tests, it is then incorporated into our list of effective prompt tips.

To optimize the testing process and ensure rigorous and systematic evaluation, we have established the following 8 rules during the test part:

\begin{itemize}

    \item \textbf{Rule 1 - The latest data policy:} Considering that GPT-4's training data does not extend beyond the year 2021, our selected dataset predominantly consists of data from 2022 onwards or from our own collected data. A smaller portion of the dataset may originate from 2021 and prior, but we intentionally do not employ this data for diagnostic tasks to prevent testing on the training set. Regarding data that may have been trained on by GPT-4 and its variants, we use it solely for tasks such as pairing, semantic recognition, and others that have no direct correlation with GPT-4's pre-training activities.

    \item \textbf{Rule 2 - Challenging High-Difficulty Projects for Superiority Proofs:} In order to demonstrate the superiority of our prompt-crafting techniques, we actively undertake challenging projects. Some of these tasks may lack immediate clinical utility; however, they serve as significant tests to evaluate the model's capabilities in handling medical images.
        
    \item \textbf{Rule 3 - Ablation Studies to Avoid Answer Leakage:} For certain scenarios, we employ ablation studies to eliminate the possibility of answer leakage. Previous research on GPT-4V has often overlooked the restriction of prompt content, leading to scenarios where correct answers could be deduced from the text alone\cite{buckley2023accuracy}, making it unclear whether the results were derived from image recognition capabilities or textual understanding. In such cases, ablation studies are essential to distinguish between these two possibilities, thereby ensuring a more accurate assessment of the model's performance.
        
    \item \textbf{Rule 4 - Focus on Conclusion Accuracy Over Process Correctness:} During our experiments, we do not delve into the correctness of each output detail. All tests generate extensive content, sometimes including correct conclusions based on partially incorrect reasoning, or vice versa. Our primary concern is the accuracy of the conclusions drawn, rather than the correctness of each step in the reasoning process.
        
    \item \textbf{Rule 5 - Ignoring Disclaimers in Medical Contexts:} Given that GPT-4V is not a modality model specifically trained for medical purposes, developers typically include disclaimers in medical-related content to prevent its use in diagnostic contexts. Since these disclaimers often occupy significant space and are not central to our research focus, we choose to disregard them in our analysis. This allows us to concentrate on the content's relevance and accuracy without the distraction of repetitive legal cautioning.
        
    \item \textbf{Rule 6 - Trialing Outputs Thrice for Stability Analysis:} In previous studies, the stability of GPT-4V's responses, particularly in natural image contexts, was not scrutinized. However, in medical imagery, we observed that GPT-4V sometimes produces different answers for the same prompt\cite{gptv2023systemcard}. Therefore, analyzing the stability of these responses is a critical aspect of our study. For lengthy outputs like video process analyses, we report the weakest result from multiple trials to highlight areas needing improvement.
        
    \item \textbf{Rule 7 - Single-Round Dialogue Tests Unless Otherwise Specified:} Unless explicitly stated as multi-round dialogue tests, all experiments are conducted as single-round dialogues. After each dialogue, the session is immediately terminated and a new session is started for the subsequent test. This approach ensures that each test is evaluated independently, allowing for a more accurate assessment of the model's performance in isolated interactions.
        
    \item \textbf{Rule 8 - Version Specification of GPT-4V:} The version of GPT4-V used in our research spans from September 27 to October 18, 2023. It is important to note that subsequent updates to the model post this period may impact the effectiveness of certain prompt tips, potentially rendering some obsolete. This time frame is critical for contextualizing our findings and understanding the specific capabilities of the model version under study.
\end{itemize}

In the manual evaluation part, we compared the outputs of GPT-4V against a set of reference answers. This process involved a scoring system wherein a match with the reference answer earned one point. Responses that were not incorrect but also not completely aligned with the reference answer neither gained nor lost points. Incorrect answers resulted in a deduction of one point. Through this scoring mechanism, we calculated the scores for each distinct prompt strategy employed during the study. The 10 prompt tips with the highest scores were then extracted. These top-scoring prompt tips form the crux of our study's conclusions, offering valuable insights into the most effective ways of interacting with the GPT-4V model. This approach not only underscores the model's capabilities but also highlights the nuances of optimizing AI interactions in complex tasks.

\subsection{Testing procedure}
In the specific context of prompt testing, the experimental process of our study typically divides into two or three distinct parts:

\begin{itemize}
    \item \textbf{Control Group:}  Under the guidelines mentioned in Section 2.1, within this group, we generally employ the simplest and most direct prompts, or sometimes only input the image without any textual prompt. Upon obtaining the corresponding outputs, these are then compared with the reference answers for scoring.
    \item  \textbf{Experiment Group:} Also following the rules outlined in Section 2.1, this group uses the same images as the control group for input. However, the textual prompts are optimized for this set. The outputs generated from these optimized prompts are then compared with the reference answers and scored accordingly.
    \item  \textbf{Ablation Study (if necessary):} In accordance with the rules specified in Section 2.1, this part of the experiment involves inputting only textual prompts without any images. The outputs are then compared with the reference answers for evaluation and scoring.
\end{itemize} 

\begin{figure}[H]
  \centering
  \includegraphics[width=1\linewidth]{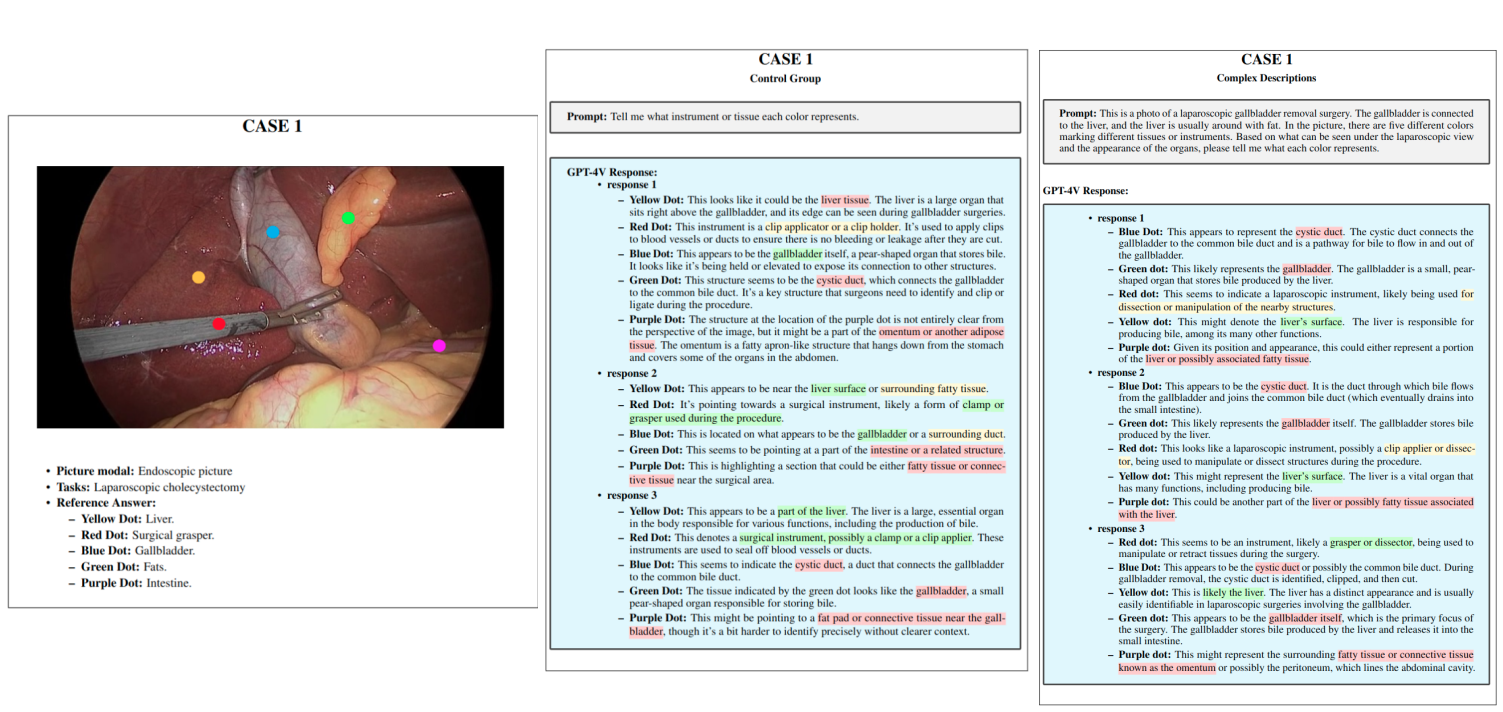}
  \label{showcase}
  \caption{A typical example of demonstration, illustrating the basic information of a case along with the results from both control and experimental groups}
\end{figure}

A typical demonstration template is shown in Figure\ref{showcase}. This structured approach enables a comprehensive assessment of the model's performance under different conditions, allowing for a clear understanding of the impact of various prompt strategies, Among all the trials, we will list out 10 tips for effective prompts.

\subsection{Testing dataset}
In our study, the endoscopy group in the test set utilized data from the \textit{Kvasir\_SEG} \cite{jha2019kvasirseg}and \textit{M2caiseg} datasets\cite{maqbool2020m2caiseg}, alongside data from the hysteroscopy procedures we collected. For CT imaging, the \textit{AutoPET}\cite{zhong2022autopet}, \textit{TotalSegmentator}\cite{Wasserthal_2023}, \textit{AbdomenCT-1K}\cite{ma2021abdomenct1k} datasets were used, supplemented with data from relevant literature. MRI data incorporated the \textit{BraTS 2021}\cite{baid2021rsnaasnrmiccai},\textit{ATLAS V2.0} and \textit{AMOS 2021}\cite{ji2022amos} datasets. It is crucial to acknowledge that this dataset compilation is not comprehensive and does not represent all medical scenarios. Nonetheless, it provides an adequate basis for testing our prompt tips. Future endeavors will aim to expand this dataset to further enrich the study.

\section{Prompt Tips}
Utilizing the aforementioned testing methodology and adhering to the established rules, we have successfully identified ten valuable prompt tips:
\begin{enumerate}
  \item Concise language is more effective than complex descriptions, emphasizing succinct, task-relevant details in image analysis.
  \item Providing the tasks can better help analyze the images
  \item Implementing step-by-step guidance in multi-round dialogue allows GPT-4V to handle complex tasks more efficiently by breaking them down into simpler operations.
  \item Don't expose your target at the very beginning when you start the multi-round dialog
  \item Describing appearances or characteristics will greatly enhance the performance.
  \item Appearance descriptors should be non-conflicting with any image annotations to avoid misinterpretations.
  \item Clarifying the contextual relationships between sequential images enhances GPT-4V's analytical accuracy and detail recognition.
  \item Splicing multiple images into one and providing a sequence order enhances task processing effectiveness compared to multiple simultaneous inputs.
  \item Providing comparative analysis opportunities, especially with temporal patient data, deepens lesion or condition assessments.
  \item Directing GPT-4V's focus to the interested areas will facilitate more targeted and relevant output.
\end{enumerate}




\section{Discussion}
\subsection{Tip 1: Concise language is more effective than complex descriptions, emphasizing succinct, task-relevant details in image analysis.}

\subsubsection{Demonstration Info}
In this demonstration, we conducted a comparative analysis of prompt outcomes using concise language versus complex descriptions within endoscopic, CT, and MRI imaging contexts, all under the condition of providing equivalent information. The study meticulously compared the effectiveness of simplified and elaborate prompt styles on GPT-4V's performance. The final results clearly indicated that, when the same amount of information is provided, prompts articulated in a concise manner significantly enhance GPT-4V's functional capabilities. This finding underscores the importance of prompt language efficiency in optimizing the performance of AI models in medical imaging scenarios, suggesting that brevity and clarity in prompts can lead to more effective AI analysis and interpretation within various medical imaging modalities.

\subsubsection{Analysis}
Based on the tests conducted, we observed that within endoscopic scenarios, simplified descriptions yielded better outcomes compared to complex descriptions. In contrast, within CT and MRI contexts, the performance of simplified descriptions was comparable to that of complex descriptions, both showing significant improvements over control groups with no or weak descriptions.

In this regard, we posit that the superior performance of simplified descriptions can be attributed to their provision of only the most critical information. Complex descriptions, conversely, offer an abundance of auxiliary information. During the generation of textual content, the weight of these auxiliary descriptive elements is also accounted for, which consequently diminishes the relative importance of core information. This can lead to slight deviations in results, potentially resulting in inaccuracies. Therefore, when inputting prompts into GPT-4V, it is advisable to include only the most relevant core information pertaining to the query at hand, minimizing the incorporation of extensive, indirectly related descriptive language.

\subsection{Tip 2: Providing the tasks can better help analyze the images}

\subsubsection{Demonstration Info}
In this demonstration, we illustrate how the task of providing medical images enhances the model's ability to analyze input medical images. We focus particularly on the identification of surgical instruments in single endoscopic images and the analysis of endoscopic sequences. An ablation study was incorporated into this demonstration to ensure that the textual prompts provided did not leak the answers. This approach underscores the effectiveness of combining visual and textual inputs in improving the model's performance in medical image analysis, while also ensuring the integrity of the results through rigorous testing methods.

\subsubsection{Analysis}
In this test, we found that specifying the type of task significantly enhances the recognition capabilities of GPT-4V. We believe that medical images, particularly endoscopic images, are challenging to differentiate based on a single image due to the limitations of the endoscope's field of view (FOV) and the similarity of most endoscopic scenes. This makes it difficult to identify the specific task of the endoscope, thereby complicating the analysis of the problem. After specifying the particular task of the endoscope, the probability distribution of the output becomes more focused on the answer, thereby increasing the accuracy of the response.

To further validate the effectiveness of our prompt strategy, we conducted an ablation study. In the ablation study, we did not input images and merely inputted prompts telling GPT-4V to pretend there was an image, then asked GPT-4V the same questions. If GPT-4V could still provide the correct answers under these circumstances, it would indicate that we had leaked answer-related information in our prompts. If it could not respond correctly, it would suggest that we did not disclose any answer-related information in our prompts.

We discovered that with pure text input, GPT-4V could not correctly answer our questions or provide analysis, indicating that our prompts did not leak additional information. This means that the correct answers were due to our prompt guiding the image analysis rather than the prompt leaking key answer-related information.

In contrast, we presented Case 4, a prompt we believe leaked some answer-related information. Compared to Case 3, we included a key piece of information, "ring-shape," which could lead GPT-4V to produce the correct answer without an image.

In summary, we believe that the conclusion that GPT-4V's performance can be further enhanced by providing the task is solid and valid.

\subsection{Tip 3: Implementing step-by-step guidance in multi-round dialogue allows GPT-4v to handle complex tasks more efficiently by breaking them down into simpler operations.}

\subsubsection{Demonstration Info}
In this demonstration, we showcased how, through step-by-step prompt tips, efficient matching of medical images with their corresponding masks can be achieved. It is important to emphasize that while this task may not hold substantial clinical significance, we selected it for demonstration due to its high level of difficulty. This task demands the utmost utilization of the multimodal model's image understanding capabilities, thereby exemplifying the value of prompt tips.

We also presented comparisons across three different modalities: endoscopy, CT, and MRI. These comparisons were based on scenarios where prompt tips were not used, where step-by-step guidance was provided in a single-round dialogue, and where step-by-step guidance was facilitated through multi-round dialogues. Our findings indicate that guiding GPT through multi-round dialogues with step-by-step instructions yielded the most effective results. This highlights the potential of structured, iterative interactions in harnessing the full capabilities of advanced AI models in complex image analysis tasks.

\subsubsection{Analysis}
In our study, the tests conducted on endoscopic imagery revealed certain limitations in GPT-4V's processing capabilities. Specifically, when tasked with identifying which mask correlates with a polyp in a colon image, GPT-4V consistently errs towards selecting the second mask. This error appears to stem from its misinterpretation of the vertical alignment of the real colonoscopy images and their corresponding masks, leading to a mistaken assumption of correlation where none exists.

Nevertheless, we observed that a gradual, step-by-step guidance approach significantly enhances GPT-4V's performance in handling complex tasks. However, this enhancement is not as pronounced when all step-by-step instructions are consolidated into a single-round dialogue. We believe this finding aligns with the conclusion drawn in Tip 1, which emphasizes that prompts should not be excessively lengthy or complex. In the process of guiding GPT-4V, the inclusion of an abundance of descriptive and medically irrelevant terminology adversely affected its performance, likely due to the undue weight these words carried in the model's processing.

Conversely, when we divided the step-by-step guidance across multiple rounds of dialogue, we noticed a marked improvement in GPT-4V's performance. This approach enabled it to accomplish challenging tasks, like accurate mask matching, with a high degree of precision—a feat not yet achievable by other large multimodal models. Such a prompting strategy effectively leverages GPT-4V's powerful capabilities.

Similarly, we observed a comparable effect in both the CT and MRI groups. Without any specifically designed prompts, GPT-4V failed to correctly identify the tasks at hand. However, with the incorporation of step-by-step guidance, there was a significant enhancement in performance, with multi-round dialogues demonstrating even more pronounced improvements. This evidence suggests that a structured, interactive approach is crucial for maximizing the potential of advanced multimodal AI models in complex medical imaging tasks.

\subsection{Tip 4: Don't expose your target at the very beginning when you start multi-round dialog}

\subsubsection{Demonstration Info}
In the previous demonstration, we showcased how step-by-step guidance within a multi-turn dialogue can effectively accomplish challenging tasks. In this demonstration, we aim to present a strategy that may potentially weaken the effectiveness of prompts: prematurely revealing the objective. We will use the same three cases and almost identical prompts as in the previous demonstration. The key difference, however, lies in our approach: we will disclose our objective -- matching masks with medical images to GPT-4V right from the start.

This setup allows us to critically evaluate the impact of early goal exposure on the performance of GPT-4V in completing the task. By comparing the outcomes of this approach with those from the previous demonstration, where the objective was gradually introduced, we aim to discern the influence of prompt structure and information sequencing on the model's efficiency and accuracy in task execution. This investigation is pivotal in understanding the nuances of effective prompt design, particularly in complex problem-solving scenarios involving AI models.

\subsubsection{Analysis}
In this demonstration, a notable difference from the previous one was observed when the objective was prematurely revealed. Despite employing our previously high-accuracy prompt strategies, the accuracy remained markedly low, comparable to the performance achieved with unrefined prompts.

Our observation led to the discovery that this decrease in performance is attributable to the "short-term memory" of GPT-4V. To elaborate, if the objective is revealed too early, GPT-4V tends to immediately generate an answer based on the exposed information. The quality of this answer is similar to that produced by unrefined prompts, which is to say, it is of lower quality. Further, if prompts are inputted after receiving this initial answer, in an attempt to guide GPT-4V, the subsequent outputs tend to be erroneous, aligning with the previously derived incorrect answer. The strong short-term memory of GPT-4V means that if the initial output is not manually identified as incorrect, it carries this erroneous prior into subsequent responses, thus affecting the quality of the answers. Therefore, when using GPT-4V with the intention of achieving a specific outcome through multi-turn dialogue, it is advisable not to disclose the task objective at the onset.

\subsection{Tip 5: Describing appearances or characteristics will greatly enhance the performance.}

\subsubsection{Demonstration Info}
In this demonstration, we illustrated that providing descriptions of appearance or characteristics significantly enhances the model's performance. For the descriptions of appearance or characteristics, we adopted definitions from Wikipedia in this showcase. Given that this prompt tip involves detailed descriptions of appearance and characteristics, it is most effectively applied in scenarios where there is a need for batch processing of data that is familiar to the user. 

\subsubsection{Analysis}
In this study, we discovered that providing specific descriptions of shapes and appearances significantly enhances GPT-4V's ability to recognize certain structures. In the control group, where no descriptions were provided, the performance was suboptimal. Specifically, within the endoscopy group, organs like the liver, fat, and intestines were difficult to identify without prompts, with only the gallbladder being relatively easier to recognize. In CT and MRI imaging, GPT-4V was able to identify easily recognizable organs like the liver without additional prompts, whereas identifying the gallbladder and pancreas proved to be more challenging.

Subsequently, we utilized organ-specific prompts based on appearance descriptions from Wikipedia and Mayo Clinic, among others. It is crucial to note that GPT-4V may output descriptions of the images in the order of the input prompts. If the order of descriptions inputted matches the order of annotations in the image, this might result in correct outputs. However, these outputs could be derived from the order of the input descriptions rather than an understanding of the image. To mitigate this, we intentionally scrambled the order of our descriptions to prevent answer leakage due to input sequencing. We observed a substantial increase in recognition accuracy upon providing appearance descriptions. In the endoscopy group, apart from the intestines which remained challenging, other structures were identified with high accuracy. In both the CT and MRI groups, complete and accurate identification was achieved, including small structures like the gallbladder in MRI images.

Furthermore, we also observed that when dealing with markers of different colors, it is advisable to avoid using color-related descriptions. Such descriptions can lead to confusion in GPT-4V's processing. We will delve into this phenomenon in greater detail in the following chapter, examining the implications of color descriptions on the model's interpretive accuracy and how they might inadvertently mislead the AI's analysis.

\subsection{Tip 6: Appearance descriptors should be non-conflicting with any image annotations to avoid misinterpretations.}

\subsubsection{Demonstration Info}
In the previous chapter, we discussed the impact of appearance descriptions on the performance of GPT-4V, noting a significant enhancement in its capabilities. Given that appearance descriptions often involve color, especially in endoscopic scenarios, it is pertinent to explore the outcomes when markers are also denoted by colors. In this round of demonstrations, we excluded data from CT and MRI modalities, as these primarily involve grayscale imaging and seldom incorporate colors beyond black and white. This focus allows us to specifically assess the influence of color descriptions in contexts where color plays a critical role, providing insights into the nuances of prompt design for color-rich medical imaging scenarios

\subsubsection{Analysis}
In this chapter, we discovered that using color descriptors in scenarios where markers are also color-coded can significantly impair performance. In this showcase, the results were even worse than those in the control group (Section 4.5.1). Notably, all instances of yellow dots were incorrectly identified as fat. This misidentification likely stems from GPT-4V interpreting the yellow dots as part of the scene rather than as markers. When we described fat as "generally pale yellow," GPT-4V mistakenly classified the yellow dots as fat. This observation underscores the importance of careful consideration in using color descriptors, particularly in contexts where color plays a dual role in both the scene and the markers.

\subsection{Tip 7: Clarifying the contextual relationships between sequential images enhances GPT-4v's analytical accuracy and detail recognition.}

\subsubsection{Demonstration Info}
This demonstration primarily focused on the processing of endoscopic sequences. We composed a single image by combining four pictures from a hysteroscopy to assess GPT-4V's capability in handling endoscopic sequences. The test was divided into two groups: one without inputting the image sequence numbers\ref{C7_CASE1} and another with the sequence numbers included \ref{C7_C8_CASE1}. Our findings indicate that image analysis is more effective when the sequence order of the images is provided. This result highlights the importance of sequential context in enhancing the model's comprehension and accuracy in processing related images in a series.

\subsubsection{Analysis}
Our findings in this chapter reveal that providing the sequential order of images significantly enhances their analysis. This approach is beneficial for processing video files as well. The underlying effectiveness of this strategy appears to stem from how GPT-4V perceives each image. Without sequence information, GPT-4V tends to treat each image as a discrete entity, leading to independent analysis with limited contextual understanding.

However, when the sequential order is specified, or when the interrelationship among the images is clarified, GPT-4V is more inclined to view the images as interconnected elements of a larger narrative. Consequently, it synthesizes a more comprehensive workflow, encompassing the entire sequence from the first to the last image. This methodology proves especially useful in the analysis of surgical procedures in videos or image sequences, allowing for a more integrated and coherent interpretation of the visual data.

\subsection{Tip 8: Splicing multiple images into one and providing a sequence order enhances task processing effectiveness compared to multiple simultaneous inputs.}

\subsubsection{Demonstration Info}
In the previous demonstration, we presented the results of splicing multiple images into one(\ref{C7_CASE1} and \ref{C7_C8_CASE1}). GPT-4V, however, possesses a built-in upload feature that can be used for analyzing image sequences. In this demonstration, we will compare the outcomes of using GPT-4V's native upload feature versus splicing multiple images together. It is important to note that, as of the September 27 to December 12, 2023 version, this upload system can handle a maximum of four images at a time. Additionally, images uploaded through this system are automatically displayed in varying sizes on the webpage, although their actual upload dimensions remain unchanged, as shown in Figure \ref{C8_CASE1_systematic_upload}. This comparison aims to highlight the differences in results obtained through these two methods of image presentation and processing.

\subsubsection{Analysis}
In our prior showcase, we created a composite image by splicing together multiple individual images. However, in this subsequent test, we employed GPT4-v's inherent image uploading feature to upload four images simultaneously, which is the system's maximum capacity, to analyze a surgical procedure depicted across these images. The results of this test indicate that the direct upload of multiple images through GPT4-v's system tends to be less effective in the analysis compared to utilizing a single, spliced composite image. This reduced effectiveness might be due to the current version of GPT4-v not being fully optimized for multi-image analysis. While GPT4-v shows remarkable proficiency in processing single images, it seems there is a need for additional development to enhance its capability to handle and interpreting multiple images concurrently.

\subsection{Tip 9: Providing comparative analysis opportunities, especially with temporal patient data, deepens lesion or condition assessments.}
\subsubsection{Demonstration Info}
In this demonstration, we highlighted the significant enhancement in GPT-4V's performance facilitated by the provision of comparative information. This comparison could involve contrasting healthy and abnormal states, or it might pertain to changes in pathological lesions. Such juxtapositions provide GPT-4V with a richer context, enabling a more accurate and nuanced understanding of the medical imagery. By presenting contrasting scenarios or changes over time, GPT-4V can better discern subtle differences, leading to improved diagnostic accuracy. This approach underscores the importance of contextual information in medical image analysis, particularly in complex cases where changes are subtle or involve gradual progression.

\subsubsection{Analysis}
In this test, we observed that providing GPT-4v with comparative images from before and after significantly improved the accuracy of diagnoses. This is intuitively reasonable since the nature of many conditions, such as the growth of pulmonary nodules, can be challenging to assess from a single, final-stage image. However, when presented with a series of images depicting the nodule over several years, GPT-4v is able to analyze the growth pattern of the nodule. Consequently, this enables a more accurate assessment of the patient's condition.

Additionally, in diagnosing lung nodules, we opted not to mark incorrect responses, focusing instead on correctly answered parts. This decision was based on the complexity inherent in diagnosing diseases solely from the growth patterns of nodules. We acknowledge that all responses provided by the model possess a certain degree of validity under these circumstances. However, it was observed that when images showing comparative growth were presented, all responses accurately identified adenocarcinoma as the correct diagnosis in the first answer. This finding suggests that providing contextual visual information significantly enhances the model's diagnostic accuracy, particularly in complex cases like lung nodule analysis.

\subsection{Tip 10: Directing GPT-4V's focus to the interested areas will facilitate more targeted and relevant output.}
\subsubsection{Demonstration Info}
In this demonstration, we examined the impact of providing spatial orientation information on GPT-4V's recognition capabilities. We observed that when directing GPT-4V's focus to the interested areas, GPT-4V's responses were notably more coherent and accurate. This finding suggests that spatial information significantly aids the AI in contextualizing and interpreting medical images. This prompt tip is particularly useful in cases where the approximate location of a lesion is known, but a detailed analysis and diagnosis are required using GPT-4V. By integrating location-based cues into the prompts, GPT-4V can focus its analysis more effectively, leading to more precise diagnostic outcomes. This approach highlights the importance of spatial context in enhancing AI-assisted medical image diagnosis.

\subsubsection{Analysis}
In the experiments conducted in this chapter, we found that providing specific spatial orientation significantly enhances the accuracy of GPT-4V's responses. In the MRI experiments, without specified location guidance, GPT-4V tended to diagnose images as normal, even when there were clear abnormalities present. Once directed to focus on the "upper left quadrant," GPT-4V promptly identified the anomalies in that region.

In the PET scan tests, without a defined area of observation, GPT-4V erroneously identified regions with naturally high metabolic activity, such as the brain, heart, and bladder, as potential pathological sites, leading to an incorrect diagnosis of possible metastatic cancer. However, when GPT-4V was instructed to concentrate on the thoracic region, it successfully differentiated between cardiac activity and lung cancer, providing a more accurate diagnosis. These findings highlight the importance of directional cues in assisting GPT-4V to accurately interpret medical imaging, emphasizing the need for precise location information in AI-assisted diagnostic processes.

\section{Summary}

In this work, we identified ten valuable prompt tips that significantly enhance GPT-4V's utility in medical contexts through extensive testing across various datasets. These prompts led to marked performance improvements in all tasks, demonstrating the power of well-designed prompts in enhancing AI analysis in medical imaging. Future research will broaden the case spectrum and employ these prompt techniques for bulk data processing using APIs. Additionally, we plan to explore the effectiveness of these prompts in non-medical, natural image settings. This exploration aims to understand the adaptability of these strategies across different image analysis scenarios. Our findings not only contribute to medical image analysis using AI models but also open new avenues for applying AI-driven strategies in a wider range of imaging applications.

\bibliographystyle{unsrt} 
\bibliography{references}

\newpage

\section{Case demonstration}
In our demonstration, each scenario is illustrated through comprehensive visuals. "Prompt" refers to specific textual inputs presented to the users. The "GPT-4V" section captures the responses elicited by the GPT-4V model. It is important to note that, due to predetermined safeguard measures, GPT-4V consistently asserts its non-expertise in specialized domains such as radiology. For clarity and succinctness, we have omitted these routine disclaimers from our presentation of results. "Reference answer" represents the standard responses or interpretations, as informed by expert sources like Radiopaedia.

For a clearer interpretation of our findings, we implemented a color-coding strategy: 
\begin{itemize}
    \item \textbf{Red:} \sethlcolor{red!20}\hl{Red} identifies errors in GPT-4V’s outputs, with corresponding information in the standard responses indicating the reason for these classifications.
    \item \textbf{Green:}  \sethlcolor{green!20}\hl{Green} emphasizes accurate responses, with identical segments in the expert responses confirming their validity.
    \item  \textbf{Yellow}  \sethlcolor{yellow!20}\hl{Yellow} is designated for portions that are ambiguous or unclear.
\end{itemize}
\subsection{Tip 1: Concise language is more effective than complex descriptions, emphasizing succinct, task-relevant details in image analysis.}
\subsubsection{Case 1: Endoscopic image annotation recognition}

\begin{center}
\fbox{%
\begin{minipage}{0.9\textwidth}
\begin{center}
    {\relsize{2}\textbf{CASE 1}}
\end{center}
\centering
\vspace{1em}
\begin{figure}[H]
  \centering
  \includegraphics[width=0.9\linewidth]{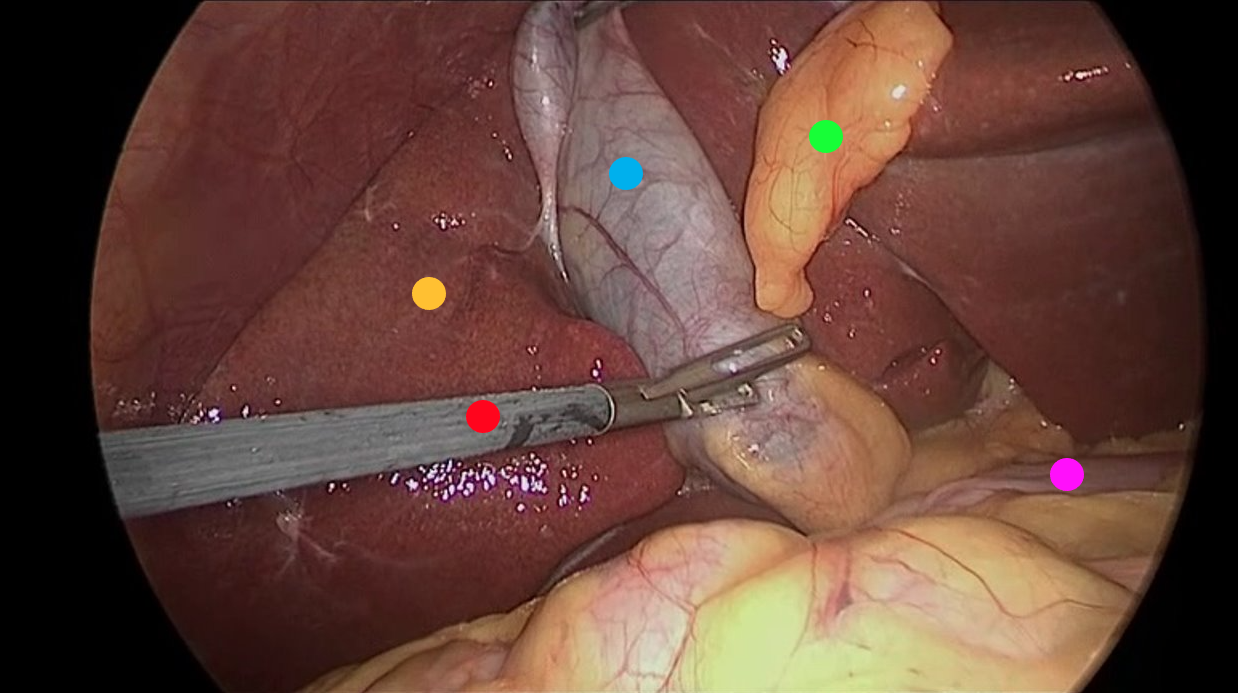}
  \label{CASE1}
\end{figure}

\vspace{1em}
\begin{flushleft}
\begin{itemize}

     \item \textbf{Picture modal:} Endoscopic picture
     \item \textbf{Tasks:} Laparoscopic cholecystectomy
     \item \textbf{Reference Answer:} 
            \begin{itemize}
                \item \textbf{Yellow Dot:} Liver.
                \item \textbf{Red Dot:} Surgical grasper.
                \item \textbf{Blue Dot:} Gallbladder.
                \item \textbf{Green Dot:} Fats.
                \item \textbf{Purple Dot:} Intestine.
            \end{itemize}

\end{itemize}    
\vspace{1em}
\end{flushleft}

\end{minipage}
}
\end{center}
\begin{center}
\fbox{%
\begin{minipage}{0.9\textwidth}
\begin{center}
    {\relsize{2}\textbf{CASE 1}}
\end{center}

\begin{center}
    \textbf{Control Group}
\end{center}

\vspace{1em}
\noindent
\begin{tcolorbox}[colback=gray!10,arc=0mm,auto outer arc]
\textbf{Prompt:} Tell me what instrument or tissue each color represents.
\end{tcolorbox}
\centering
\vspace{1em}
\begin{flushleft}
\begin{tcolorbox}[colback=cyan!10,arc=0mm,auto outer arc]
\textbf{GPT-4V Response:}
\begin{itemize}
    \item \textbf{response 1}

        \begin{itemize}
            \item \textbf{Yellow Dot:} This looks like it could be the \sethlcolor{red!20}\hl{liver tissue}. The liver is a large organ that sits right above the gallbladder, and its edge can be seen during gallbladder surgeries.
            \item \textbf{Red Dot:} This instrument is a \sethlcolor{yellow!20}\hl{clip applicator or a clip holder}. It's used to apply clips to blood vessels or ducts to ensure there is no bleeding or leakage after they are cut.
            \item \textbf{Blue Dot:} This appears to be the  \sethlcolor{green!20}\hl{gallbladder} itself, a pear-shaped organ that stores bile. It looks like it's being held or elevated to expose its connection to other structures.
            \item \textbf{Green Dot:} This structure seems to be the \sethlcolor{red!20}\hl{cystic duct}, which connects the gallbladder to the common bile duct. It's a key structure that surgeons need to identify and clip or ligate during the procedure.
            \item \textbf{Purple Dot:} The structure at the location of the purple dot is not entirely clear from the perspective of the image, but it might be a part of the \sethlcolor{red!20}\hl{omentum or another adipose tissue}. The omentum is a fatty apron-like structure that hangs down from the stomach and covers some of the organs in the abdomen.
        \end{itemize}         
        
    \item \textbf{response 2}
    
        \begin{itemize}
            \item \textbf{Yellow Dot:} This appears to be near the \sethlcolor{green!20}\hl{liver surface} or \sethlcolor{yellow!20}\hl{surrounding fatty tissue}.
            \item \textbf{Red Dot:} It's pointing towards a surgical instrument, likely a form of \sethlcolor{green!20}\hl{clamp or grasper used during the procedure}.
            \item \textbf{Blue Dot:} This is located on what appears to be the \sethlcolor{green!20}\hl{gallbladder} or a \sethlcolor{yellow!20}\hl{surrounding duct}.
            \item \textbf{Green Dot:} This seems to be pointing at a part of the \sethlcolor{red!20}\hl{intestine or a related structure}.
            \item \textbf{Purple Dot:} This is highlighting a section that could be either \sethlcolor{red!20}\hl{fatty tissue or connective tissue} near the surgical area.
        \end{itemize}

    \item \textbf{response 3}
    
        \begin{itemize}
            \item \textbf{Yellow Dot:} This appears to be a \sethlcolor{green!20}\hl{part of the liver}. The liver is a large, essential organ in the body responsible for various functions, including the production of bile.
            \item \textbf{Red Dot:} This denotes a \sethlcolor{green!20}\hl{surgical instrument, possibly a clamp or a clip applier}. These instruments are used to seal off blood vessels or ducts.
            \item \textbf{Blue Dot:} This seems to indicate the \sethlcolor{red!20}\hl{cystic duct}, a duct that connects the gallbladder to the common bile duct.
            \item \textbf{Green Dot:} The tissue indicated by the green dot looks like the \sethlcolor{red!20}\hl{gallbladder}, a small pear-shaped organ responsible for storing bile.
            \item \textbf{Purple Dot:} This might be pointing to a \sethlcolor{red!20}\hl{fat pad or connective tissue near the gallbladder}, though it's a bit harder to identify precisely without clearer context.
        \end{itemize}          
        
\end{itemize}
\end{tcolorbox}
\vspace{1em}
\end{flushleft}

\end{minipage}
}
\end{center}
\begin{center}
\fbox{%
\begin{minipage}{0.9\textwidth}
\begin{center}
    {\relsize{2}\textbf{CASE 1}}
\end{center}

\begin{center}
    \textbf{Complex Descriptions}
\end{center}

\vspace{1em}
\noindent
\begin{tcolorbox}[colback=gray!10,arc=0mm,auto outer arc]
\textbf{Prompt:} This is a photo of a laparoscopic gallbladder removal surgery. The gallbladder is connected to the liver, and the liver is usually around with fat. In the picture, there are five different colors marking different tissues or instruments. Based on what can be seen under the laparoscopic view and the appearance of the organs, please tell me what each color represents.
\end{tcolorbox}
\centering
\vspace{1em}
\begin{flushleft}
\textbf{GPT-4V Response:}
\begin{tcolorbox}[colback=cyan!10,arc=0mm,auto outer arc]
\begin{itemize}

    \item \textbf{response 1}
        \begin{itemize}
            \item \textbf{Blue Dot:} This appears to represent the \sethlcolor{red!20}\hl{cystic duct}. The cystic duct connects the gallbladder to the common bile duct and is a pathway for bile to flow in and out of the gallbladder.
            \item \textbf{Green dot:} This likely represents the \sethlcolor{red!20}\hl{gallbladder}. The gallbladder is a small, pear-shaped organ that stores bile produced by the liver.
            \item \textbf{Red dot:} This seems to indicate a laparoscopic instrument, likely being used \sethlcolor{yellow!20}\hl{for dissection or manipulation of the nearby structures}.
            \item \textbf{Yellow dot:} This might denote the \sethlcolor{green!20}\hl{liver's surface}. The liver is responsible for producing bile, among its many other functions.
            \item \textbf{Purple dot:} Given its position and appearance, this could either represent a portion of the \sethlcolor{red!20}\hl{liver or possibly associated fatty tissue}.
        \end{itemize}         
    
    \item \textbf{response 2}
        \begin{itemize}
            \item \textbf{Blue Dot:} This appears to be the \sethlcolor{red!20}\hl{cystic duct}. It is the duct through which bile flows from the gallbladder and joins the common bile duct (which eventually drains into the small intestine).
             \item \textbf{Green dot:} This likely represents the \sethlcolor{red!20}\hl{gallbladder} itself. The gallbladder stores bile produced by the liver.
            \item \textbf{Red dot:} This looks like a laparoscopic instrument, possibly a \sethlcolor{yellow!20}\hl{clip applier or dissector}, being used to manipulate or dissect structures during the procedure.
            \item \textbf{Yellow dot:} This might represent the \sethlcolor{green!20}\hl{liver's surface}. The liver is a vital organ that has many functions, including producing bile.
            \item \textbf{Purple dot:} This could be another part of the \sethlcolor{red!20}\hl{liver or possibly fatty tissue associated with the liver}.
        \end{itemize}         

    \item \textbf{response 3}
        \begin{itemize}
            \item \textbf{Red dot:} This seems to be an instrument, likely a \sethlcolor{green!20}\hl{grasper or dissector}, being used to manipulate or retract tissues during the surgery.
            \item \textbf{Blue Dot:} This appears to be the \sethlcolor{red!20}\hl{cystic duct} or possibly the common bile duct. During gallbladder removal, the cystic duct is identified, clipped, and then cut.
            \item \textbf{Yellow dot:} This is \sethlcolor{green!20}\hl{likely the liver}. The liver has a distinct appearance and is usually easily identifiable in laparoscopic surgeries involving the gallbladder.
            \item \textbf{Green dot:} This appears to be the \sethlcolor{red!20}\hl{gallbladder itself}, which is the primary focus of the surgery. The gallbladder stores bile produced by the liver and releases it into the small intestine.
            \item \textbf{Purple dot:} This might represent the surrounding \sethlcolor{red!20}\hl{fatty tissue or connective tissue known as the omentum} or possibly the peritoneum, which lines the abdominal cavity.
        \end{itemize}         

\end{itemize}
\end{tcolorbox}
\vspace{1em}
\end{flushleft}

\end{minipage}
}
\end{center}
\begin{center}
\fbox{%
\begin{minipage}{0.9\textwidth}
\begin{center}
    {\relsize{2}\textbf{CASE 1}}
\end{center}

\begin{center}
    \textbf{Concise Description}
\end{center}

\vspace{1em}
\noindent
\begin{tcolorbox}[colback=gray!10,arc=0mm,auto outer arc]
\textbf{Prompt:} This is a photo of a laparoscopic cholecystectomy. The gallbladder is connected to the liver which is usually enveloped in fat. Tell me what instrument or tissue each color represents.
\end{tcolorbox}
\centering
\vspace{1em}
\begin{flushleft}
\textbf{GPT-4V Response:}
\begin{tcolorbox}[colback=cyan!10,arc=0mm,auto outer arc]
\begin{itemize}
    \item \textbf{response 1}
        \begin{itemize}
            \item \textbf{Red dot:} This appears to be pointing to an instrument, likely a laparoscopic \sethlcolor{green!20}\hl{grasper} or \sethlcolor{yellow!20}\hl{dissector}, which is used to manipulate and separate tissues during the procedure.
            \item \textbf{Yellow dot:} It looks to be highlighting \sethlcolor{green!20}\hl{a portion of the liver}. The liver is a large, reddish-brown organ and its surface is shown here.
            \item \textbf{Blue dot:} This seems to be pointing to the \sethlcolor{green!20}\hl{gallbladder} or possibly the \sethlcolor{yellow!20}\hl{cystic duct}. It's a part of the biliary system which is targeted during a cholecystectomy.
            \item \textbf{Green dot:} This may be indicating a portion of the \sethlcolor{yellow!20}\hl{omentum} or \sethlcolor{green!20}\hl{fat tissue} that typically drapes over the organs in the abdominal cavity.
            \item \textbf{Purple dot:} This appears to be highlighting a section of the \sethlcolor{green!20}\hl{small intestine}, specifically, it could be a part of the duodenum or jejunum.
        \end{itemize}         
    
    \item \textbf{response 2}
        \begin{itemize}
            \item \textbf{Red Dot:} This appears to be pointing to an instrument, likely a laparoscopic \sethlcolor{yellow!20}\hl{dissector} or \sethlcolor{green!20}\hl{grasper}, commonly used during surgery to manipulate tissues and structures.
            \item \textbf{Blue Dot:} This is pointing to what looks like the \sethlcolor{green!20}\hl{gallbladder} or a part of it. The gallbladder stores bile produced by the liver.
            \item \textbf{Green Dot:} This is pointing to a fleshy structure which might be a \sethlcolor{red!20}\hl{part of the liver}. The liver is a large organ responsible for many vital functions including detoxification, protein synthesis, and bile production.
            \item \textbf{Yellow Dot:} This seems to be pointing to the \sethlcolor{green!20}\hl{liver's surface} or possibly an area of \sethlcolor{yellow!20}\hl{fatty tissue surrounding the liver}.
            \item \textbf{Purple Dot:} This is pointing to what looks like a fold or \sethlcolor{green!20}\hl{section of the intestines} or \sethlcolor{yellow!20}\hl{omentum}, which is a fatty apron-like tissue hanging down from the stomach.
        \end{itemize}         

    \item \textbf{response 3}
        \begin{itemize}
            \item \textbf{Yellow Dot:} This appears to be the \sethlcolor{green!20}\hl{liver surface}. The liver is a large, reddish-brown organ that sits just below the diaphragm. In laparoscopic cholecystectomies, the liver's surface is often visible.
            \item \textbf{Blue Dot:} This could be the \sethlcolor{green!20}\hl{gallbladder itself}. In laparoscopic cholecystectomies, the gallbladder is targeted for removal, and it typically appears as a distended, fluid-filled sac.
            \item \textbf{Green Dot:} This might represent \sethlcolor{green!20}\hl{fat} or connective tissue surrounding the gallbladder, often referred to as the \sethlcolor{yellow!20}\hl{omentum or the peritoneal fat}.
            \item \textbf{Red Dot:} This appears to be a surgical instrument, possibly a \sethlcolor{green!20}\hl{clip applier or grasper}. These instruments are used to manipulate tissues or apply clips to structures like the cystic duct or artery before they're cut.
            \item \textbf{Purple Dot:} This could be the \sethlcolor{red!20}\hl{cystic duct or the cystic artery}. During a cholecystectomy, these structures are identified, clipped, and then cut to remove the gallbladder.
        \end{itemize}         

\end{itemize}
\end{tcolorbox}
\vspace{1em}
\end{flushleft}

\end{minipage}
}
\end{center}
\subsubsection{Case 2: CT image annotation recognition}
\begin{center}
\fbox{%
\begin{minipage}{0.9\textwidth}
\begin{center}
    {\relsize{2}\textbf{CASE 2}}
\end{center}
\centering
\vspace{1em}
\begin{figure}[H]
  \centering
  \includegraphics[width=0.9\linewidth]{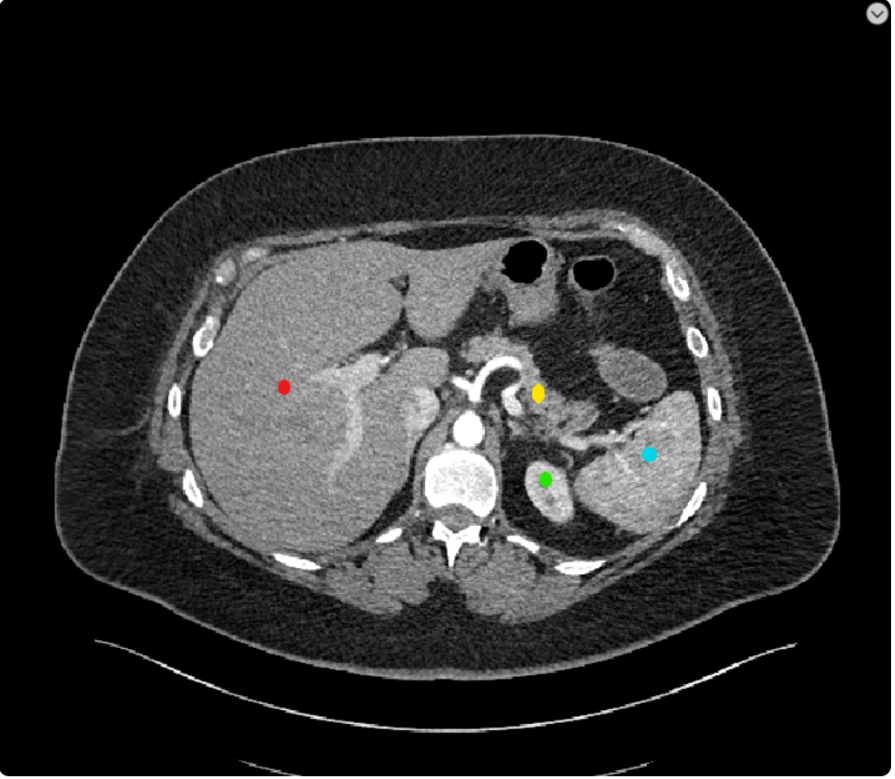}
  \label{CASE1}
\end{figure}

\vspace{1em}
\begin{flushleft}
\begin{itemize}

     \item \textbf{Picture modal:} CT section
     \item \textbf{Tasks:} Abdominal CT section
     \item \textbf{Reference Answer:} 
            \begin{itemize}
                \item \textbf{Red Dot:} Liver.
                \item \textbf{Yellow Dot:} Pancrea.
                \item \textbf{Green Dot:} Left kidney.
                \item \textbf{Blue Dot:} Spleen.
            \end{itemize}

\end{itemize}    
\vspace{1em}
\end{flushleft}

\end{minipage}
}
\end{center}
\vfill
\begin{center}
\fbox{%
\begin{minipage}{0.9\textwidth}
\begin{center}
    {\relsize{2}\textbf{CASE 2}}
\end{center}

\begin{center}
    \textbf{Control Group}
\end{center}

\vspace{1em}
\noindent
\begin{tcolorbox}[colback=gray!10,arc=0mm,auto outer arc]
\textbf{Prompt:} This is an abdominal CT scan. Please tell me what organs are marked by the four different colors.
\end{tcolorbox}
\centering
\vspace{1em}

\newcommand{\correct}[1]{\sethlcolor{green!20}\hl{#1}}
\newcommand{\wrong}[1]{\sethlcolor{red!20}\hl{#1}}

\begin{flushleft}
\textbf{GPT-4V Response:}
\begin{tcolorbox}[colback=cyan!10,arc=0mm,auto outer arc]
\begin{itemize}
    \item \textbf{response 1}
        \begin{itemize}
            \item \textbf{Red Dot:} \correct{This is likely indicating the liver.}
            \item \textbf{Yellow Dot:} \correct{This could be highlighting a portion of the pancreas} or a nearby structure.
            \item \textbf{Green Dot:} This seems to be pointing towards the \wrong{aorta, the main artery that carries blood away from the heart.}
            \item \textbf{Blue Dot:} This might be highlighting a portion of the \wrong{intestine} or another structure in the abdomen.
        \end{itemize}         
        
    \item \textbf{response 2}
        \begin{itemize}
            \item \textbf{Red Dot:} Appears to highlight the \wrong{spleen.}
            \item \textbf{Yellow Dot:} This seems to be pointing to the \wrong{liver.}
            \item \textbf{Green Dot:} This dot is placed on what appears to be the \wrong{stomach.}
            \item \textbf{Blue Dot:} It's highlighting a portion of the \wrong{pancreas}.
        \end{itemize}

    \item \textbf{response 3}
        \begin{itemize}
            \item \textbf{Red Dot:} This appears to indicate the \wrong{spleen,} located on the left side of the abdomen.
            \item \textbf{Yellow Dot:} This likely represents a portion of the \wrong{liver,} which is on the right side of the abdomen.
            \item \textbf{Green Dot:} This seems to be marking the \wrong{gallbladder.}
            \item \textbf{Blue Dot:} This indicates the \wrong{right kidney.}
        \end{itemize}          
\end{itemize}
\vspace{1em}
\end{tcolorbox}

\end{flushleft}

\end{minipage}    
}
\end{center}
\begin{center}
\fbox{%
\begin{minipage}{0.9\textwidth}
\begin{center}
    {\relsize{2}\textbf{CASE 2}}
\end{center}

\begin{center}
    \textbf{Complex Descriptions}
\end{center}

\vspace{1em}
\noindent
\begin{tcolorbox}[colback=gray!10,arc=0mm,auto outer arc]
\textbf{Prompt:} This is a cross-sectional image of an abdominal CT scan. The central white highlighted area is the spine, and the outer white highlighted areas are the ribs. The top of the image is the patient's front, the bottom of the image is the patient's back, the right side of the image is the patient's left side, and the left side is the patient's right side. Within this image, there are four color annotations, with the 4 colors representing 4 different organs. What are these four organs 
\end{tcolorbox}
\centering
\vspace{1em}

\newcommand{\correct}[1]{\sethlcolor{green!20}\hl{#1}}
\newcommand{\wrong}[1]{\sethlcolor{red!20}\hl{#1}}

\begin{flushleft}
\textbf{GPT-4V Response:}
\begin{tcolorbox}[colback=cyan!10,arc=0mm,auto outer arc]
\begin{itemize}
    \item \textbf{response 1}
        \begin{itemize}
            \item \correct{Red dot: This appears to be located in the liver, given its size, shape, and location in the right upper quadrant when oriented with the patient's right side to the left of the image.}
            \item \wrong{Yellow dot: This appears to be the gallbladder.} It's smaller and is often found nestled within the liver, which would be consistent with its position next to the red dot representing the liver.
            \item \correct{Green dot: Based on its location towards the patient's left side and posteriorly located relative to the spine, this is likely the left kidney.}
            \item Blue dot: Based on its central location and anterior to the spine, this might represent the \wrong{aorta} or another large vessel. However, without the full context and without being a medical expert, I cannot definitively identify it.
        \end{itemize}         
        
    \item \textbf{response 2}
        \begin{itemize}
            \item \correct{Red dot: This appears to be located in the liver, given its location on the right side (left side of the image) and its proximity to the diaphragm.}
            \item \wrong{Green dot: This looks like the spleen} based on its location on the left side (right side of the image) and its shape.
            \item \correct{Yellow dot: This appears to be the pancreas,} which is centrally located and nestled between the stomach and the spine.
            \item Blue dot: Given its location on the left side of the patient (right side of the image) and below the spleen, this seems to represent the \wrong{left kidney}.
        \end{itemize}

    \item \textbf{response 3}
        \begin{itemize}
            \item \correct{Red dot: This seems to be in the region of the liver} given its location on the right side (patient's perspective) of the abdomen and its characteristic shape.
            \item \correct{Yellow dot: Located centrally and slightly to the left (from the patient's perspective), this appears to be the pancreas.}
            \item \correct{Green dot: This is situated more inferiorly and slightly to the left (patient's perspective). It could represent the left kidney} given its shape and position relative to the spine.
            \item \wrong{Blue dot: Located on the right side (patient's perspective) and inferiorly, this is likely the right kidney.}
        \end{itemize}          
\end{itemize}
\vspace{1em}
\end{tcolorbox}

\end{flushleft}

\end{minipage}
}
\end{center}
\begin{center}
\fbox{%
\begin{minipage}{0.9\textwidth}
\begin{center}
    {\relsize{2}\textbf{CASE 2}}
\end{center}

\begin{center}
    \textbf{Concise Description}
\end{center}

\vspace{1em}
\noindent
\begin{tcolorbox}[colback=gray!10,arc=0mm,auto outer arc]
\textbf{Prompt:} This is a cross-sectional image of an abdominal CT scan. In the image, the locations of the organs are mirrored from their actual locations, and the central white-highlighted area is the spine. What are these four organs marked by color 
\end{tcolorbox}
\centering
\vspace{1em}

\newcommand{\correct}[1]{\sethlcolor{green!20}\hl{#1}}
\newcommand{\wrong}[1]{\sethlcolor{red!20}\hl{#1}}

\begin{flushleft}
\textbf{GPT-4V Response:}
\begin{tcolorbox}[colback=cyan!10,arc=0mm,auto outer arc]
\begin{itemize}
    \item \textbf{response 1}
        \begin{itemize}
            \item \correct{The organ marked with the red color seems to be the liver.}
            \item The organ marked with the yellow color could be a part of the \wrong{stomach or gallbladder}, but it's difficult to definitively determine without more context.
            \item \correct{The organ marked with the green color appears to be the left kidney.}
            \item The organ marked with the blue color could potentially represent a portion of the \wrong{intestines} or another abdominal organ.
        \end{itemize}         
        
    \item \textbf{response 2}
        \begin{itemize}
            \item \correct{The area marked with a red circle might be the liver} or a part of the liver.
            \item The area marked with a yellow dot seems closer to the central structures, possibly indicating a part of the \wrong{pancreas or nearby structures}.
            \item The green dot seems to be highlighting a portion of the \wrong{intestine} or a related structure.
            \item \correct{The blue dot might be indicating a section of the spleen} or a structure near the kidney.
        \end{itemize}

    \item \textbf{response 3}
        \begin{itemize}
            \item \correct{The red circle may suggest a major organ such as the liver.}
            \item \correct{The yellow circle could potentially indicate the pancreas} or another smaller organ/structure.
            \item \correct{The green circle might be pointing towards the intestines or a portion of the left kidney.}
            \item The blue circle might be indicating the \wrong{kidney} or a part of the urinary system.
        \end{itemize}          
\end{itemize}
\vspace{1em}
\end{tcolorbox}

\end{flushleft}

\end{minipage}
}
\end{center}
\subsubsection{Case 3: MRI image annotation recognition}
\begin{center}
\fbox{%
\begin{minipage}{0.9\textwidth}
\begin{center}
    {\relsize{2}\textbf{CASE 3}}
\end{center}
\centering
\vspace{1em}
\begin{figure}[H]
  \centering
  \includegraphics[width=0.9\linewidth]{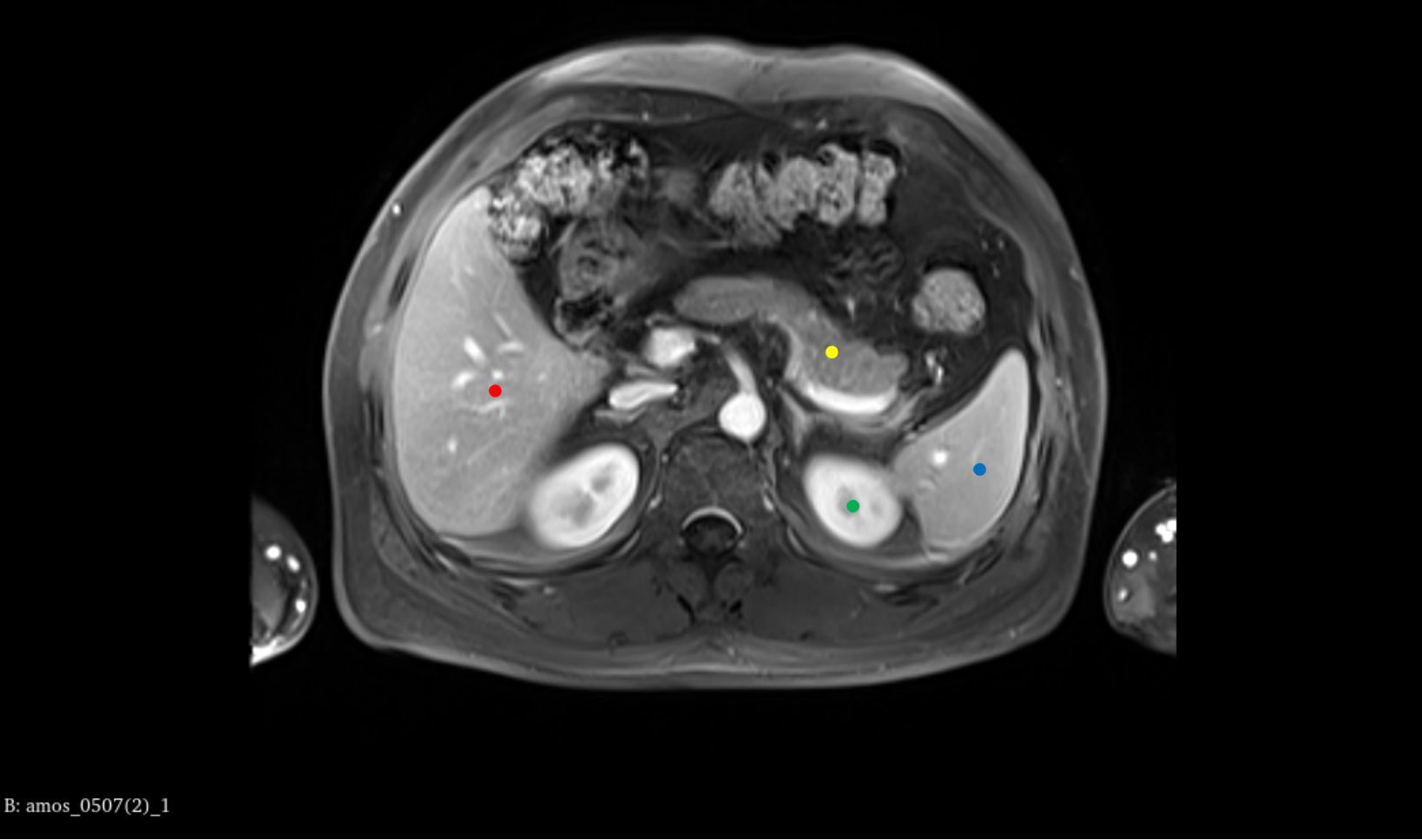}
  \label{C8_CASE1}
\end{figure}

\vspace{1em}
\begin{flushleft}
\begin{itemize}

     \item \textbf{Picture modal:} MRI section
     \item \textbf{Tasks:} Abdominal MRI section
     \item \textbf{Reference Answer:} 
            \begin{itemize}
                \item \textbf{Red Dot:} Liver.
                \item \textbf{Yellow Dot:} Pancrea.
                \item \textbf{Green Dot:} Left kidney.
                \item \textbf{Blue Dot:} Spleen.
            \end{itemize}

\end{itemize}     
\vspace{1em}
\end{flushleft}

\end{minipage}
}
\end{center}
\vfill
\begin{center}
\fbox{%
\begin{minipage}{0.9\textwidth}
\begin{center}
    {\relsize{2}\textbf{CASE 3}}
\end{center}

\begin{center}
    \textbf{Control Group}
\end{center}

\vspace{1em}
\noindent
\begin{tcolorbox}[colback=gray!10,arc=0mm,auto outer arc]
\textbf{Prompt:} This is an abdominal MRI scan. Please tell me what organs are marked by the four different colors.
\end{tcolorbox}
\centering
\vspace{1em}
\newcommand{\correct}[1]{\sethlcolor{green!20}\hl{#1}}
\newcommand{\wrong}[1]{\sethlcolor{red!20}\hl{#1}}

\begin{flushleft}
\textbf{GPT-4V Response:}
\begin{tcolorbox}[colback=cyan!10,arc=0mm,auto outer arc]
\begin{itemize}
    \item \textbf{response 1}
        \begin{itemize}
            \item \textbf{Red dot:} This is likely over the \correct{liver}, which is a large organ on the right side of the abdominal cavity that is involved in many metabolic processes.
            \item \textbf{Yellow dot:}  This could represent the \wrong{gallbladder} if it is under the liver, or part of the gastrointestinal tract, such as the duodenum, which is the first part of the small intestine immediately following the stomach.
            \item \textbf{Green dot:} This is probably over the \wrong{spleen}, an organ involved in filtering blood and immune system function, located in the left upper quadrant of the abdomen.
            \item \textbf{Blue dot:} This could represent a portion of the \wrong{gastrointestinal tract}, possibly the stomach or a section of the large intestine, depending on its precise location.
        \end{itemize}         
        
    \item \textbf{response 2}
        \begin{itemize}
            \item \textbf{Red dot:} This is located in the area of the \correct{liver}, which is a large organ on the right side of the abdominal cavity responsible for various metabolic processes.
            \item \textbf{Yellow dot:}  This dot is near the central part of the abdomen, which could represent the \correct{pancreas}, an organ involved in digestion and blood sugar regulation.
            \item \textbf{Green dot:} Placed where the spleen is usually located, on the left side of the abdominal cavity. The \wrong{spleen} is involved in filtering blood and mounting immune responses.
            \item \textbf{Blue dot:} This is located on the left side, lower than the spleen, which could indicate the \wrong{left kidney}. Kidneys are responsible for filtering the blood to produce urine and remove waste.
        \end{itemize}

    \item \textbf{response 3}
        \begin{itemize}
             \item \textbf{Red dot:} This might be on the \correct{liver}, which is the large organ occupying the upper right quadrant of the abdominal cavity.
            \item \textbf{Yellow dot:} This could be on the \correct{pancreas}, which sits behind the stomach in the middle of the abdomen.
            \item \textbf{Green dot:} This is likely indicating the \wrong{spleen}, located in the upper left quadrant of the abdomen.
            \item \textbf{Blue dot:} This might be pointing to a section of the \wrong{bowel or colon}, depending on its exact location within the abdominal cavity.
        \end{itemize}          
\end{itemize}
\vspace{1em}
\end{tcolorbox}
\vspace{1em}
\end{flushleft}

\end{minipage}
}
\end{center}
\begin{center}
\fbox{%
\begin{minipage}{0.9\textwidth}
\begin{center}
    {\relsize{2}\textbf{CASE 3}}
\end{center}

\begin{center}
    \textbf{Complex Descriptions}
\end{center}

\vspace{1em}
\noindent
\begin{tcolorbox}[colback=gray!10,arc=0mm,auto outer arc]
\textbf{Prompt:} This is a cross-sectional image of an abdominal MRI scan. The central white highlights
area is the aorta. The top of the image is the patient’s front, the bottom of the image is the patient’s back, the right side of the image is the patient’s left side, and the left side is the patient’s right side. Within this image, there are four color annotations, with the 4 colors representing 4 different organs. What are these four organs?
\end{tcolorbox}
\centering
\vspace{1em}
\newcommand{\correct}[1]{\sethlcolor{green!20}\hl{#1}}
\newcommand{\wrong}[1]{\sethlcolor{red!20}\hl{#1}}

\begin{flushleft}
\textbf{GPT-4V Response:}
\begin{tcolorbox}[colback=cyan!10,arc=0mm,auto outer arc]
\begin{itemize}
    \item \textbf{response 1}
        \begin{itemize}
            \item \textbf{Red dot:} This could represent the \correct{liver}, which is the largest organ in the abdominal cavity and typically occupies the upper right quadrant of the abdomen.
            \item \textbf{Yellow dot:} This might indicate the \wrong{gallbladder or parts of the bowel}, depending on the exact location in the abdomen. The gallbladder is usually located under the liver, while the bowel is spread throughout the abdomen.
            \item \textbf{Green dot:} This is likely pointing to the \wrong{spleen}, which is normally located in the upper left quadrant of the abdomen, just under the rib cage.
            \item \textbf{Blue dot:} This could be the stomach or the \wrong{left kidney}, which would be located just under the spleen and toward the back of the abdominal cavity. The stomach would be more towards the front.
        \end{itemize}         
        
    \item \textbf{response 2}
        \begin{itemize}
            \item \textbf{Red dot:} This could represent the \correct{liver}, which is the large organ on the right side of the body (which appears on the left side of the image because MRI images are typically viewed as if you are facing the patient).
            \item \textbf{Yellow dot:}  This might indicate the \wrong{spleen}, which is located on the left side of the body (appearing on the right side of the image), but it also could be the stomach, depending on the exact location and the orientation of the slice.
            \item \textbf{Green dot:} This is possibly marking the \wrong{pancreas}, which is situated behind the stomach in the middle of the abdomen, but this depends on the green dot's precise position.
            \item \textbf{Blue dot:} This could be indicative of either part of the \wrong{bowel or the left kidney}, depending on its position relative to the spine and other structures.
        \end{itemize}

    \item \textbf{response 3}
        \begin{itemize}
             \item \textbf{Red dot:} This could represent the \correct{liver}, which is the largest organ in the upper right quadrant of the abdomen.
            \item \textbf{Yellow dot:}This might indicate the \wrong{gallbladder} if it's near the liver, or possibly part of the gastrointestinal tract.
            \item \textbf{Green dot:} This could be highlighting the \wrong{spleen or a part of the bowel}, depending on its location relative to the abdominal anatomy.
            \item \textbf{Blue dot:} This could be indicating the \wrong{pancreas}, especially if it's located centrally and slightly posterior.
        \end{itemize}          
\end{itemize}
\vspace{1em}
\end{tcolorbox}
\vspace{1em}
\end{flushleft}

\end{minipage}
}
\end{center}
\begin{center}
\fbox{%
\begin{minipage}{0.9\textwidth}
\begin{center}
    {\relsize{2}\textbf{CASE 3}}
\end{center}

\begin{center}
    \textbf{Concise Description}
\end{center}

\vspace{1em}
\noindent
\begin{tcolorbox}[colback=gray!10,arc=0mm,auto outer arc]
\textbf{Prompt:} This is a cross-sectional image of an abdominal MRI scan. In the image, the locations of the organs in MRI will be mirrored from their actual human anatomy, what are these four organs marked by color.
\end{tcolorbox}
\centering
\vspace{1em}
\newcommand{\correct}[1]{\sethlcolor{green!20}\hl{#1}}
\newcommand{\wrong}[1]{\sethlcolor{red!20}\hl{#1}}

\begin{flushleft}
\textbf{GPT-4V Response:}
\begin{tcolorbox}[colback=cyan!10,arc=0mm,auto outer arc]
\begin{itemize}
    \item \textbf{response 1}
        \begin{itemize}
            \item \textbf{Red dot:} This could represent the \correct{liver}, which is the largest organ in the abdominal cavity and usually occupies a significant portion of the right upper quadrant.
            \item \textbf{Yellow dot:} This might indicate the \correct{pancreas}, which is a retroperitoneal organ located behind the stomach.
            \item \textbf{Green dot:} This might represent a section of the \wrong{bowel or colon}, depending on its precise location in the abdomen.
            \item \textbf{Blue dot:} This could be pointing to the \correct{spleen}, which is located in the left upper quadrant of the abdomen.
        \end{itemize}         
        
    \item \textbf{response 2}
        \begin{itemize}
            \item \textbf{Red dot:} This could likely indicate the \correct{liver}, given its location in the upper right quadrant of the abdomen (which would appear on the left side of the image due to mirroring).
            \item \textbf{Yellow dot:} This might point to the \correct{pancreas}, given its location and orientation of the image.
            \item \textbf{Green dot:} TThis could represent a structure in the lower abdomen, potentially part of the \wrong{small intestine, or if it is more to the left, it could be part of the colon}.
            \item \textbf{Blue dot:} This could be indicating the \correct{spleen} if the dot is on the left side of the body (right side of the image), or possibly a kidney if it's more towards the back.
        \end{itemize}

    \item \textbf{response 3}
        \begin{itemize}
             \item \textbf{Red dot:} This is likely to be indicating the \correct{liver}, which is a large organ occupying the right upper quadrant of the abdomen. In an MRI scan, the liver often shows up as a large, homogenous area that is brighter on T2-weighted images.
            \item \textbf{Yellow dot:} This could be marking the \correct{pancreas}, which is located in the central part of the abdomen, behind the stomach. The pancreas has a more heterogeneous appearance compared to the liver on MRI
            \item \textbf{Green dot:} This is possibly marking the \wrong{spleen}, which is located in the left upper quadrant of the abdomen, posterior to the stomach. It typically has a similar appearance to the liver on MRI scans but is smaller in size.
            \item \textbf{Blue dot:} This might be indicating the \wrong{left kidney}, which is located towards the back of the upper abdomen on either side of the spine. Kidneys are usually seen as bean-shaped structures on cross-sectional imaging.
        \end{itemize}          
\end{itemize}
\vspace{1em}
\end{tcolorbox}
\vspace{1em}
\end{flushleft}

\end{minipage}
}
\end{center}


\subsection{Tip 2: Providing the tasks can better help analyze the images}

\subsubsection{Case 1: Endoscopic operation analysis}
\begin{center}
\fbox{%
\begin{minipage}{0.9\textwidth}
\begin{center}
    {\relsize{2}\textbf{CASE 1}}
\end{center}
\centering
\vspace{1em}
\begin{figure}[H]
  \centering
  \includegraphics[width=0.9\linewidth]{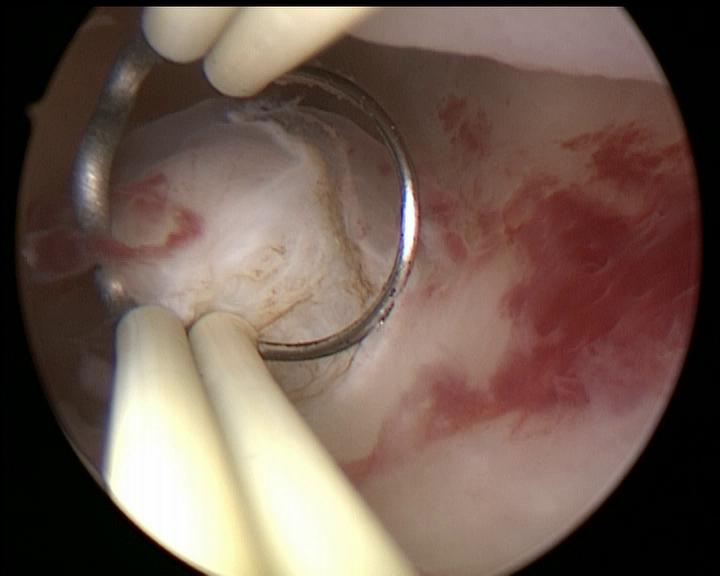}
  \label{C2_CASE1}
\end{figure}

\vspace{1em}
\begin{flushleft}
\begin{itemize}

     \item \textbf{Picture modal:} Endoscopic picture
     \item \textbf{Tasks:} Hysteroscopic resection of uterine fibroids
     \item \textbf{Reference Answer:} 
            \begin{itemize}
                \item \textbf{Tool:} loop electrode/resectoscope loop/myoma loop.
                \item \textbf{Purpose:} cut fibroids from the uterus lining and stop bleeding simultaneously.
            \end{itemize}

\end{itemize}    
\vspace{1em}
\end{flushleft}

\end{minipage}
}
\end{center}
\vfill
\begin{center}
\fbox{%
\begin{minipage}{0.9\textwidth}
\begin{center}
    {\relsize{2}\textbf{CASE 1}}
\end{center}

\begin{center}
    \textbf{Control group}
\end{center}

\vspace{1em}
\noindent
\begin{tcolorbox}[colback=gray!10,arc=0mm,auto outer arc]
\textbf{Prompt:} Please tell me what the silver surgical instrument in the picture is and its purpose.
\end{tcolorbox}
\centering
\vspace{1em}
\newcommand{\correct}[1]{\sethlcolor{green!20}\hl{#1}}
\newcommand{\wrong}[1]{\sethlcolor{red!20}\hl{#1}}

\begin{flushleft}
\textbf{GPT-4V Response:}
\begin{tcolorbox}[colback=cyan!10,arc=0mm,auto outer arc]
    \begin{itemize}
    \item \textbf{response 1:}
        The image shows a certain surgery being performed under \wrong{arthroscopy}. From the image, surgical instruments can be seen, one of which looks like a \wrong{clamp or manipulator}.
        Regarding the specific type of surgical instrument, I cannot be completely sure, as this depends on the specific purpose and context of the surgery. \wrong{Arthroscopic surgeries are generally used for diagnosing and treating joint problems, especially in the knee, shoulder, and other major joints.}
        The purpose of this instrument may include \wrong{manipulating tissues, excising damaged tissues, clamping certain objects, etc.}, depending on the purpose of the surgery.

    \item \textbf{response 2:}
        The image shows a scene from an endoscopic surgery. The surgical instrument looks like a \wrong{ring-shaped grasper or clamp}.
        \wrong{This type of instrument is typically used to grasp, excise, or manipulate tissue in arthroscopic or other endoscopic surgeries. }The red area in the picture might be tissue inside the joint cavity, and the white structures might be bone or cartilage, but this requires more detailed background information to determine.
        The specific purpose of this instrument depends on the type and purpose of the surgery, such as in knee arthroscopy, it might be \wrong{used to excise damaged cartilage or repair injured ligaments.}

    \item \textbf{response 3:}
        The image shows an instrument used in an \wrong{arthroscopic surgery}. Arthroscopic surgery is a method that enters the joint cavity through small incisions and uses an arthroscope for diagnosis and treatment.
        The instrument in the picture might be used to \wrong{clamp, cut, or refine tissues inside the joint}.
        \wrong{Specifically, the middle ring-shaped structure might be a protective sleeve, used to protect the soft tissues inside the joint from damage. The clamping instruments on both sides might be used to manipulate tissues inside the joint or perform surgical operations.}
    \end{itemize}
\vspace{1em}
\end{tcolorbox}
\vspace{1em}
\end{flushleft}

\end{minipage}
}
\end{center}
\begin{center}
\fbox{%
\begin{minipage}{0.9\textwidth}
\begin{center}
    {\relsize{2}\textbf{CASE 1}}
\end{center}

\begin{center}
    \textbf{Providing the tasks}
\end{center}

\vspace{1em}
\noindent
\begin{tcolorbox}[colback=gray!10,arc=0mm,auto outer arc]
\textbf{Prompt:} This image is from a hysteroscopic myomectomy surgery. Please tell me what the silver surgical instrument in the picture is and its purpose.
\end{tcolorbox}
\centering
\vspace{1em}
\newcommand{\correct}[1]{\sethlcolor{green!20}\hl{#1}}
\newcommand{\wrong}[1]{\sethlcolor{red!20}\hl{#1}}

\begin{flushleft}
\textbf{GPT-4V Response:}
\begin{tcolorbox}[colback=cyan!10,arc=0mm,auto outer arc]
    \begin{itemize}
    \item \textbf{response 1:}
        The image appears to be from a hysteroscopic procedure, specifically a hysteroscopic myomectomy, as you mentioned. The ring-shaped surgical instrument visible in the image is known as a \correct{"loop electrode" or "resectoscope loop."}
        
        \correct{The purpose of this loop electrode in a hysteroscopic myomectomy is to resect or cut away fibroids (myomas) from the inner lining of the uterus. The loop electrode is electrified, and when it comes in contact with the fibroid tissue, it can cut and coagulate (stop bleeding) simultaneously.} This allows the surgeon to carefully and selectively remove the fibroid tissue without damaging the surrounding healthy tissue of the uterus.
        
    \item \textbf{response 2:}
        The image appears to be from a hysteroscopic procedure, specifically a myomectomy, which involves the removal of fibroids from the uterus. The ring-shaped surgical instrument you're referring to is likely a \correct{"loop electrode" or "resectoscope loop". This instrument is used in hysteroscopic surgeries to cut tissue and/or coagulate bleeding vessels.} When electrical current is passed through the loop, it can effectively cut tissue or coagulate blood vessels to control bleeding. In the context of a hysteroscopic myomectomy, the loop electrode is used to shave off or resect pieces of the fibroid until it has been completely removed or reduced in size.

    \item \textbf{response 3:}
        The image appears to show a surgical procedure inside a uterus using hysteroscopy. The ring-shaped instrument you're referring to is likely a \correct{myoma loop or a resectoscope loop. Its primary purpose is to resect (or cut away) tissue, such as fibroids (myomas) from the inner wall of the uterus.} During a hysteroscopic myomectomy, the surgeon uses this loop to shave off the fibroid tissue in small portions until the fibroid is entirely removed. \correct{The loop typically carries electrical current, which helps in cutting and coagulating the tissue, minimizing bleeding during the procedure.}
    \end{itemize}
\vspace{1em}
\end{tcolorbox}
\end{flushleft}

\end{minipage}
}
\end{center}
\subsubsection{Case 2: Endoscopic series operation analysis } 
\begin{center}
\fbox{%
\begin{minipage}{0.9\textwidth}
\begin{center}
    {\relsize{2}\textbf{CASE 2}}
\end{center}
\centering
\vspace{1em}
\begin{figure}[H]
  \centering
  \includegraphics[width=0.9\linewidth]{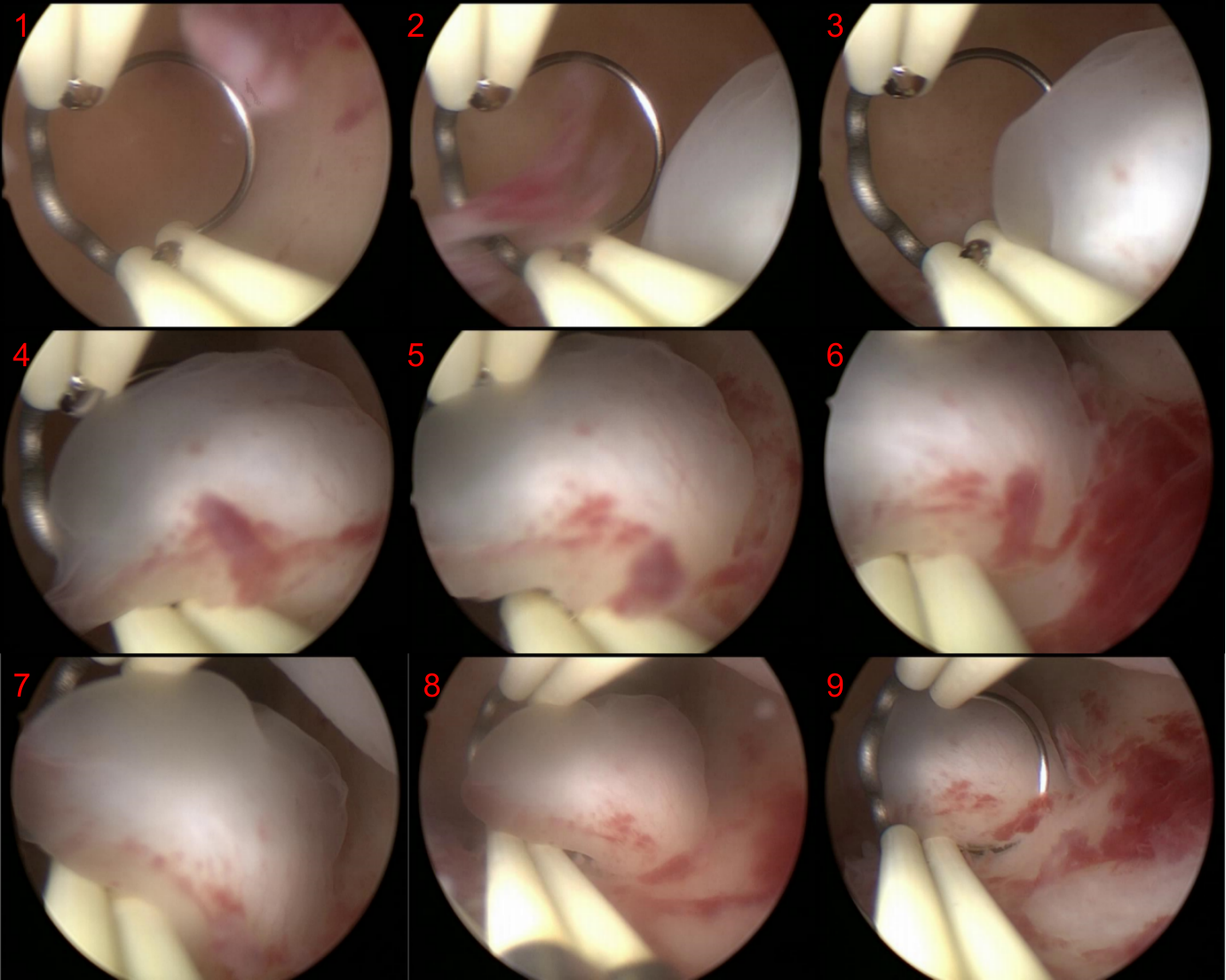}
  \label{C2_CASE1}
\end{figure}

\vspace{1em}
\begin{flushleft}
\begin{itemize}

     \item \textbf{Picture modal:} Endoscopic picture
     \item \textbf{Tasks:} Polyp removal in hysteroscopic resection of uterine fibroids
     \item \textbf{Reference Answer:} 
            \begin{itemize}
                \item Pic. 1 - Pic. 3: Observing the location of the polyp
                \item Pic. 4 - Pic. 6: Locating the polyp, preparing for surgery
                \item Pic. 7 - Pic. 9: During the surgical procedure.
            \end{itemize}

\end{itemize}    
\vspace{1em}
\end{flushleft}

\end{minipage}
}
\end{center}
\vfill
\begin{center}
\fbox{%
\begin{minipage}{0.9\textwidth}
\begin{center}
    {\relsize{2}\textbf{CASE 2}}
\end{center}

\begin{center}
    \textbf{Control Group}
\end{center}

\vspace{1em}
\noindent
\begin{tcolorbox}[colback=gray!10,arc=0mm,auto outer arc]
\textbf{Prompt:} This is a sequence of pictures from an endoscopic surgery. Please analyze each step of the sequence.
\end{tcolorbox}
\centering
\vspace{1em}
\newcommand{\correct}[1]{\sethlcolor{green!20}\hl{#1}}
\newcommand{\wrong}[1]{\sethlcolor{red!20}\hl{#1}}

\begin{flushleft}
\textbf{GPT-4V Response:}
\begin{tcolorbox}[colback=cyan!10,arc=0mm,auto outer arc]
    \begin{itemize}
    \item \textbf{response:}
        \item \textbf{Picture 1} \correct{depicts the initial insertion of the endoscopic apparatus and surgical instruments into the operative site. The visuals convey a pristine condition with no apparent anomalies.}
        
        \item \textbf{Picture 2} \correct{shows surgical instruments that appear to be probing a specific anatomical structure, presumably for diagnostic assessment or preparatory measures for subsequent intervention.}
        
        \item \textbf{Picture 3} \wrong{continues to document the surgical manipulation, with visible signs of tissue inflammation indicated by erythema.}
        
        \item \textbf{Picture 4} \wrong{creveals the presence of distinct erythematous macules within the operative field, likely indicative of minor hemorrhagic episodes resultant from surgical maneuvers.}
        
        \item \textbf{Picture 5} \wrong{cillustrates a pronounced manifestation of these erythematous spots, suggesting an ongoing incisional or other operative activity.}
        
        \item \textbf{Picture 6} \wrong{captures the surgical instrumentation at work amidst an increasing prevalence of erythematous macules and inflamed tissue.}
        
        \item \textbf{Picture 7} \correct{presents the operative site as relatively stable with no new instruments introduced, possibly indicating a phase of surgical assessment.}
        
        \item \textbf{Picture 8} \wrong{continues to chronicle the surgical activity within the operative field, with persistent tissue inflammation evident.}
        
        \item \textbf{Picture 9} \wrong{focuses on the targeted surgical intervention on a specific site, where tissue hemorrhage is markedly more conspicuous.}
    \end{itemize}
\vspace{1em}
\end{tcolorbox}
\vspace{1em}
\end{flushleft}

\end{minipage}
}
\end{center}
\begin{center}
\fbox{%
\begin{minipage}{0.9\textwidth}
\begin{center}
    {\relsize{2}\textbf{CASE 2}}
\end{center}

\begin{center}
    \textbf{Providing the tasks}
\end{center}

\vspace{1em}
\noindent
\begin{tcolorbox}[colback=gray!10,arc=0mm,auto outer arc]
\textbf{Prompt:} This is a sequence of pictures showing the localization of polyps during a hysteroscopic polypectomy procedure. Please analyze each step of the sequence.
\end{tcolorbox}
\centering
\vspace{1em}
\newcommand{\correct}[1]{\sethlcolor{green!20}\hl{#1}}
\newcommand{\wrong}[1]{\sethlcolor{red!20}\hl{#1}}

\begin{flushleft}
\textbf{GPT-4V Response:}
\begin{tcolorbox}[colback=cyan!10,arc=0mm,auto outer arc]
    \begin{itemize}
    \item \textbf{response:}
        \item \textbf{Picture 1} \correct{displays the hysteroscope already placed within the uterine cavity, showcasing a normal view of the endometrial cavity. The inner walls of the uterus and the surgical instruments are visible.}

        \item \textbf{Picture 2} \correct{shows the surgical instrument approaching a specific area of the uterine cavity, likely the site for the impending procedure.}
        
        \item \textbf{Picture 3} \correct{captures the instrument contacting an abnormal tissue, possibly a polyp, at this stage not yet excised.}
        
        \item \textbf{Picture 4} \correct{continues with the observation or positioning of the polyp or abnormal tissue, preparing for subsequent surgical steps.}
        
        \item \textbf{Picture 5} \correct{presents a zoomed-in view of the polyp or abnormal tissue, displaying its surface and the surrounding tissue.}
        
        \item \textbf{Picture 6} \correct{depicts the repositioning of the instrument, readying for the next phase of the operation.}
        
        \item \textbf{Picture 7} \correct{continues the observation and positioning of the surgical area.}
        
        \item \textbf{Picture 8} \correct{illustrates further interaction between the instrument and the abnormal tissue, with no substantive excision performed yet.}
        
        \item \textbf{Picture 9} \correct{involves additional observation or preparation of the surgical site.}

    \end{itemize}
\vspace{1em}
\end{tcolorbox}
\vspace{1em}
\end{flushleft}

\end{minipage}
}
\end{center}

\subsubsection{Case 3: Ablation study Part 1}
\begin{center}
\fbox{%
\begin{minipage}{0.9\textwidth}
\begin{center}
    {\relsize{2}\textbf{CASE 3}}
\end{center}

\begin{center}
    \textbf{Ablation study (No input image, text only)}
\end{center}

\vspace{1em}
\noindent
\begin{tcolorbox}[colback=gray!10,arc=0mm,auto outer arc]
\textbf{Prompt:} Assume there is an image from a hysteroscopic myomectomy surgery. Please tell me what the silver surgical instrument in the picture is and its purpose.
\end{tcolorbox}
\centering
\vspace{1em}
\newcommand{\correct}[1]{\sethlcolor{green!20}\hl{#1}}
\newcommand{\wrong}[1]{\sethlcolor{red!20}\hl{#1}}

\begin{flushleft}
\textbf{GPT-4V Response:}
\begin{tcolorbox}[colback=cyan!10,arc=0mm,auto outer arc]
    \begin{itemize}
    \item \textbf{response 1:}

        \wrong{In a hysteroscopic myomectomy surgery image, the silver surgical instrument you're most likely to see is a hysteroscope. A hysteroscope is a thin, lighted telescope-like device that is inserted into the uterus through the cervix.} It allows the surgeon to see the inside of the uterus and is used to diagnose and treat causes of abnormal bleeding. During a myomectomy, the hysteroscope enables the surgeon to locate and remove fibroids or polyps from within the uterine cavity. It's a key instrument in such procedures, providing a minimally invasive option for surgery.
        
    \item \textbf{response 2:}

        \wrong{In a hysteroscopic myomectomy surgery image, the silver surgical instrument you are referring to is most likely a hysteroscope. A hysteroscope is a thin, lighted telescope-like device that is inserted through the vagina and cervix into the uterus.} It allows the surgeon to view the inside of the uterus and operate with minimal invasion. During a myomectomy, the hysteroscope is used to locate and remove fibroids or polyps from the uterine wall.

    \item \textbf{response 3:}

        \wrong{In a hysteroscopic myomectomy surgery, the most likely silver surgical instrument you might see in an image is a hysteroscope. A hysteroscope is a thin, lighted telescope-like device that is inserted into the uterus through the cervix.} It allows the surgeon to view the inside of the uterus and operate with precision. This instrument is crucial for hysteroscopic myomectomy, a procedure used to remove fibroids (noncancerous growths) from the uterus. The hysteroscope is used both to visualize the fibroids and to guide other instruments that are used to excise or destroy the fibroids.

    \end{itemize}
\vspace{1em}
\end{tcolorbox}
\end{flushleft}

\end{minipage}
}
\end{center}
\vfill
\subsubsection{Case 4: Ablation study Part 2}
\begin{center}
\fbox{%
\begin{minipage}{0.9\textwidth}
\begin{center}
    {\relsize{2}\textbf{CASE 4}}
\end{center}

\begin{center}
    \textbf{Ablation study (No input image, text only)}
\end{center}

\vspace{1em}
\noindent
\begin{tcolorbox}[colback=gray!10,arc=0mm,auto outer arc]
\textbf{Prompt:} Assume there is an image from a myomectomy surgery. Please tell me what the ring-shaped surgical instrument in the picture is and its purpose.
\end{tcolorbox}
\centering
\vspace{1em}
\newcommand{\correct}[1]{\sethlcolor{green!20}\hl{#1}}
\newcommand{\wrong}[1]{\sethlcolor{red!20}\hl{#1}}

\begin{flushleft}
\textbf{GPT-4V Response:}
\begin{tcolorbox}[colback=cyan!10,arc=0mm,auto outer arc]
    \begin{itemize}
    \item \textbf{response 1:}

        In a hypothetical scenario where there's an image from a hysteroscopic myomectomy, a ring-shaped surgical instrument could likely be a \correct{"loop resectoscope."} The loop resectoscope is equipped with an electric wire loop. \correct{Its purpose is to cut and coagulate tissue during the procedure.}

        During a hysteroscopic myomectomy, this instrument is specifically used to shave off the fibroid tissue from the uterus. The electric loop cuts the fibroid, and the electric current ensures that there's minimal bleeding by coagulating the blood vessels as it cuts.
        
    \item \textbf{response 2:}

        A ring-shaped instrument in the context of hysteroscopic myomectomy could possibly be a tenaculum or a \correct{loop electrode}. The loop electrode is often part of a resectoscope, which is commonly used in hysteroscopic myomectomy to remove fibroids from the uterine wall. \correct{The loop can be used to cut and remove the fibroid tissue with the use of electric current, while the hysteroscope allows the surgeon to see inside the uterus during the procedure.}

    \item \textbf{response 3:}

        In the context of a hysteroscopic myomectomy, a ring-shaped surgical instrument is typically referred to as a \correct{"loop resectoscope."} The loop resectoscope is an instrument used in conjunction with a hysteroscope.

        Purpose:

        \correct{The loop resectoscope has an electric wire loop at its end. It is used to cut and coagulate tissue during the procedure. During a hysteroscopic myomectomy, the loop resectoscope is utilized to shave off or remove fibroid tissue from the uterus. The electric current that passes through the wire loop allows for both cutting and coagulation, making the removal of fibroids efficient and minimizing bleeding.}
    \end{itemize}
\vspace{1em}
\end{tcolorbox}
\end{flushleft}

\end{minipage}
}
\end{center}
\vfill

\subsection{Tip 3: Implementing step-by-step guidance in multi-round dialogue allows GPT-4v to handle complex tasks more efficiently by breaking them down into simpler operations.}

\subsubsection{Case 1: Endoscopic mask-polyp matching challenge}
\begin{center}
\fbox{%
\begin{minipage}{0.9\textwidth}
\begin{center}
    {\relsize{2}\textbf{CASE 1}}
\end{center}
\centering
\vspace{1em}
\begin{figure}[H]
  \centering
  \includegraphics[width=0.9\linewidth]{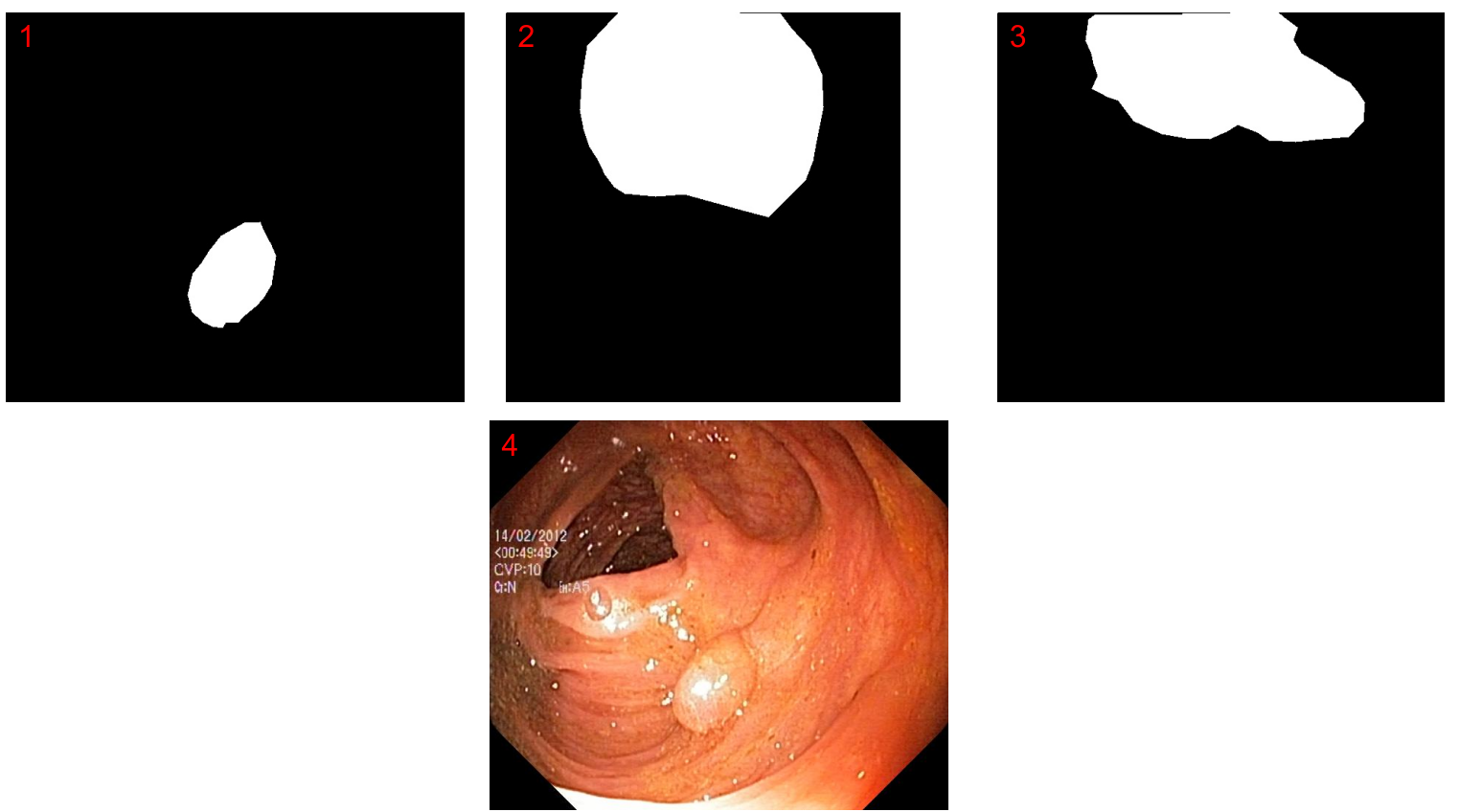}
  \label{C3_CASE1}
\end{figure}

\vspace{1em}
\begin{flushleft}
\begin{itemize}

     \item \textbf{Picture modal:} Endoscopic picture
     \item \textbf{Tasks:} polyp segmentation (Mask-polyp matching)
     \item \textbf{Reference Answer:} 
            \begin{itemize}
                \item The mask corresponding to Pic.4 is Pic.1 
            \end{itemize}

\end{itemize}    
\vspace{1em}
\end{flushleft}

\end{minipage}
}
\end{center}
\vfill
\begin{center}
\fbox{%
\begin{minipage}{0.9\textwidth}
\begin{center}
    {\relsize{2}\textbf{CASE 1}}
\end{center}

\begin{center}
    \textbf{Control group}
\end{center}

\vspace{1em}
\noindent
\begin{tcolorbox}[colback=gray!10,arc=0mm,auto outer arc]
\textbf{Prompt:} Please tell me which mask at the top corresponds to the polyp below.
\end{tcolorbox}
\centering
\vspace{1em}
\newcommand{\correct}[1]{\sethlcolor{green!20}\hl{#1}}
\newcommand{\wrong}[1]{\sethlcolor{red!20}\hl{#1}}

\begin{flushleft}
\textbf{GPT-4V Response:}
\begin{tcolorbox}[colback=cyan!10,arc=0mm,auto outer arc]
    \begin{itemize}
    \item \textbf{response 1:}
        \wrong{The polyp in image 4 appears to match the mask in image 2.} The reason is that the shape and overall silhouette of the polyp in image 4 closely resemble the white mask region shown in image 2. The other masks in images 1 and 3 do not have a similar configuration or shape to that of the polyp in image 4.

    \item \textbf{response 2:}
        \wrong{The polyp in image 4 appears to match the mask in image 2.} The reasoning behind this assessment is based on the shape and size of the polyp in relation to the masks. The mask in image 2 seems to have a shape and contour that closely resembles the visual boundaries of the polyp as seen in image 4. The other masks (1 and 3) either appear too small or have a different contour that doesn't match as closely.
    \item \textbf{response 3:}
        \wrong{The polyp in image 4 appears to match the mask in image 2.} The reason for this is based on the spatial configuration and the overall shape. The mask in image 2 has a rounded contour that is similar to the form of the polyp in image 4. The other masks (1 and 3) don't seem to have the same configuration as the polyp's visible structure.
        
    \end{itemize}
\vspace{1em}
\end{tcolorbox}
\vspace{1em}
\end{flushleft}

\end{minipage}
}
\end{center}

\begin{center}
\fbox{%
\begin{minipage}{0.9\textwidth}
\begin{center}
    {\relsize{2}\textbf{CASE 1}}
\end{center}

\begin{center}
    \textbf{Step-by-step guidance (single round)}
\end{center}

\vspace{1em}
\noindent
\begin{tcolorbox}[colback=gray!10,arc=0mm,auto outer arc]
\textbf{Prompt 1:} Picture 123 is the mask template for polyps, and Picture 4 is an image of a colonoscopy, if I divide each image into nine equal parts, which block would contain the polyp in Picture 4  

If I divide each image into nine equal parts, which block would contain the polyp in
Picture 1 2 and 3 

Which mask would correspond to the Picture 4 

\end{tcolorbox}
\centering
\vspace{1em}
\newcommand{\correct}[1]{\sethlcolor{green!20}\hl{#1}}
\newcommand{\wrong}[1]{\sethlcolor{red!20}\hl{#1}}

\begin{flushleft}
\textbf{GPT-4V Response:}
\begin{tcolorbox}[colback=cyan!10,arc=0mm,auto outer arc]
    \begin{itemize}
    \item \textbf{response 1:}
        For Picture 4:
        When the image is divided into nine equal parts, the polyp in Picture 4 appears to be located in the bottom right block.
        
        For Pictures 1, 2, and 3:
        
        The polyp in Picture 1 is located in the bottom center block.
        The polyp in Picture 2 is located in the top center block.
        The polyp in Picture 3 is located in the top right block.
        As for the correspondence between the mask templates and Picture 4, the mask in \wrong{Picture 3 appears to best match the shape and location of the polyp in Picture 4.}

    \item \textbf{response 2:}
        Picture 4 (Colonoscopy image):
        If we were to divide the image into nine equal parts, the polyp is located in the bottom middle block.
        
        Picture 1:
        Upon dividing the image into nine equal parts, the white shape (which represents the polyp) is located in the center block.
        
        Picture 2:
        The white shape in this Picture is positioned in the top middle block when divided into nine parts.
        
        Picture 3:
        For this image, when segmented into nine equal blocks, the white shape is situated in the top right block.
        
        Corresponding mask for Picture 4:
        Comparing the position and shape of the polyp in Picture 4 with the templates in Pictures 1, 2, and 3, \wrong{none of the masks from Pictures 1, 2, or 3 precisely match the shape and position of the polyp in Picture 4}. \sethlcolor{yellow!20}\hl{However, the closest in terms of position would be Picture 1 as it's centrally located, but the shapes differ.}

    \item \textbf{response 3:}
        To answer your questions:

        For Picture 4: When dividing the image into nine equal parts, the polyp is located in the center block.
        
        For Picture 1: When dividing the image into nine equal parts, the polyp is located in the bottom center block.
        
        For Picture 2: When dividing the image into nine equal parts, the polyp is located in the top center block.
        
        For Picture 3: When dividing the image into nine equal parts, the polyp is located in the center-right block.
        
        \wrong{The mask that corresponds to Picture 4 is Picture 2 (the top center block).}
        
    \end{itemize}
\vspace{1em}
\end{tcolorbox}
\vspace{1em}
\end{flushleft}

\end{minipage}
}
\end{center}

\begin{center}
\fbox{
\begin{minipage}{0.9\textwidth}
\begin{center}
    {\relsize{2}\textbf{CASE 1}}
\end{center}

\begin{center}
    \textbf{Step-by-step guidance (multi-round)}
\end{center}

\vspace{1em}
\noindent
\begin{tcolorbox}[colback=gray!10,arc=0mm,auto outer arc]
\textbf{Prompt 1:} Picture 123 is the mask template for polyps, and Picture 4 is an image of a colonoscopy, if I divide each image into nine equal parts, which block would contain the polyp in Picture 4

\textbf{Prompt 2:} If I divide each image into nine equal parts, which block would contain the polyp in Pictures 1 2, and 3

\textbf{Prompt 3:} Which mask would correspond to Picture 4
\end{tcolorbox}
\centering
\vspace{1em}
\newcommand{\correct}[1]{\sethlcolor{green!20}\hl{#1}}
\newcommand{\wrong}[1]{\sethlcolor{red!20}\hl{#1}}

\begin{flushleft}
\textbf{GPT-4V Response:}
\begin{tcolorbox}[colback=cyan!10,arc=0mm,auto outer arc]
    \begin{itemize}
    \item \textbf{response 1:}
        \begin{itemize}
            \item \textbf{Answering prompt 1:} 
            To identify the location of the polyp in Picture 4 based on the mask template, let's divide each image into nine equal parts (3x3 grid) and see where the corresponding pattern from the template appears in Picture 4.

            Upon examining the images:
            
            Pictures 1-3 serve as a mask template for polyps.
            Picture 4 is an image from a colonoscopy.
            Based on the mask template provided in Pictures 1-3 and the colonoscopy image in Picture 4, the polyp in Picture 4 would be located in the bottom center block (or the 8th block if numbering from top left to bottom right in a 3x3 grid).
            \item \textbf{Answering prompt 2:}
            Let's determine the location of the polyp in Pictures 1, 2, and 3 by dividing each image into nine equal parts (3x3 grid):
            
            Picture 1: The polyp is located in the bottom center block (or the 8th block if numbering from top left to bottom right in a 3x3 grid).
            
            Picture 2: The polyp is located in the top center block (or the 2nd block if numbering from top left to bottom right in a 3x3 grid).
            
            Picture 3: The polyp is located in the center-right block (or the 6th block if numbering from top left to bottom right in a 3x3 grid).
            \item \textbf{Answering prompt 3:}
            Upon comparing the mask templates in Pictures 1, 2, and 3 with the colonoscopy image in Picture 4, the mask in \correct{Picture 1 appears to correspond most closely to the location and shape of the polyp in Picture 4.}
        \end{itemize}

    \item \textbf{response 2:}
        \begin{itemize}
            \item \textbf{Answering prompt 1:} 
            Based on the image you provided, Picture 4 appears to be a colonoscopy image with a polyp.
            
            If we divide Picture 4 into nine equal parts, the polyp is primarily located in the central block. So, the correct answer is the center block
            \item \textbf{Answering prompt 2:}
            Given the images:
            
            In Picture 1, the white shape (presumed to be the polyp) is located in the bottom center block.
            In Picture 2, the white shape is located in the top center block.
            In Picture 3, the white shape extends across the top right and center-right blocks.
            So, for Pictures 1, 2, and 3, the blocks containing the polyps are the bottom center, top center, and top right/center-right respectively.
            \item \textbf{Answering prompt 3:}
            Comparing the mask templates in Pictures 1, 2, and 3 to the colonoscopy image in Picture 4:
            
            \correct{Picture 4 appears to most closely correspond with the mask in Picture 1. The polyp shape and location in Picture 1 seem to be the best match for the polyp observed in Picture 4 when compared to the other masks.}
        \end{itemize}        

    \item \textbf{response 3:}
        \begin{itemize}
            \item \textbf{Answering prompt 1:} 
            The polyp in Picture 4 is located in the center block.
            \item \textbf{Answering prompt 2:}
            For Pictures 1, 2, and 3:
            Picture 1: The polyp is located in the center block.
            Picture 2: The polyp is located in the top-center block.
            Picture 3: The polyp is located in the top-right block. respectively.
            \item \textbf{Answering prompt 3:}
            \correct{The mask in Picture 1 would correspond to Picture 4.}
        \end{itemize} 
        
    \end{itemize}
\vspace{1em}
\end{tcolorbox}
\vspace{1em}
\end{flushleft}
\end{minipage}
}
\end{center}

\subsubsection{Case 2: CT mask-organs matching challenge}
\begin{center}
\fbox{%
\begin{minipage}{0.9\textwidth}
\begin{center}
    {\relsize{2}\textbf{CASE 2}}
\end{center}
\centering
\vspace{1em}
\begin{figure}[H]
  \centering
  \includegraphics[width=0.9\linewidth]{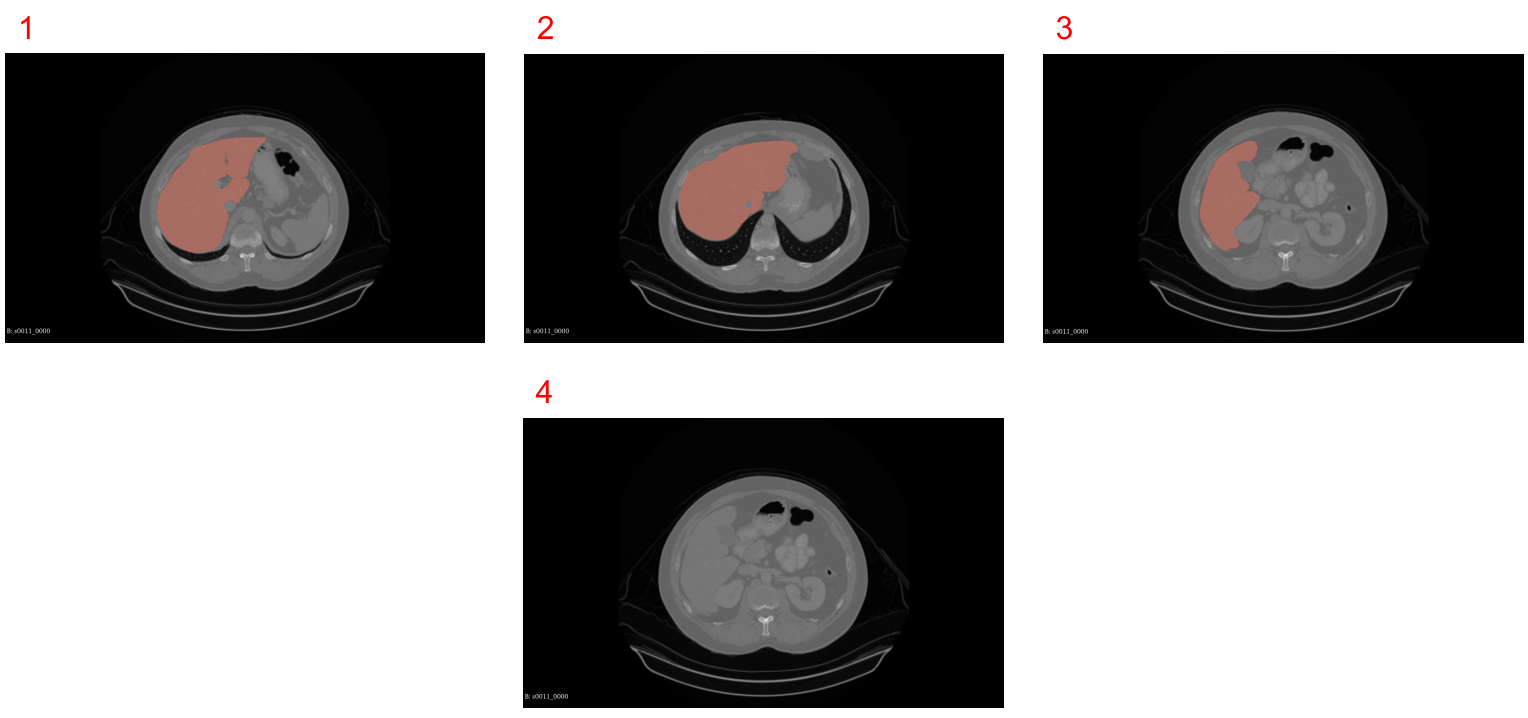}
  \label{C3_CASE2}
\end{figure}

\vspace{1em}
\begin{flushleft}
\begin{itemize}

     \item \textbf{Picture modal:} CT
     \item \textbf{Tasks:} organ segmentation
     \item \textbf{Reference Answer:} 
            \begin{itemize}
                \item The mask corresponding to Pic.4 is Pic.3 
            \end{itemize}

\end{itemize}    
\vspace{1em}
\end{flushleft}

\end{minipage}
}
\end{center}
\vfill
\begin{center}
\fbox{%
\begin{minipage}{0.9\textwidth}
\begin{center}
    {\relsize{2}\textbf{CASE 2}}
\end{center}

\begin{center}
    \textbf{Control Group}
\end{center}

\vspace{1em}
\noindent
\begin{tcolorbox}[colback=gray!10,arc=0mm,auto outer arc]
\textbf{Prompt:} Please tell me which mask at the top corresponds to the liver below, consider the red part only.
\end{tcolorbox}
\centering
\vspace{1em}
\newcommand{\correct}[1]{\sethlcolor{green!20}\hl{#1}}
\newcommand{\wrong}[1]{\sethlcolor{red!20}\hl{#1}}

\begin{flushleft}
\textbf{GPT-4V Response:}
\begin{tcolorbox}[colback=cyan!10,arc=0mm,auto outer arc]
    \begin{itemize}
    \item \textbf{response 1:}
    
        Upon examining the CT images provided, it is evident that the red mask in Image 1 corresponds to the liver's anatomical location. This mask covers the right upper quadrant of the abdomen, which aligns with the liver's predominant placement in the human body. The red part follows the expected shape and region of the liver, extending slightly towards the left upper quadrant, which is consistent with the liver's anatomy. The other masks either cover additional areas not occupied by the liver (as seen in Image 2) or highlight regions to the left (as seen in Image 3), which do not correspond to the primary location of the liver. \wrong{Therefore, the correct liver mask is the one shown in Image 1.}
        
    \item \textbf{response 2:}

        Based on the anatomy visible in the CT images, the red part of the mask that corresponds to the liver is found in Image 1 at the top. \wrong{The liver is situated predominantly in the right upper quadrant of the abdominal cavity, and the red mask in Image 1 accurately overlays this area, conforming to the expected anatomical location and shape of the liver.} The other images either highlight areas that extend beyond the typical boundaries of the liver or, in the case of Image 3, cover a section on the left side where the liver is not present. Image 4 serves as a reference as it does not have any highlighted area, confirming that the highlighted region in Image 1 is indeed representative of the liver’s location.

    \item \textbf{response 3:}
        Upon examining the provided images, \wrong{it is clear that the mask in Image 1 corresponds to the liver in the images below, considering only the red part.} This conclusion is reached by observing the anatomical position and shape of the red highlighted area, which closely matches the typical location and size of the liver in the right upper quadrant of the abdomen on a transverse CT scan. The liver is the largest organ in the abdominal cavity and has a characteristic shape that is well represented by the mask in Image 1, which covers the right lobe of the liver and extends slightly to the left, consistent with the liver's anatomical position. Images 2 and 3 highlight areas that either extend beyond the typical boundaries of the liver or are situated in an atypical location for liver tissue. Image 4 does not have any highlighted areas to compare.
        
    \end{itemize}
\vspace{1em}
\end{tcolorbox}
\vspace{1em}
\end{flushleft}

\end{minipage}
}
\end{center}
\begin{center}
\fbox{%
\begin{minipage}{0.9\textwidth}
\begin{center}
    {\relsize{2}\textbf{CASE 2}}
\end{center}

\begin{center}
    \textbf{Step-by-step guidance (single round)}
\end{center}

\vspace{1em}
\noindent
\begin{tcolorbox}[colback=gray!10,arc=0mm,auto outer arc]
\textbf{Prompt:} Picture 123 is the mask template for the liver(the red part), and Picture 4 is an image of an abdominal CT section, please tell me the location and size of the liver in Picture 4. 

Please tell me the location and size of the livers in pictures 1 2 3

Which mask would correspond to Picture 4

\end{tcolorbox}
\centering
\vspace{1em}
\newcommand{\correct}[1]{\sethlcolor{green!20}\hl{#1}}
\newcommand{\wrong}[1]{\sethlcolor{red!20}\hl{#1}}

\begin{flushleft}
\textbf{GPT-4V Response:}
\begin{tcolorbox}[colback=cyan!10,arc=0mm,auto outer arc]
    \begin{itemize}
    \item \textbf{response 1:}        
        For picture 4, which does not have a mask, the liver is typically located in the upper right quadrant of the abdominal cavity, just under the diaphragm. In a CT image like the one shown, it would usually appear as a large, well-defined area of soft tissue that is more radiodense (i.e., appears lighter) than the surrounding tissues and organs.

        Based on the common anatomy of the liver, the corresponding location in picture 4 would be the large area on the right side of the image (which would be the patient's left side, as images are typically viewed as if you're facing the person), extending from just below the diaphragm down towards the right kidney.

        The mask from \correct{picture 3 would likely correspond to picture 4 because it shows the liver extending across the midline to the left side of the body}, which is a normal variation due to the size and shape of the liver. The positioning and the apparent size of the liver in mask 3 seems most similar to the unmarked CT image in picture 4, considering the anatomical position and the typical appearance of the liver in such scans.
        
    \item \textbf{response 2:}
        In the provided CT images, the liver is represented by the red-shaded area in pictures 1, 2, and 3. The liver is located in the upper right portion of the abdominal cavity, just below the diaphragm and occupies a large area of the upper quadrant of the abdomen.
         
        \begin{itemize}
            \item \textbf{Picture 1:} In picture 1, the liver is the large red area taking up most of the right side of the image (the patient's left side, as medical images are typically viewed from the perspective of facing the patient).
            
            \item \textbf{Picture 2:} In picture 2, again, the liver is the large red area, which is on the right side of the image 

            \item \textbf{Picture 3:} In picture 3, the liver is shown with two separate red areas, indicating the larger right lobe of the liver and the smaller left lobe.
        \end{itemize}
        
        Based on the position and shape of the liver in pictures 1, 2, and 3, \wrong{the mask that would correspond to picture 4 would likely be similar to the ones in pictures 1 and 2, as these show a complete liver, which is how it would normally appear in an unsegmented CT image.}

    \item \textbf{response 3:}
        In the provided images, the red-shaded areas are indicative of the liver's location in the axial (horizontal) cross-section of the abdominal CT scans. Here's how the liver is positioned in each of the images:
        
        \begin{itemize}
            \item \textbf{Picture 1:} The liver is large and occupies the right upper quadrant of the abdomen. It is the red area that spans across the right side of the picture, just beneath the diaphragm.
            
            \item \textbf{Picture 2:} Similar to picture 1, the liver is shown in red, also occupying the right upper quadrant, extending slightly to the left of the midline of the body.

            \item \textbf{Picture 3:} The liver is again highlighted in red, located in the right upper part of the abdomen, but in this image, the red area extends more towards the left, crossing the midline slightly, which is typical as the liver has a left lobe that extends to the left side.
        \end{itemize}

        For picture 4, which is an unmarked CT scan, the liver would typically be located in the same general area as in pictures 1, 2, and 3, meaning it should be in the upper right quadrant of the abdomen, extending across the midline towards the left upper quadrant beneath the diaphragm. Based on the location in the marked images, \wrong{the corresponding mask for picture 4 would likely be most similar to the mask in picture 1 or 2, since these show a large portion of the liver in its typical position.}

    \end{itemize}
\vspace{1em}
\end{tcolorbox}
\vspace{1em}
\end{flushleft}

\end{minipage}
}
\end{center}

\begin{center}
\fbox{%
\begin{minipage}{0.9\textwidth}
\begin{center}
    {\relsize{2}\textbf{CASE 2}}
\end{center}

\begin{center}
    \textbf{Step-by-step guidance (multi-round)}
\end{center}

\vspace{1em}
\noindent
\begin{tcolorbox}[colback=gray!10,arc=0mm,auto outer arc]
\textbf{Prompt 1:} Picture 123 is the mask template for the liver(the red part), and Picture 4 is an image of an abdominal CT section, please tell me the location and size of the liver in picture 4.(answer in 50 words)

\textbf{Prompt 2:}Please tell me the location and size of the livers in picture 1 2 3 (answer in 50 words)

\textbf{Prompt 3:}Which mask would correspond to Picture 4 (answer in 50 words)

\end{tcolorbox}
\centering
\vspace{1em}
\newcommand{\correct}[1]{\sethlcolor{green!20}\hl{#1}}
\newcommand{\wrong}[1]{\sethlcolor{red!20}\hl{#1}}

\begin{flushleft}
\textbf{GPT-4V Response:}
\begin{tcolorbox}[colback=cyan!10,arc=0mm,auto outer arc]
    \begin{itemize}
    \item \textbf{response 1:}

        \begin{itemize}
            \item \textbf{Answering prompt 1:}  In Picture 4, the liver is located in the upper right quadrant of the abdominal cavity, extending across the midline to the upper left quadrant. It is the large, dark gray area beneath the diaphragm. The size cannot be accurately determined without specific measurements or scale, but it occupies a significant portion of the upper abdominal area.
            
            \item \textbf{Answering prompt 2:} In Pictures 1, 2, and 3, the liver (highlighted in red) is primarily located in the right upper quadrant of the abdominal cavity. The size appears large, spanning across to the left upper quadrant, below the diaphragm. Without a scale, precise size cannot be determined, but it occupies a substantial portion of the abdominal cavity in each image.

            \item \textbf{Answering prompt 3:}  Picture 1's mask seems to cover the largest area, extending quite far to the left, which might be too large for Picture 4.
            Picture 2's mask has a more confined area on the right side, which may not cover all of the liver in Picture 4.
            Picture 3's mask is in between the two in terms of coverage, possibly being a closer match to the liver in Picture 4.
            \correct{Based on this visual assessment, Picture 3's mask might correspond most closely to the liver in Picture 4.}
            
        \end{itemize}

    \item \textbf{response 2:}

        \begin{itemize}
            \item \textbf{Answering prompt 1:}  The liver in Picture 4, based on the mask template provided in Pictures 1-3, is not delineated with a red mask. However, typically, the liver is located in the upper right quadrant of the abdomen in a CT image, extending from just under the diaphragm down towards the right kidney. It is not possible to accurately determine the size of the liver from Picture 4 without the mask, but the liver is usually the largest solid organ in the abdominal cavity
            
            \item \textbf{Answering prompt 2:}  In pictures 1, 2, and 3, the liver (highlighted in red) occupies the upper right quadrant of the abdomen. It extends across the midline to the left upper quadrant. The size appears substantial, filling much of the space below the diaphragm and above the stomach and intestines.
            
            \item \textbf{Answering prompt 3:}  Based on typical anatomical position and the visible structures in Picture 4, \wrong{the liver mask that would correspond most closely appears to be the mask in Picture 1,} considering the liver's location and the visible vasculature branching within the liver tissue.
            
        \end{itemize}

    \item \textbf{response 3:}
    
        \begin{itemize}
            \item \textbf{Answering prompt 1:} In Picture 4, the liver is located in the upper right quadrant of the abdomen, just below the diaphragm. It spans from the right side across the midline towards the left, extending slightly past the stomach. The size of the liver appears to be within normal limits, occupying a significant portion of the upper abdominal cavity.
            
            \item \textbf{Answering prompt 2:}  In Pictures 1-3, the liver, indicated by the red mask, occupies the upper right quadrant of the abdomen, extending across the midline to the upper left quadrant. It varies in size across the images, with Picture 2 showing the largest coverage and Picture 3 showing the least.

            \item \textbf{Answering prompt 3:} Without medical imaging software to analyze the images in detail, it's challenging to accurately determine which mask corresponds to Picture 4. \correct{However, based on visual approximation, Picture 1 appears to have a liver mask that most closely matches the visible liver contours in Picture 4.}
            
        \end{itemize}
    \end{itemize}
\vspace{1em}
\end{tcolorbox}
\vspace{1em}
\end{flushleft}

\end{minipage}
}
\end{center}


\subsubsection{Case 3: MRI mask-organs matching challenge}
\begin{center}
\fbox{%
\begin{minipage}{0.9\textwidth}
\begin{center}
    {\relsize{2}\textbf{CASE 3}}
\end{center}
\centering
\vspace{1em}
\begin{figure}[H]
  \centering
  \includegraphics[width=0.9\linewidth]{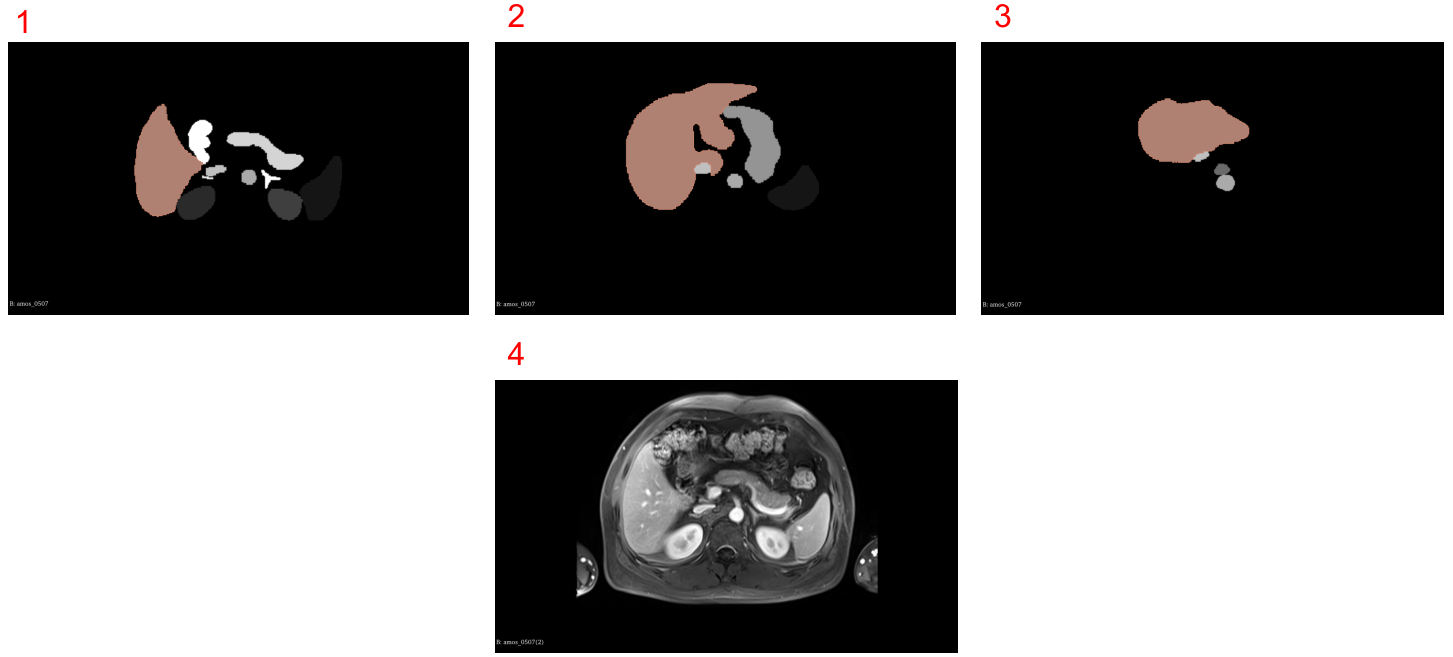}
  \label{C3_CASE3}
\end{figure}

\vspace{1em}
\begin{flushleft}
\begin{itemize}

     \item \textbf{Picture modal:} MRI
     \item \textbf{Tasks:} organ segmentation (mask-organ matching)
     \item \textbf{Reference Answer:} 
            \begin{itemize}
                \item The mask corresponding to Pic.4 is Pic.1 
            \end{itemize}

\end{itemize}    
\vspace{1em}
\end{flushleft}

\end{minipage}
}
\end{center}
\vfill
\begin{center}
\fbox{%
\begin{minipage}{0.9\textwidth}
\begin{center}
    {\relsize{2}\textbf{CASE 3}}
\end{center}

\begin{center}
    \textbf{Control Group}
\end{center}

\vspace{1em}
\noindent
\begin{tcolorbox}[colback=gray!10,arc=0mm,auto outer arc]
\textbf{Prompt:} Please tell me which mask at the top corresponds to the liver below, consider the red part only.
\end{tcolorbox}
\centering
\vspace{1em}
\newcommand{\correct}[1]{\sethlcolor{green!20}\hl{#1}}
\newcommand{\wrong}[1]{\sethlcolor{red!20}\hl{#1}}

\begin{flushleft}
\textbf{GPT-4V Response:}
\begin{tcolorbox}[colback=cyan!10,arc=0mm,auto outer arc]
    \begin{itemize}
    \item \textbf{response 1:}
    
        The images you've provided appear to be masks generated from an MRI scan, with the bottom image showing an actual MRI scan of the abdomen. To determine which mask at the top corresponds to the liver in the actual MRI scan, we need to match the shape and position of the liver in the MRI scan with the red masks above.
        
        The liver is typically located in the upper right quadrant of the abdomen, just under the diaphragm. It has a characteristic shape that is usually wider at the bottom and tapers off near the top.
        
        By comparing the shapes and positions, it looks like the mask in:
        
        \wrong{Image 2 most closely corresponds to the liver's shape and position in the MRI scan.}
        This mask shows a larger mass on the right side, which tapers off towards the left, closely mirroring the liver's typical presentation in an axial MRI slice.
    \item \textbf{response 2:}

        The images you've provided appear to be from a medical imaging study, where the top three images are masks representing different anatomical structures in the liver, and the bottom image seems to be a cross-sectional MRI scan of the abdomen, including the liver.
        
        To determine which mask corresponds to the liver on the MRI scan, one needs to consider the anatomical position, shape, and relative size of the liver in the MRI scan and match it to the provided masks. The liver is typically located in the right upper quadrant of the abdomen.
        
        Based on visual inspection, it seems that \wrong{mask 2 at the top most closely corresponds to the shape and position of the liver in the MRI scan below.} This mask appears to cover the area of the liver without including other adjacent organs or structures, which aligns with the typical anatomical location of the liver as shown in the MRI image.

    \item \textbf{response 3:}
        The images provided show a series of masks corresponding to segments in an MRI scan. To identify which mask corresponds to the liver in the MRI scan, one would look for the shape and location that matches the liver's appearance and anatomical position in the lower image.
        
        In the MRI scan, the liver is typically located in the upper right quadrant of the abdomen, just under the diaphragm. It appears as a large, dark mass on T2-weighted images like the one shown, due to its relatively homogeneous texture compared to other organs and tissues which may appear brighter due to higher water content.
        
        Comparing the shapes and positions in the masks to the liver in the MRI:

        Mask 1 shows a large area that extends across the midline, which is not characteristic of the liver's location.

        Mask 2 shows a large area that is positioned to the right but extends too far to the left, which is not typical for the liver.

        Mask 3 shows a smaller area positioned on the right side only, which could be a closer match to the typical location of the liver.

        Given this information and without more specific details about the orientation and type of MRI sequence, \wrong{Mask 3 seems to be the most likely candidate that corresponds to the liver}, considering only the red part of the mask. However, such assessments are best confirmed by a medical professional with expertise in radiology.
        
    \end{itemize}
\vspace{1em}
\end{tcolorbox}
\vspace{1em}
\end{flushleft}

\end{minipage}
}
\end{center}
\begin{center}
\fbox{%
\begin{minipage}{0.9\textwidth}
\begin{center}
    {\relsize{2}\textbf{CASE 3}}
\end{center}

\begin{center}
    \textbf{Step-by-step guidance (single round)}
\end{center}

\vspace{1em}
\noindent
\begin{tcolorbox}[colback=gray!10,arc=0mm,auto outer arc]
\textbf{Prompt:} Picture 123 is the mask template for the liver(the red part), and Picture 4 is an image of an abdominal MRI section, please tell me the location of the liver in picture 4. 

Please tell me the location of the livers in picture 1 2 3

Which mask would correspond to Picture 4 

\end{tcolorbox}
\centering
\vspace{1em}
\newcommand{\correct}[1]{\sethlcolor{green!20}\hl{#1}}
\newcommand{\wrong}[1]{\sethlcolor{red!20}\hl{#1}}

\begin{flushleft}
\textbf{GPT-4V Response:}
\begin{tcolorbox}[colback=cyan!10,arc=0mm,auto outer arc]
    \begin{itemize}
    \item \textbf{response 1:}
        In picture 4, the liver is the large, darker area that takes up the upper part of the abdomen on the right side (from the perspective of the image, which would be the patient's left side since MRI images are typically mirrored).
        
        Regarding pictures 1, 2, and 3, each of them shows a red area that represents the liver's location as a mask on a black background. Here's how they would match:
        \begin{itemize}
            \item \textbf{Picture 1}  shows a mask where the liver is represented in the correct anatomical position, extending across the right side of the image with segments appearing on the left side.
            
            \item \textbf{Picture 2}  has a similar representation, with the liver's position slightly more centralized but still predominantly on the right side.   

            \item \textbf{Picture 3}  depicts the liver significantly smaller and more to the central and left side, which is less typical for a standard anatomical position.          
        \end{itemize}
        
        For picture 4, the mask that would correspond to the MRI image would be the one that shows the liver in the anatomical position that matches the dark area on the MRI. \correct{This would most likely be the mask from picture 1}, as it covers the area where the liver is typically located and appears to be the correct size and shape to match the liver's appearance in an axial MRI section.

    \item \textbf{response 2:}
        In the provided MRI image, the liver is located in the upper right quadrant of the abdomen. It appears as a prominent, dark area in the scan, situated just below the diaphragm and to the right of where the stomach would be, which is usually less distinct in an MRI. The masks in Pictures 1, 2, and 3 offer abstract representations of the liver's location and shape within the human body, following a color-coded template.
        
        The liver's relative position is consistent across the masks, although the shapes differ due to individual anatomical variations or the specific liver section shown. 
        \begin{itemize}
            \item \textbf{Picture 1}  mask shows a large red area on the right, corresponding to the liver's position from the viewer's perspective. 
            
            \item \textbf{Picture 2}  mask has a similar red area representing the liver. 

            \item \textbf{Picture 3}  presents a smaller red section, suggesting a different or smaller liver portion.
        \end{itemize}        
        
        Determining the most accurate mask for Picture 4 depends on the MRI's orientation and section. The liver is the largest solid organ with a distinctive shape, so the correct mask would closely resemble the large, dark area on the right side of the MRI image. \correct{Picture 1 is likely the most accurate representation, with its extensive red area aligning with the liver's visible section in the MRI.}

    \item \textbf{response 3:}
        The liver in an abdominal MRI, like Picture 4, is located on the upper right from the viewer’s angle, corresponding to the patient's left, and shows up as a large, darker area. 
        
        For Pictures 1, 2, and 3, the red overlay marks the liver differently: to the left, centered with a right lean, and mainly on the right, respectively.
        
        Determining which mask aligns with Picture 4 depends on the MRI's plane—axial, coronal, or sagittal. \correct{Visually, Picture 1's red overlay looks most like the liver in Picture 4}, but definitive matching requires specialized software to consider the liver’s 3D shape and position.
        
    \end{itemize}
\vspace{1em}
\end{tcolorbox}
\vspace{1em}
\end{flushleft}

\end{minipage}
}
\end{center}

\begin{center}
\fbox{%
\begin{minipage}{0.9\textwidth}
\begin{center}
    {\relsize{2}\textbf{CASE 3}}
\end{center}

\begin{center}
    \textbf{Step-by-step guidance (multi-round)}
\end{center}

\vspace{1em}
\noindent
\begin{tcolorbox}[colback=gray!10,arc=0mm,auto outer arc]
\textbf{Prompt 1:} Picture 123 is the mask template for the liver(the red part), and Picture 4 is an image of an abdominal MRI section, please tell me the location of the liver in picture 4.(answer in 50 words)

\textbf{Prompt 2:}Please tell me the location of the livers in picture 1 2 3  (answer in 50 words)

\textbf{Prompt 3:}Which mask would correspond to Picture 4  (answer in 50 words)

\end{tcolorbox}
\centering
\vspace{1em}
\newcommand{\correct}[1]{\sethlcolor{green!20}\hl{#1}}
\newcommand{\wrong}[1]{\sethlcolor{red!20}\hl{#1}}

\begin{flushleft}
\textbf{GPT-4V Response:}
\begin{tcolorbox}[colback=cyan!10,arc=0mm,auto outer arc]
    \begin{itemize}
    \item \textbf{response 1:}

        \begin{itemize}
            \item \textbf{Answering prompt 1:}  In picture 4, the liver is located in the upper right quadrant of the abdomen, occupying most of the space under the diaphragm. It appears as the large, dark area on the right side of the image (from the viewer's perspective), extending towards the left side across the midline.
            
            \item \textbf{Answering prompt 2:} In pictures 1, 2, and 3, the liver is represented by the red-colored mask. It is positioned on the right side of the body (left side from the viewer's perspective) and extends toward the center, varying slightly in each image due to the perspective and section of the scan

            \item \textbf{Answering prompt 3:}  \correct{The mask that corresponds to Picture 4 would be Picture 1.} This mask shows the liver covering a large area across the upper right quadrant, which matches the liver's position and size in the MRI image of Picture 4, extending towards the central and left portions of the abdominal cavity.
            
        \end{itemize}

    \item \textbf{response 2:}

        \begin{itemize}
            \item \textbf{Answering prompt 1:}  In Picture 4, the liver is located in the upper right quadrant of the abdominal cavity, directly under the diaphragm. In the MRI image, it is the large, dark area at the top, extending from the right side of the image towards the center, and is the brightest and most voluminous organ visible.
            
            \item \textbf{Answering prompt 2:}  In pictures 1, 2, and 3, the liver is represented by the red area. It is the largest shape in each picture, typically located towards the upper part of the image, indicating its position in the upper right quadrant of the body's abdominal cavity in a real-life orientation.
            
            \item \textbf{Answering prompt 3:}  \correct{The mask that corresponds to Picture 4 is the one in Picture 1.} This mask closely matches the shape and position of the liver in the MRI image, with the liver's broad, right-sided location in the abdominal cavity being accurately represented by the large red area in the mask.
            
        \end{itemize}

    \item \textbf{response 3:}
    
        \begin{itemize}
            \item \textbf{Answering prompt 1:}  In picture 4, the liver is the large organ located in the upper right quadrant of the abdominal cavity. It is the dark grayish-brown structure in the MRI image, extending across the midline to the left upper quadrant but predominantly situated on the right side of the image.
            
            \item \textbf{Answering prompt 2:}  In pictures 1, 2, and 3, the liver is represented by the reddish-brown area. In all images, it is located toward the top portion of the mask template, with the position slightly varying in each: centrally to the left in picture 1, more to the left in picture 2, and predominantly on the left in picture 3.

            \item \textbf{Answering prompt 3:} \correct{The mask that corresponds to Picture 4 is the one in Picture 1.} This mask shows a liver representation on the right side of the body (left side of the image from our perspective), which matches the typical location of the liver as seen in the MRI image in Picture 4.
            
        \end{itemize}
    \end{itemize}
\vspace{1em}
\end{tcolorbox}
\vspace{1em}
\end{flushleft}

\end{minipage}
}
\end{center}

\subsection{Tip 4: Don't expose your target at the very beginning when you start multi-round dialog}

\subsubsection{Case 1: Endoscopic mask-polyp matching challenge}

\begin{center}
\fbox{%
\begin{minipage}{0.9\textwidth}
\begin{center}
    {\relsize{2}\textbf{CASE 1}}
\end{center}

\begin{center}
    \textbf{Same as CASE 1 in conclusion 3}
\end{center}

\centering
\vspace{1em}
\begin{figure}[H]
  \centering
  \includegraphics[width=0.9\linewidth]{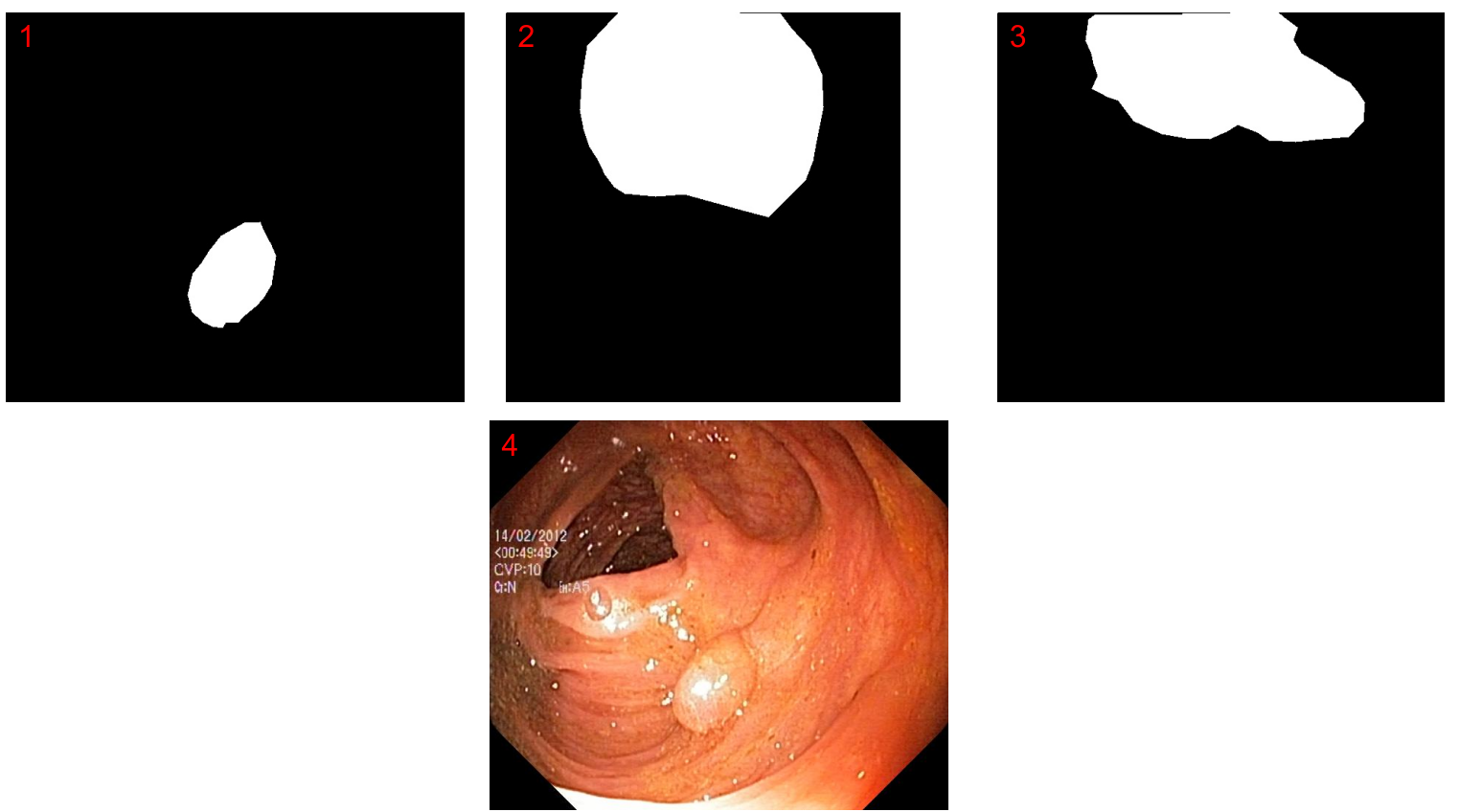}
  \label{C4_CASE1}
\end{figure}

\vspace{1em}
\begin{flushleft}
\begin{itemize}

     \item \textbf{Picture modal:} Endoscopic picture
     \item \textbf{Tasks:} polyp segmentation (Mask-polyp matching)
     \item \textbf{Reference Answer:} 
            \begin{itemize}
                \item the mask corresponding to Pic.4 is Pic.1
            \end{itemize}

\end{itemize}    
\vspace{1em}
\end{flushleft}

\end{minipage}
}
\end{center}
\vfill
\begin{center}
\fbox{%
\begin{minipage}{0.9\textwidth}
\begin{center}
    {\relsize{2}\textbf{CASE 1}}
\end{center}

\begin{center}
    \textbf{Target exposed (marked by orange)}
\end{center}

\vspace{1em}
\noindent
\begin{tcolorbox}[colback=gray!10,arc=0mm,auto outer arc]
\textbf{Prompt 1:} \sethlcolor{orange!20}\hl{Following the procedure to find which Picture in 1-3 is the mask of 4.} Picture 123 is the mask template for polyps, and Picture 4 is an image of a colonoscopy, if I divide each image into nine equal parts, which block would contain the polyp in Picture 4  

\textbf{Prompt 2:}If I divide each image into nine equal parts, which block would contain the polyp in
Picture 1 2 and 3 

\textbf{Prompt 3:}Which mask would correspond to the Picture 4 
\end{tcolorbox}
\centering
\vspace{1em}
\newcommand{\correct}[1]{\sethlcolor{green!20}\hl{#1}}
\newcommand{\wrong}[1]{\sethlcolor{red!20}\hl{#1}}

\begin{flushleft}
\textbf{GPT-4V Response:}
\begin{tcolorbox}[colback=cyan!10,arc=0mm,auto outer arc]
    \begin{itemize}
    \item \textbf{response 1:}
        \begin{itemize}
            \item \textbf{Answering prompt 1:} 
            Based on the mask templates in Pictures 1-3 and the colonoscopy image in Picture 4, the mask that matches the location of the polyp in \wrong{Picture 4 is Picture 2.} The polyp in Picture 4 is located in the central block when the image is divided into nine equal parts.
            \item \textbf{Answering prompt 2:}
            Dividing each image into nine equal parts, the blocks containing the polyp would be:
            
            Picture 1: Bottom center block
            Picture 2: Top center block
            Picture 3: Right center block.
            \item \textbf{Answering prompt 3:}
            \wrong{Based on the provided images, Picture 2's mask (top center block) corresponds to the location of the polyp in Picture 4.}
        \end{itemize}          
    
    \item \textbf{response 2:}
        \begin{itemize}
            \item \textbf{Answering prompt 1:} 
            Based on the mask templates provided in Pictures 1-3 and the colonoscopy image in Picture 4, \wrong{Picture 2 appears to be the mask corresponding to the polyp's location in Picture 4.} The polyp in Picture 4 is located in the bottom-right block when divided into nine equal parts.
            \item \textbf{Answering prompt 2:}
            When dividing each image into nine equal parts:
            
            Picture 1: The polyp mask is located in the center block.
            Picture 2: The polyp mask is located in the top-middle block.
            Picture 3: The polyp mask is located in the top-right block.
            \item \textbf{Answering prompt 3:}
            \wrong{Based on the colonoscopy image in Picture 4, Picture 2 appears to be the mask corresponding to the polyp's location in Picture 4.}
        \end{itemize}  

    \item \textbf{response 3:}
        \begin{itemize}
            \item \textbf{Answering prompt 1:} 
            \wrong{Based on the provided images, Picture 2 is the mask template that matches the polyp location in Picture 4.} The polyp in Picture 4 is located in the bottom middle block when divided into nine equal parts.
            \item \textbf{Answering prompt 2:}
            When dividing each image into nine equal parts (3x3 grid):
            
            Picture 1: The polyp mask is in the bottom middle block.
            Picture 2: The polyp mask is in the top middle block.
            Picture 3: The polyp mask is in the top right block.
            \item \textbf{Answering prompt 3:}
            \correct{Based on the provided images, Picture 1 is the mask template that corresponds to the polyp location in Picture 4.}
        \end{itemize}  
        
    \end{itemize}
\vspace{1em}
\end{tcolorbox}
\vspace{1em}
\end{flushleft}

\end{minipage}
}
\end{center}
\subsubsection{Case 2: CT mask-organs matching challenge}
\begin{center}
\fbox{%
\begin{minipage}{0.9\textwidth}
\begin{center}
    {\relsize{2}\textbf{CASE 2}}
\end{center}
\centering
\vspace{1em}
\begin{figure}[H]
  \centering
  \includegraphics[width=0.9\linewidth]{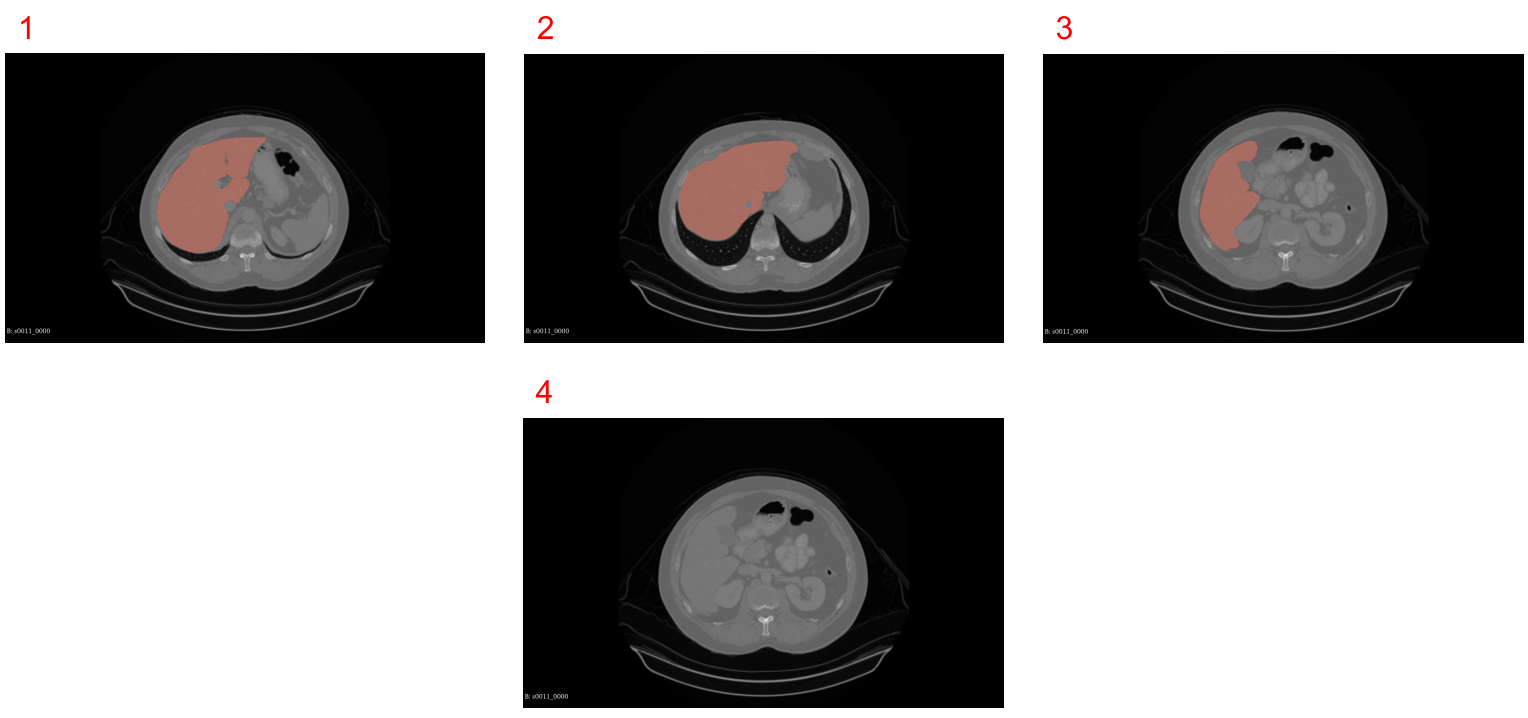}
  \label{C3_CASE2}
\end{figure}

\vspace{1em}
\begin{flushleft}
\begin{itemize}

     \item \textbf{Picture modal:} CT
     \item \textbf{Tasks:} organ segmentation
     \item \textbf{Reference Answer:} 
            \begin{itemize}
                \item The mask corresponding to Pic.4 is Pic.3 
            \end{itemize}

\end{itemize}    
\vspace{1em}
\end{flushleft}

\end{minipage}
}
\end{center}
\vfill
\begin{center}
\fbox{%
\begin{minipage}{0.9\textwidth}
\begin{center}
    {\relsize{2}\textbf{CASE 2}}
\end{center}

\begin{center}
    \textbf{Target exposed (marked by orange)}
\end{center}

\vspace{1em}
\noindent
\begin{tcolorbox}[colback=gray!10,arc=0mm,auto outer arc]
\textbf{Prompt 1:} \sethlcolor{orange!20}\hl{Following the procedure to find which Picture in 1-3 is the mask of 4.} Picture 123 is the mask template for livers, and Picture 4 is an image of a CT section, please tell me the location and size of the liver in Picture 4  (answer in 50 words)

\textbf{Prompt 2:} Please tell me the location and size of the livers in pictures 1 2 3. (answer in 50 words)

\textbf{Prompt 3:}Which mask would correspond to the Picture 4  (answer in 50 words)
\end{tcolorbox}
\centering
\vspace{1em}
\newcommand{\correct}[1]{\sethlcolor{green!20}\hl{#1}}
\newcommand{\wrong}[1]{\sethlcolor{red!20}\hl{#1}}

\begin{flushleft}
\textbf{GPT-4V Response:}
\begin{tcolorbox}[colback=cyan!10,arc=0mm,auto outer arc]
    \begin{itemize}
    \item \textbf{response 1:}
        \begin{itemize}
            \item \textbf{Answering prompt 1:} 
            \wrong{The liver in Picture 4 is likely to correspond with the mask shown in Picture 1.} The liver is positioned predominantly on the right side of the abdomen, extends across the midline towards the left, and occupies a large space in the upper abdomen, as indicated by the shape and location of the red area in Picture 1.
            \item \textbf{Answering prompt 2:}
            In pictures 1-3, the red areas representing the liver are located in the upper abdomen, with the \wrong{largest coverage in picture 3, suggesting a lower axial slice.} Picture 2 shows a smaller section, likely higher near the liver's dome, and picture 1 indicates a mid-level section with substantial bilateral extension.
            \item \textbf{Answering prompt 3:}
            Based on the typical anatomical position of the liver in axial CT images and without the ability to overlay the images directly, \wrong{Picture 1 appears to be the corresponding mask for Picture 4 due to the large size and the right-sided dominance of the red area, which matches the usual liver location.}
        \end{itemize}          
    
    \item \textbf{response 2:}
        \begin{itemize}
            \item \textbf{Answering prompt 1:} 
            \wrong{The liver in Picture 4 appears to be most closely matching the mask in Picture 1}, suggesting it occupies the upper right quadrant of the abdomen. It is not possible to provide an accurate size measurement without specific tools; however, visually, it takes up a significant portion of the upper abdominal area.
            \item \textbf{Answering prompt 2:}
            In Pictures 1, 2, and 3, the liver (red part) is predominantly located in the right upper quadrant of the abdomen, extending slightly to the left upper quadrant. The size varies in each, covering a large portion of the upper abdomen, with Picture 2 showing the largest coverage.
            \item \textbf{Answering prompt 3:}
            \wrong{The mask in Picture 1 appears to correspond to Picture 4}, as the shape and position of the red area most closely match the anatomical location and size of the liver as visualized on the CT image in Picture 4.
        \end{itemize}  

    \item \textbf{response 3:}
        \begin{itemize}
            \item \textbf{Answering prompt 1:} 
            The liver in Picture 4 is located in the upper right quadrant of the abdominal cavity, extending across the midline towards the left side. \wrong{It appears to cover a large portion of the upper abdomen, under the diaphragm, similar to the mask shown in Picture 2.} The liver's size looks to be consistent with the standard anatomical proportions in comparison to the rest of the organs present in the CT image.
            \item \textbf{Answering prompt 2:}
            In Pictures 1, 2, and 3, the liver is highlighted by a red mask. In Picture 1, the liver is large, extending across the right upper quadrant to the midline. Picture 2 shows a large liver, dominating the right side. Picture 3's liver is also large, crossing the midline slightly
            \item \textbf{Answering prompt 3:}
            \wrong{The mask that corresponds to Picture 4 appears to be the one in Picture 2.} This is because the highlighted area in Picture 2 closely matches the size and shape of the liver as seen in Picture 4, covering the right upper quadrant and extending towards the midline.
        \end{itemize}  
        
    \end{itemize}
\vspace{1em}
\end{tcolorbox}
\vspace{1em}
\end{flushleft}

\end{minipage}
}
\end{center}

\subsubsection{Case 3: MRI mask-organs matching challenge}
\begin{center}
\fbox{%
\begin{minipage}{0.9\textwidth}
\begin{center}
    {\relsize{2}\textbf{CASE 3}}
\end{center}
\centering
\vspace{1em}
\begin{figure}[H]
  \centering
  \includegraphics[width=0.9\linewidth]{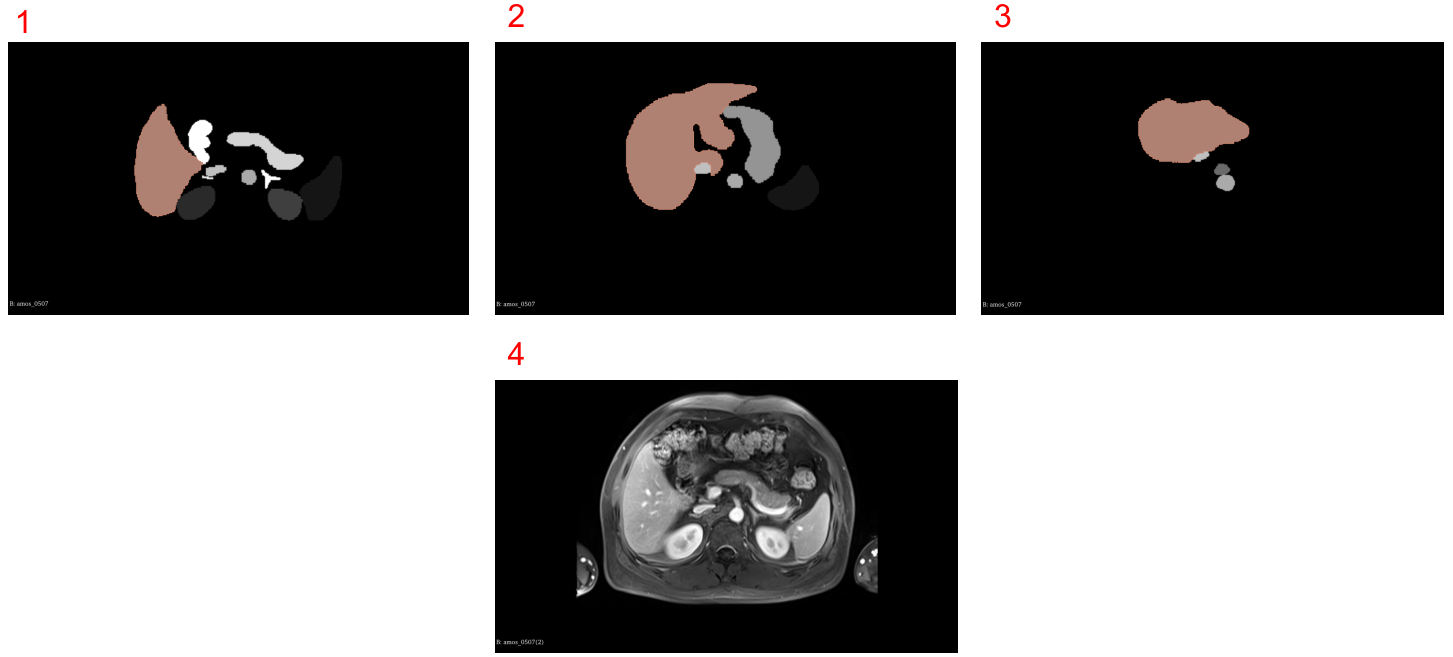}
  \label{C3_CASE3}
\end{figure}

\vspace{1em}
\begin{flushleft}
\begin{itemize}

     \item \textbf{Picture modal:} MRI
     \item \textbf{Tasks:} organ segmentation (mask-organ matching)
     \item \textbf{Reference Answer:} 
            \begin{itemize}
                \item The mask corresponding to Pic.4 is Pic.1 
            \end{itemize}

\end{itemize}    
\vspace{1em}
\end{flushleft}

\end{minipage}
}
\end{center}
\vfill
\begin{center}
\fbox{%
\begin{minipage}{0.9\textwidth}
\begin{center}
    {\relsize{2}\textbf{CASE 3}}
\end{center}

\begin{center}
    \textbf{Target exposed (marked by orange)}
\end{center}

\vspace{1em}
\noindent
\begin{tcolorbox}[colback=gray!10,arc=0mm,auto outer arc]
\textbf{Prompt 1:} \sethlcolor{orange!20}\hl{Following the procedure to find which Picture in 1-3 is the mask of 4.} Picture 123 is the mask template for livers, and Picture 4 is an image of an MRI section, please tell me the location and size of the liver in Picture 4  (answer in 50 words)

\textbf{Prompt 2:} Please tell me the location and size of the livers in pictures 1 2 3. (answer in 50 words)

\textbf{Prompt 3:}Which mask would correspond to the Picture 4  (answer in 50 words)
\end{tcolorbox}
\centering
\vspace{1em}
\newcommand{\correct}[1]{\sethlcolor{green!20}\hl{#1}}
\newcommand{\wrong}[1]{\sethlcolor{red!20}\hl{#1}}

\begin{flushleft}
\textbf{GPT-4V Response:}
\begin{tcolorbox}[colback=cyan!10,arc=0mm,auto outer arc]
    \begin{itemize}
    \item \textbf{response 1:}
        \begin{itemize}
            \item \textbf{Answering prompt 1:} 
            In Picture 4, the liver is the large organ located in the upper right quadrant of the abdomen, spanning almost the entire width of the image. It appears to be the largest organ visible. \wrong{The mask corresponding to the liver in Pictures 1-3 is likely to be Picture 2}, due to its size and shape which closely matches the liver's position and form in Picture 4.
            \item \textbf{Answering prompt 2:}
            In pictures 1, 2, and 3, the liver representations are abstracted as shaded areas against a black background. Picture 1 shows a large liver mask on the left side, Picture 2 shows a slightly smaller mask towards the center, and Picture 3 displays the smallest mask on the right side.
            \item \textbf{Answering prompt 3:}
            \wrong{The mask that corresponds to Picture 4 is the one in Picture 2.} It most accurately represents the size and shape of the liver in the MRI image, with its larger, more central position compared to the others, and with the characteristic lobe structure on the right side.
        \end{itemize}          
    
    \item \textbf{response 2:}
        \begin{itemize}
            \item \textbf{Answering prompt 1:} 
            The liver in Picture 4 is centrally located, extending across the midline of the body and predominantly on the right side. It is the large, dark area on the MRI, taking up a significant portion of the upper abdomen, corresponding most closely to the mask in Picture 2, indicating a typical size and location for a liver.
            \item \textbf{Answering prompt 2:}
            In the mask templates (Pictures 1-3), the liver representations are as follows:
            
            Large, right-sided, extending from the upper to the lower mid abdomen.

            Large, predominantly right-sided but with a lobe crossing midline to the left.

            Smaller, strictly right-sided, located in the upper abdomen.
            \item \textbf{Answering prompt 3:}
            \wrong{The mask in Picture 2 corresponds to Picture 4.} It shows a large liver extending across the midline with a realistic representation of both the left and right lobes, matching the anatomical position and size of the liver as seen in the MRI image.
        \end{itemize}  

    \item \textbf{response 3:}
        \begin{itemize}
            \item \textbf{Answering prompt 1:} 
            In Picture 4, which is an MRI cross-sectional image, the liver is represented by the large, dark area on the right side of the image (patient's left side). It occupies a significant portion of the abdominal cavity, extending from the upper right quadrant towards the central upper abdomen. \wrong{The corresponding mask that best matches the liver's shape and position appears to be Picture 2. }
            \item \textbf{Answering prompt 2:}
            In pictures 1, 2, and 3, the liver masks are depicted in different shades of brown. Picture 1 shows a large liver mask occupying the majority of the image space, extending centrally. Picture 2 shows a more compact liver shape towards the top. Picture 3 displays the smallest liver mask, positioned towards the upper right
            \item \textbf{Answering prompt 3:}
            \wrong{The mask that corresponds to Picture 4 is the one in Picture 2.} It most closely matches the size, shape, and orientation of the liver as seen in the MRI image, covering the right section of the abdomen and wrapping around the central axis of the body.
        \end{itemize}  
        
    \end{itemize}
\vspace{1em}
\end{tcolorbox}
\vspace{1em}
\end{flushleft}

\end{minipage}
}
\end{center}


\subsection{Tip 5: Describing appearances or characteristics will greatly enhance the performance.}

\subsubsection{Case 1: Endoscopic image annotation recognition}
\begin{center}
\fbox{%
\begin{minipage}{0.9\textwidth}
\begin{center}
    {\relsize{2}\textbf{CASE 1}}
\end{center}
\centering
\vspace{1em}
\begin{figure}[H]
  \centering
  \includegraphics[width=0.9\linewidth]{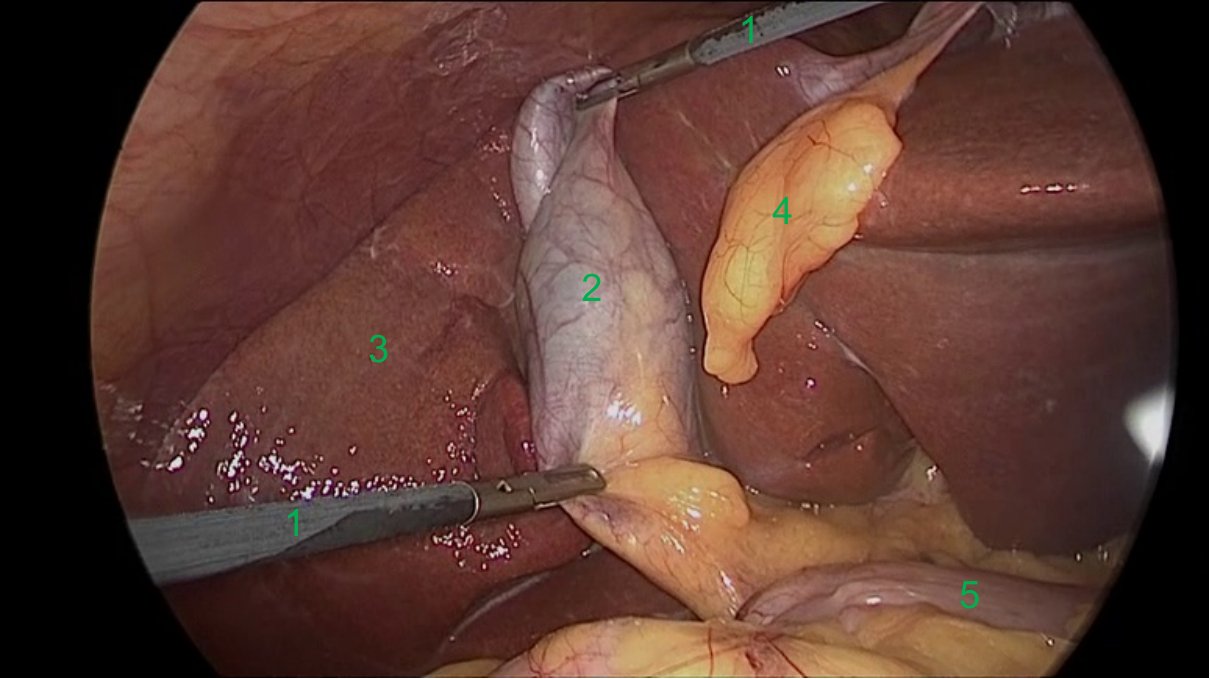}
  \label{C5_CASE1}
\end{figure}

\vspace{1em}
\begin{flushleft}
\begin{itemize}

     \item \textbf{Picture modal:} Endoscopic picture
     \item \textbf{Tasks:} Laparoscopic cholecystectomy
     \item \textbf{Reference Answer:} 
            \begin{itemize}
                \item \textbf{1:} Grasper.
                \item \textbf{2:} Gallbladder.
                \item \textbf{3:} Liver.
                \item \textbf{4:} Fats.
                \item \textbf{5:} Intestine.
            \end{itemize}

\end{itemize}     
\vspace{1em}
\end{flushleft}

\end{minipage}
}
\end{center}
\vfill
\begin{center}
\fbox{%
\begin{minipage}{0.9\textwidth}
\begin{center}
    {\relsize{2}\textbf{CASE 1}}
\end{center}

\begin{center}
    \textbf{Control group}
\end{center}

\vspace{1em}
\noindent
\begin{tcolorbox}[colback=gray!10,arc=0mm,auto outer arc]
\textbf{Prompt:} This is a picture of cholecystectomy In the picture, there are 5 marked numbers. Please tell me which tissues or instruments correspond to the marked numbers in the picture.
\end{tcolorbox}
\centering
\vspace{1em}
\newcommand{\correct}[1]{\sethlcolor{green!20}\hl{#1}}
\newcommand{\wrong}[1]{\sethlcolor{red!20}\hl{#1}}

\begin{flushleft}
\textbf{GPT-4V Response:}
\begin{tcolorbox}[colback=cyan!10,arc=0mm,auto outer arc]
    \begin{itemize}
    \item \textbf{response 1:}
        \begin{itemize}
            \item \textbf{1:} This appears to be a laparoscopic instrument, specifically a \correct{grasper} or dissector, used to manipulate or hold tissue during surgery.
            \item \textbf{2:} This seems to be the \correct{gallbladder}, which is the main target of a cholecystectomy.
            \item \textbf{3:} This could represent the \wrong{cystic duct, cystic artery, or one of the connective structures leading to the gallbladder}.
            \item \textbf{4:} These appear to be loops of the \wrong{small intestine, possibly the duodenum or jejunum.}
            \item \textbf{5:} This could be another \wrong{part of the biliary tree or surrounding vasculature,} but it's difficult to discern exactly without more context.
        \end{itemize}
        
    \item \textbf{response 2:}
        \begin{itemize}
            \item \textbf{1:} Instrument (probably a laparoscopic \correct{grasper} or dissector): This tool is commonly used during laparoscopic surgery to hold, dissect, or manipulate tissues.
            \item \textbf{2:} \correct{Gallbladder}, this pear-shaped organ stores bile produced by the liver. During a cholecystectomy, it's removed due to gallstones or other gallbladder diseases.
            \item \textbf{3:} \wrong{Cystic duct}, this duct is a connection between the gallbladder and the common bile duct. It's usually clipped and cut during the procedure.
            \item \textbf{4:} \wrong{Liver}, The large organ at the top-right is the liver, which produces bile that is stored in the gallbladder.
            \item \textbf{5:} \wrong{Instrument (likely another laparoscopic tool).} It's hard to determine the exact type from this angle, but it's another surgical instrument used in the procedure.
        \end{itemize}

    \item \textbf{response 3:}
        \begin{itemize}
            \item \textbf{1:} This appears to be a laparoscopic instrument, possibly a dissector or \correct{grasper}, used to manipulate or hold tissues during the surgery.
            \item \textbf{2:} This tissue looks like the \correct{gallbladder}. During a cholecystectomy, the gallbladder is often dissected free from its attachments and removed.
            \item \textbf{3:} This area seems to represent the \correct{liver}, which the gallbladder sits underneath.
            \item \textbf{4:} This could be a part of the \correct{omentum} or another adjacent structure within the abdominal cavity.
            \item \textbf{5:} This region seems to be another \wrong{laparoscopic instrument} or a piece of surgical equipment.
        \end{itemize}
        
    \end{itemize}
\vspace{1em}
\end{tcolorbox}
\vspace{1em}
\end{flushleft}

\end{minipage}
}
\end{center}

\begin{center}
\fbox{%
\begin{minipage}{0.9\textwidth}
\begin{center}
    {\relsize{2}\textbf{CASE 1}}
\end{center}

\begin{center}
    \textbf{Appearances/characteristics described}
\end{center}

\vspace{1em}
\noindent
\begin{tcolorbox}[colback=gray!10,arc=0mm,auto outer arc]
\textbf{Prompt:} This is a picture of cholecystectomy In the picture, there are 5 marked numbers. Please tell me which tissues or instruments correspond to the marked numbers in the picture. The liver is a dark reddish-brown organ, the gallbladder is a pear-shaped organ, both of them are under the abdomen and covered by fatty omentum.
\end{tcolorbox}
\centering
\vspace{1em}
\newcommand{\correct}[1]{\sethlcolor{green!20}\hl{#1}}
\newcommand{\wrong}[1]{\sethlcolor{red!20}\hl{#1}}

\begin{flushleft}
\textbf{GPT-4V Response:}
\begin{tcolorbox}[colback=cyan!10,arc=0mm,auto outer arc]
    \begin{itemize}
    \item \textbf{response 1:}
        \begin{itemize}
            \item \textbf{1:} This appears to be a surgical instrument, possibly a \correct{laparoscopic grasper} or dissector.
            \item \textbf{2:} This appears to be the \correct{gallbladder}, which is a pear-shaped organ.
            \item \textbf{3:} This area appears to be part of the \correct{liver}, the dark reddish-brown organ you mentioned.
            \item \textbf{4:} This might represent the \correct{fatty omentum} or some other tissue covering the abdominal organs.
            \item \textbf{5:} It's a bit challenging to ascertain without additional context, but it could be another \wrong{part of the liver or adjacent tissue.}
        \end{itemize}

    \item \textbf{response 2:}
        \begin{itemize}
            \item \textbf{1:} \sethlcolor{yellow!20}\hl{Instrument (looks like a laparoscopic instrument, possibly for retraction or dissection).}
            \item \textbf{2:} \correct{Gallbladder (pear-shaped organ as mentioned).}
            \item \textbf{3:} \correct{Liver (dark reddish-brown organ as described).}
            \item \textbf{4:} \correct{Fatty omentum or surrounding adipose tissue.}
            \item \textbf{5:} \wrong{Another section of the liver or possibly the edge of the liver. }
        \end{itemize}

    \item \textbf{response 3:}
        \begin{itemize}
            \item \textbf{1:} Surgical instrument, likely a \correct{grasper} or dissector used to manipulate or hold tissues.
            \item \textbf{2:} \correct{Gallbladder}, The pear-shaped organ you mentioned. It stores bile produced by the liver.
            \item \textbf{3:} \correct{Liver}, The dark reddish-brown organ that produces bile and performs various other metabolic functions
            \item \textbf{4:} \correct{Fat} or connective tissue surrounding the liver and gallbladder.
            \item \textbf{5:} Possibly a part of the \wrong{bile duct} or another structure close to the gallbladder. 
        \end{itemize}
        
    \end{itemize}
\vspace{1em}
\end{tcolorbox}
\vspace{1em}
\end{flushleft}

\end{minipage}
}
\end{center}
\subsubsection{Case 2: CT image annotation recognition}
\begin{center}
\fbox{%
\begin{minipage}{0.9\textwidth}
\begin{center}
    {\relsize{2}\textbf{CASE 2}}
\end{center}
\centering
\vspace{1em}
\begin{figure}[H]
  \centering
  \includegraphics[width=0.9\linewidth]{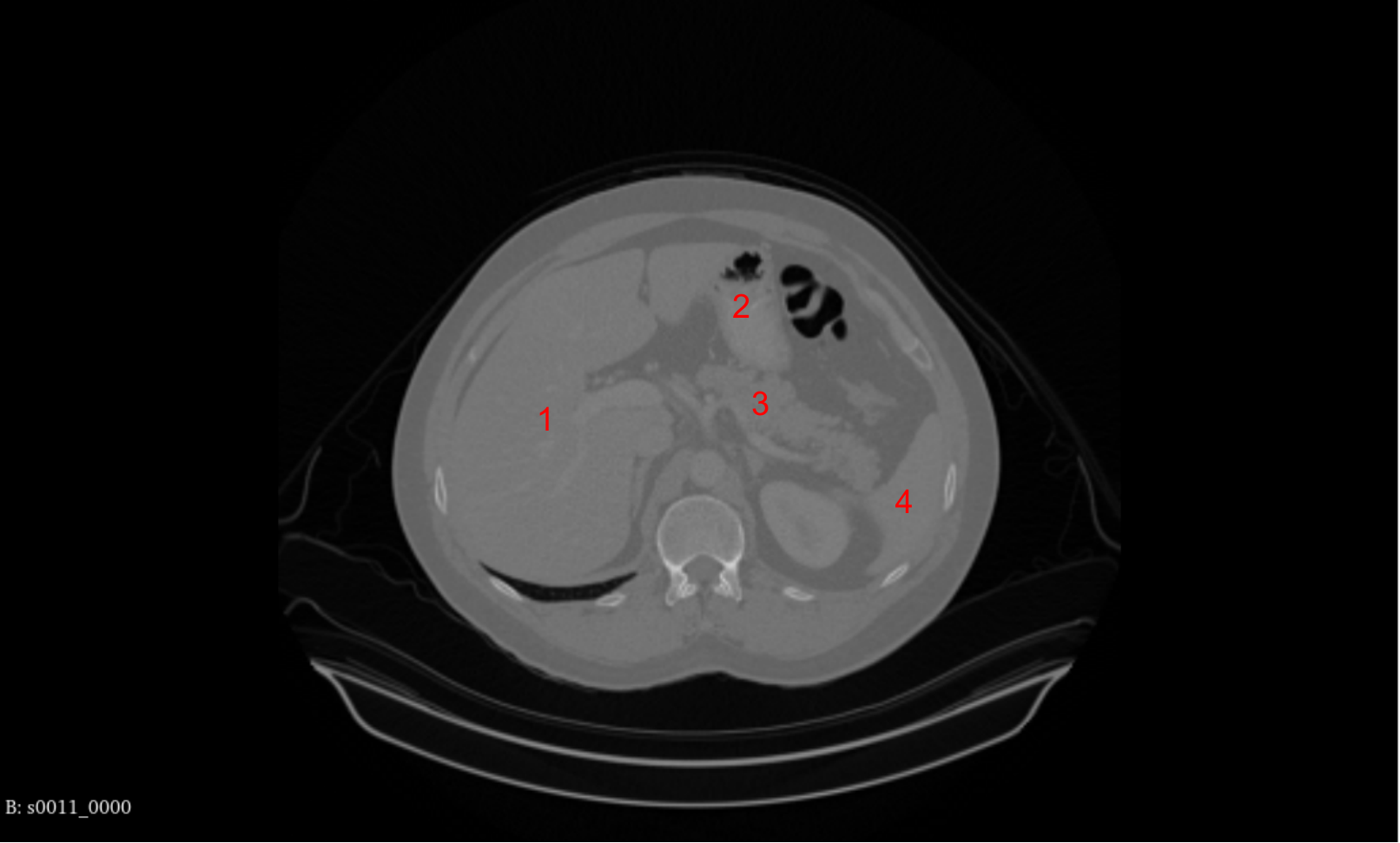}
  \label{C5_CASE2}
\end{figure}

\vspace{1em}
\begin{flushleft}
\begin{itemize}

     \item \textbf{Picture modal:} CT section
     \item \textbf{Reference Answer:} 
            \begin{itemize}
                \item \textbf{1:} Liver.
                \item \textbf{2:} Stomach.
                \item \textbf{3:} Pancreas.
                \item \textbf{4:} Spleen.
            \end{itemize}

\end{itemize}     
\vspace{1em}
\end{flushleft}

\end{minipage}
}
\end{center}
\vfill
\begin{center}
\fbox{%
\begin{minipage}{0.9\textwidth}
\begin{center}
    {\relsize{2}\textbf{CASE 2}}
\end{center}

\begin{center}
    \textbf{Control Group}
\end{center}

\vspace{1em}
\noindent
\begin{tcolorbox}[colback=gray!10,arc=0mm,auto outer arc]
\textbf{Prompt:} This is a picture of an axial CT scan. In the picture, there are 4 marked numbers. Please tell me which tissues correspond to the marked numbers in the picture.
\end{tcolorbox}
\centering
\vspace{1em}
\newcommand{\correct}[1]{\sethlcolor{green!20}\hl{#1}}
\newcommand{\wrong}[1]{\sethlcolor{red!20}\hl{#1}}

\begin{flushleft}
\textbf{GPT-4V Response:}
\begin{tcolorbox}[colback=cyan!10,arc=0mm,auto outer arc]
    \begin{itemize}
    \item \textbf{response 1:}
        \begin{itemize}
            \item \textbf{1:} \correct{Liver: This is the largest organ in the abdominal cavity and is usually located in the upper right quadrant of the abdomen.} It appears as a large, uniform area on a CT scan
            \item \textbf{2:} \correct{Stomach: Often containing air or ingested materials, which makes it appear darker (radiolucent) on a CT scan compared to the surrounding tissues.}
            \item \textbf{3:} \wrong{Spine: Visible in the center of the image as a circular structure, which is the vertebral body.} The spinal canal can also be seen as a small dark (radiolucent) circle within the vertebral body.
            \item \textbf{4:} \wrong{Kidneys: Typically, the kidneys are located on either side of the spine.} They have a bean-like shape and are less dense than the liver but more dense than the surrounding fat.
        \end{itemize}
        
    \item \textbf{response 2:}
        \begin{itemize}
            \item \textbf{1:} \correct{This likely indicates the liver}, which is a large organ situated in the upper right quadrant of the abdomen.
            \item \textbf{2:} \correct{This appears to be the stomach}, which often contains air or ingested materials, making it appear darker on a CT scan due to the air contrast.
            \item \textbf{3:} \wrong{This could be the aorta}, the main artery that carries blood from the heart to the rest of the body, although without contrast it's a bit harder to identify.
            \item \textbf{4:} \wrong{This seems to be the vertebral body, part of the spine.}
        \end{itemize}

    \item \textbf{response 3:}
        \begin{itemize}
            \item \textbf{1:} \wrong{This area typically represents the abdominal wall or possibly subcutaneous tissue.} It could also be part of the liver, depending on the specific location and context of the slice.
            \item \textbf{2:} \correct{This is likely to be the stomach}, especially if the dark areas within it represent ingested material or air.
            \item \textbf{3:} \wrong{This could be a part of the intestines or bowel}; the exact part (small vs. large intestine) would depend on the location within the abdomen and the presence of any distinguishing features.
            \item \textbf{4:} \wrong{This could be part of the musculature of the back}, like the psoas muscle, or potentially the kidney if it's located more posteriorly and laterally.
        \end{itemize}
        
    \end{itemize}
\vspace{1em}
\end{tcolorbox}
\vspace{1em}
\end{flushleft}

\end{minipage}
}
\end{center}

\begin{center}
\fbox{%
\begin{minipage}{0.9\textwidth}
\begin{center}
    {\relsize{2}\textbf{CASE 2}}
\end{center}

\begin{center}
    \textbf{Appearances/Characteristics Described}
\end{center}

\vspace{1em}
\noindent
\begin{tcolorbox}[colback=gray!10,arc=0mm,auto outer arc]
\textbf{Prompt:} This is a picture of an axial CT scan. In the picture, there are 4 marked numbers. Please tell me which tissues correspond to the marked numbers in the picture. The stomach typically appears as a hollow structure located near the midline in the upper part of the abdomen. The liver is a wedge-shaped organ with a structure that can be divided into two main lobes. The spleen has a shape similar to a thick oval or convex lens. The pancreas presents horizontally, resembling an axe lying on the spine.

\end{tcolorbox}
\centering
\vspace{1em}
\newcommand{\correct}[1]{\sethlcolor{green!20}\hl{#1}}
\newcommand{\wrong}[1]{\sethlcolor{red!20}\hl{#1}}

\begin{flushleft}
\textbf{GPT-4V Response:}
\begin{tcolorbox}[colback=cyan!10,arc=0mm,auto outer arc]
    \begin{itemize}
    \item \textbf{response 1:}
        \begin{itemize}
            \item \textbf{1:}\correct{Liver: The liver is typically the largest organ seen in the upper abdomen on the right side of the body in axial CT images}. It has a characteristic wedge shape and is divided into lobes. It's usually homogenous in texture on a CT scan without contrast.
            \item \textbf{2:}\correct{Stomach: The stomach is usually identified as a hollow, air-filled or fluid-filled structure}. It can vary in shape and size depending on the amount of content it holds at the time of the scan. It's located near the midline, typically more on the left side of the upper abdomen.
            \item \textbf{3:} \correct{Pancreas: The pancreas lies horizontally in the upper abdomen and is located posterior to the stomach.} It's often described as having an elongated shape resembling an axe or a tadpole, lying against the spine.
            \item \textbf{4:} \correct{Spleen: The spleen is usually located in the upper left part of the abdomen, appearing as a well-defined, homogenous soft tissue mass.} It has a convex lens-like or ovoid shape on axial images.
        \end{itemize}

    \item \textbf{response 2:}
        \begin{itemize}
            \item \textbf{1:} \correct{Might correspond to the liver}, which is usually visible as a large organ in the right upper quadrant of the abdomen.
            \item \textbf{2:} \correct{Could indicate the stomach}, especially if it appears as a hollow, air-filled structure.
            \item \textbf{3:} \correct{May represent the pancreas}, which would be seen as a softer tissue density sitting posterior to the stomach, and anterior to the spine.
            \item \textbf{4:} \correct{Could be pointing to the spleen}, located in the left upper quadrant of the abdomen, showing up as a soft tissue density similar to the liver.
        \end{itemize}

    \item \textbf{response 3:}
        \begin{itemize}
            \item \textbf{1:} \correct{Liver: The liver is typically visible on the right side of the abdomen when viewed on a CT scan}. It has a homogeneous, soft tissue density and is larger than other organs in the abdominal cavity.
            \item \textbf{2:} \correct{Stomach: The stomach is usually located on the left side under the ribs, appearing as a hollow}, sometimes air-filled structure on CT scans. Depending on whether it is filled with ingested material or not, its appearance may vary.
            \item \textbf{3:} \correct{Pancreas: The pancreas lies horizontally across the posterior abdominal wall and will generally appear as a soft tissue density structure that lies posterior to the stomach and anterior to the spine.}
            \item \textbf{4:} \correct{Spleen: The spleen is typically found in the left upper quadrant of the abdomen and may resemble a thick oval or lens shape on cross-sectional imaging like CT scans.}
        \end{itemize}
        
    \end{itemize}
\vspace{1em}
\end{tcolorbox}
\vspace{1em}
\end{flushleft}

\end{minipage}
}
\end{center}
\subsubsection{Case 3: MRI image annotation recognition}
\begin{center}
\fbox{%
\begin{minipage}{0.9\textwidth}
\begin{center}
    {\relsize{2}\textbf{CASE 3}}
\end{center}
\centering
\vspace{1em}
\begin{figure}[H]
  \centering
  \includegraphics[width=0.9\linewidth]{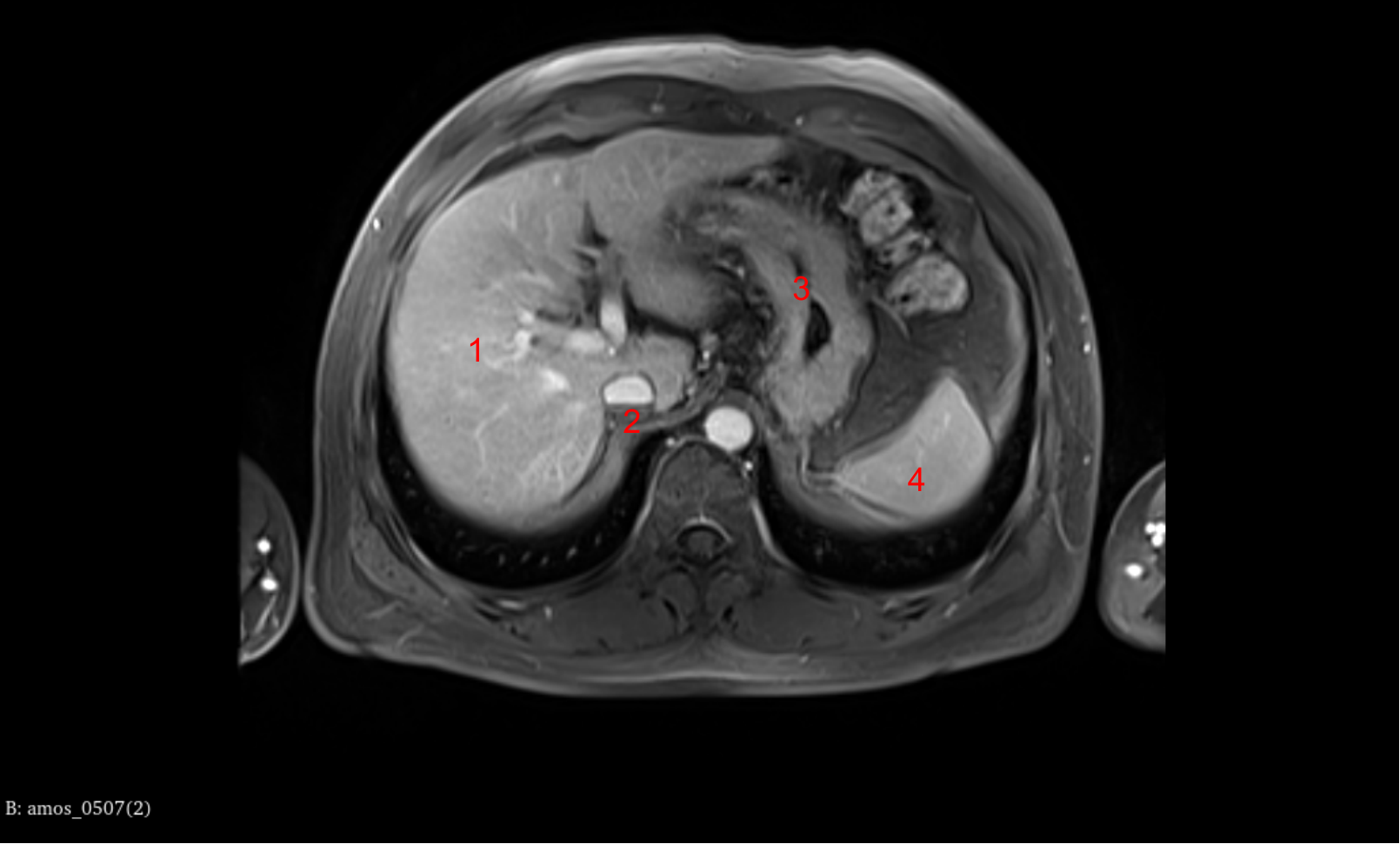}
  \label{C5_CASE3}
\end{figure}

\vspace{1em}
\begin{flushleft}
\begin{itemize}

     \item \textbf{Picture modal:} MRI section 
     \item \textbf{Reference Answer:} 
            \begin{itemize}
                \item \textbf{1:} Liver.
                \item \textbf{2:} Gallbladder.
                \item \textbf{3:} Pancreas.
                \item \textbf{4:} Spleen.
            \end{itemize}

\end{itemize}     
\vspace{1em}
\end{flushleft}

\end{minipage}
}
\end{center}
\vfill
\begin{center}
\fbox{%
\begin{minipage}{0.9\textwidth}
\begin{center}
    {\relsize{2}\textbf{CASE 3}}
\end{center}

\begin{center}
    \textbf{Control Group}
\end{center}

\vspace{1em}
\noindent
\begin{tcolorbox}[colback=gray!10,arc=0mm,auto outer arc]
\textbf{Prompt:} This is a picture of an axial MRI scan. In the picture, there are 4 marked numbers. Please tell me which tissues correspond to the marked numbers in the picture.
\end{tcolorbox}
\centering
\vspace{1em}
\newcommand{\correct}[1]{\sethlcolor{green!20}\hl{#1}}
\newcommand{\wrong}[1]{\sethlcolor{red!20}\hl{#1}}
\newcommand{\halfcorrect}[1]{\sethlcolor{yellow!20}\hl{#1}}

\begin{flushleft}
\textbf{GPT-4V Response:}
\begin{tcolorbox}[colback=cyan!10,arc=0mm,auto outer arc]
    \begin{itemize}
    \item \textbf{response 1:}
        \begin{itemize}
            \item \textbf{1:} \correct{This number is pointing to the region that is likely the liver}. It is a large organ located in the upper right quadrant of the abdomen.
            \item \textbf{2:} \wrong{The area labeled here is central and could represent the abdominal aorta or the spine}, depending on the exact location and orientation of the scan.
            \item \textbf{3:}\halfcorrect{This is likely pointing towards the stomach or pancreas}, which are located in the upper left quadrant of the abdomen.
            \item \textbf{4:}The location of this number \halfcorrect{seems to be over the spleen or the left kidney}, depending on the exact axial level of the MRI scan.
        \end{itemize}
        
    \item \textbf{response 2:}
        \begin{itemize}
            \item \textbf{1:} \correct{This area is typically where the liver is located}. The brightness on MRI suggests it could be a T2 weighted image where fluids appear bright, or a post-contrast T1 weighted image where the liver parenchyma enhances with contrast.
            \item \textbf{2:} \wrong{This could be the aorta}, the main artery that runs from the heart down through the chest and abdomen.
            \item \textbf{3:} \wrong{This is likely the stomach}, which can be identified by its location and the dark (on T2) or bright (on T1 post-contrast) appearance due to the fluid or air inside.
            \item \textbf{4:} \correct{This area could correspond to the spleen}, which sits in the left upper quadrant of the abdomen, posterior to the stomach.
        \end{itemize}

    \item \textbf{response 3:}
        \begin{itemize}
            \item \textbf{1:} \correct{This number is pointing to an area that is part of the liver}, which is a large organ in the upper right part of the abdomen, involved in many metabolic processes.
            \item \textbf{2:} \wrong{This could be pointing to the aorta or the inferior vena cava}, depending on its exact anatomical location as these are the major blood vessels that run anterior to the spine through the abdomen.
            \item \textbf{3:} \wrong{This number seems to be pointing to the spine or vertebrae}, which would be the bony structure in the center of the body at this level of an axial slice.
            \item \textbf{4:} \wrong{This appears to be pointing to the subcutaneous tissue and skin at the back of the body}, which is the lower-density area surrounding the body on an MRI scan.
        \end{itemize}
        
    \end{itemize}
\vspace{1em}
\end{tcolorbox}
\vspace{1em}
\end{flushleft}

\end{minipage}
}
\end{center}

\begin{center}
\fbox{%
\begin{minipage}{0.9\textwidth}
\begin{center}
    {\relsize{2}\textbf{CASE 3}}
\end{center}

\begin{center}
    \textbf{Appearances/Characteristics Described}
\end{center}

\vspace{1em}
\noindent
\begin{tcolorbox}[colback=gray!10,arc=0mm,auto outer arc]
\textbf{Prompt:} This is a picture of an axial MRI scan. In the picture, there are 4 marked numbers. Please tell me which tissues correspond to the marked numbers in the picture. The spleen typically presents as an oval or crescent-like structure, with its signal intensity usually slightly higher than that of the liver, exhibiting a fairly uniform texture. The liver typically displays a uniform texture with moderate to slightly high signal intensity. Its shape is similar to a large, somewhat irregular wedge. The pancreas appears as a slightly flattened, elongated structure. The gallbladder is presented as a small, streamlined or pear-shaped structure, located beneath the liver.

\end{tcolorbox}
\centering
\vspace{1em}
\newcommand{\correct}[1]{\sethlcolor{green!20}\hl{#1}}
\newcommand{\wrong}[1]{\sethlcolor{red!20}\hl{#1}}

\begin{flushleft}
\textbf{GPT-4V Response:}
\begin{tcolorbox}[colback=cyan!10,arc=0mm,auto outer arc]
    \begin{itemize}
    \item \textbf{response 1:}
        \begin{itemize}
            \item \textbf{1:}\correct{This could represent the liver}, as it's a large area with a somewhat irregular shape, consistent with the liver's appearance on an MRI scan.
            \item \textbf{2:}\correct{This marker could indicate the gallbladder}, given its location beneath the liver and its small, streamlined shape.
            \item \textbf{3:}\correct{This may correspond to the pancreas}, which would appear as a slightly flattened, elongated structure on an MRI scan.
            \item \textbf{4:}\correct{This likely represents the spleen}, which typically appears as an oval or crescent-like structure and is usually situated laterally in the abdomen, to the left of the body.
        \end{itemize}

    \item \textbf{response 2:}
        \begin{itemize}
            \item \textbf{1:}\correct{This number might indicate the liver}, based on its location in the upper right quadrant of the abdomen and the description of the liver's appearance on MRI.
            \item \textbf{2:}\correct{This could represent the gallbladder} if it's located under the liver and has the described pear-shaped appearance.
            \item \textbf{3:}\correct{This might be the spleen}, especially if it's on the upper left side of the abdomen and has a higher signal intensity than the liver.
            \item \textbf{4:}This number could point to the pancreas, which typically lies centrally within the abdomen, posterior to the stomach.
        \end{itemize}

    \item \textbf{response 3:}
        \begin{itemize}
            \item \textbf{1:} \correct{This could represent the liver}, given its large, somewhat irregular wedge shape, and it typically has a moderate to slightly high signal intensity on most MRI sequences.
            \item \textbf{2:} \correct{This might be the gallbladder} if it's located beneath the liver and appears as a small, streamlined or pear-shaped structure.
            \item \textbf{3:} \correct{The pancreas is often more central in the abdomen}, slightly flattened and elongated, and might correspond to this number.
            \item \textbf{4:} \correct{The spleen is typically located to the left of the body (which would be on the right side of the image as we look at it)}, and it may appear as an oval or crescent-like structure with signal intensity slightly higher than that of the liver.
        \end{itemize}
        
    \end{itemize}
\vspace{1em}
\end{tcolorbox}
\vspace{1em}
\end{flushleft}

\end{minipage}
}
\end{center}

\subsection{Tip 6: Appearance descriptors should be non-conflicting with any image annotations to avoid misinterpretations.}


\subsubsection{Case 1: Endoscopic image annotation recognition}
\begin{center}
\fbox{%
\begin{minipage}{0.9\textwidth}
\begin{center}
    {\relsize{2}\textbf{CASE 1}}
\end{center}
\centering
\vspace{1em}
\begin{figure}[H]
  \centering
  \includegraphics[width=0.9\linewidth]{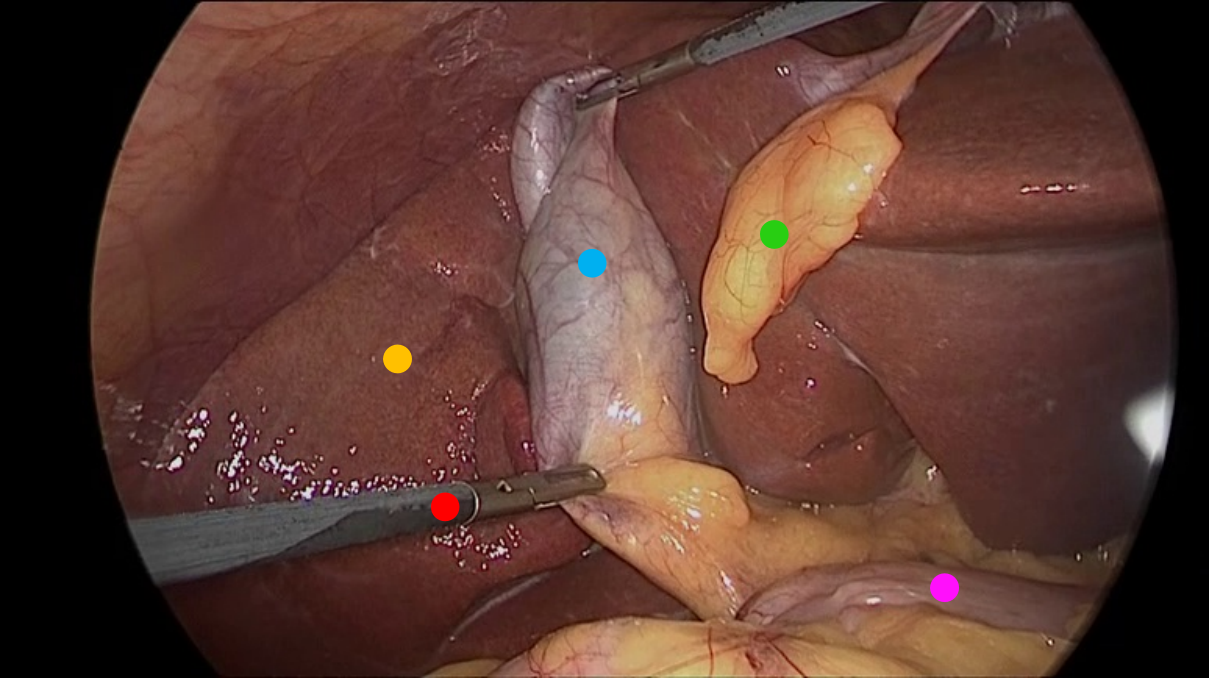}
  \label{CASE6}
\end{figure}

\vspace{1em}
\begin{flushleft}
\begin{itemize}

     \item \textbf{Picture modal:} Endoscopic picture
     \item \textbf{Tasks:} Laparoscopic cholecystectomy
     \item \textbf{Reference Answer:} 
            \begin{itemize}
                \item \textbf{Yellow Dot:} Liver.
                \item \textbf{Red Dot:} Surgical grasper.
                \item \textbf{Blue Dot:} Gallbladder.
                \item \textbf{Green Dot:} Fats.
                \item \textbf{Purple Dot:} Intestine.
            \end{itemize}

\end{itemize}    
\vspace{1em}
\end{flushleft}

\end{minipage}
}
\end{center}
\vfill

\begin{center}
\fbox{%
\begin{minipage}{0.9\textwidth}
\begin{center}
    {\relsize{2}\textbf{CASE 1}}
\end{center}

\begin{center}
    \textbf{Conflicted with annotation}
\end{center}

\vspace{1em}
\noindent
\begin{tcolorbox}[colback=gray!10,arc=0mm,auto outer arc]
\textbf{Prompt:} This is a picture of a cholecystectomy. In this picture, there are markers of different colors. Please tell me which tissues or instruments correspond to the annotations in the image. Intestinal tissue is typically pale pink, the liver is usually reddish-brown, the gallbladder is commonly grayish-green, and fat is generally pale yellow.
\end{tcolorbox}
\centering
\vspace{1em}
\newcommand{\correct}[1]{\sethlcolor{green!20}\hl{#1}}
\newcommand{\wrong}[1]{\sethlcolor{red!20}\hl{#1}}

\begin{flushleft}
\textbf{GPT-4V Response:}
\begin{tcolorbox}[colback=cyan!10,arc=0mm,auto outer arc]
    \begin{itemize}
    \item \textbf{response 1:}
        \begin{itemize}
            \item \textbf{Blue Dot:} This appears to be the \correct{gallbladder}, which is typically of a grayish-green hue.
            \item \textbf{Green Dot:} This section might represent \wrong{intestinal tissue}, often presented in a pale pink color.
            \item \textbf{Yellow Dot:} This could likely be \wrong{adipose tissue or fat}, as it usually appears pale yellow.
            \item \textbf{Red Dot:} This is a \sethlcolor{yellow!20}\hl{surgical instrument.}
            \item \textbf{Purple Dot:} This section might represent the \wrong{liver}, typically seen in a reddish-brown hue.
        \end{itemize}

    \item \textbf{response 2:}
        \begin{itemize}
            \item \textbf{Blue Dot:} \correct{Gallbladder}, typically seen in a grayish-green hue.
            \item \textbf{Green Dot:} \wrong{Intestinal tissue}, often presented in a pale pink color.
            \item \textbf{Yellow Dot:} \wrong{Adipose tissue, or fat}, usually of a pale yellow appearance.
            \item \textbf{Red Dot:} \correct{This instrument is a surgical clamp, used for grasping or holding tissue in place.}
            \item \textbf{Purple Dot:} \wrong{Liver, often seen in a reddish-brown hue.}
        \end{itemize}

    \item \textbf{response 3:}
        \begin{itemize}
            \item \textbf{Blue Dot:} This appears to resemble the \correct{gallbladder}, typically of a grayish-green hue.
            \item \textbf{Green Dot:} This section seems to be the \wrong{intestinal tissue}, seen in a pale pink hue.
            \item \textbf{Yellow Dot:} This could be \wrong{adipose tissue, given its pale yellow appearance}.
            \item \textbf{Red Dot:} This tool appears to be a \sethlcolor{yellow!20}\hl{medical instrument used for surgical procedures.}
            \item \textbf{Purple Dot:} This region \wrong{likely represents the liver, with its reddish-brown appearance.}
        \end{itemize}
        
    \end{itemize}
\vspace{1em}
\end{tcolorbox}
\vspace{1em}
\end{flushleft}

\end{minipage}
}
\end{center}
\vfill
\subsection{Tip 7: Clarifying the contextual relationships between sequential images enhances GPT-4v's analytical accuracy and detail recognition.}


\subsubsection{Case 1: Endoscopic video sequence analysis}
\begin{center}
\begin{framed}
\begin{minipage}{0.9\textwidth}
\begin{center}
    {\relsize{2}\textbf{CASE 1}}
\end{center}
\centering
\vspace{1em}
\begin{figure}[H]
  \centering
  \includegraphics[width=0.9\linewidth]{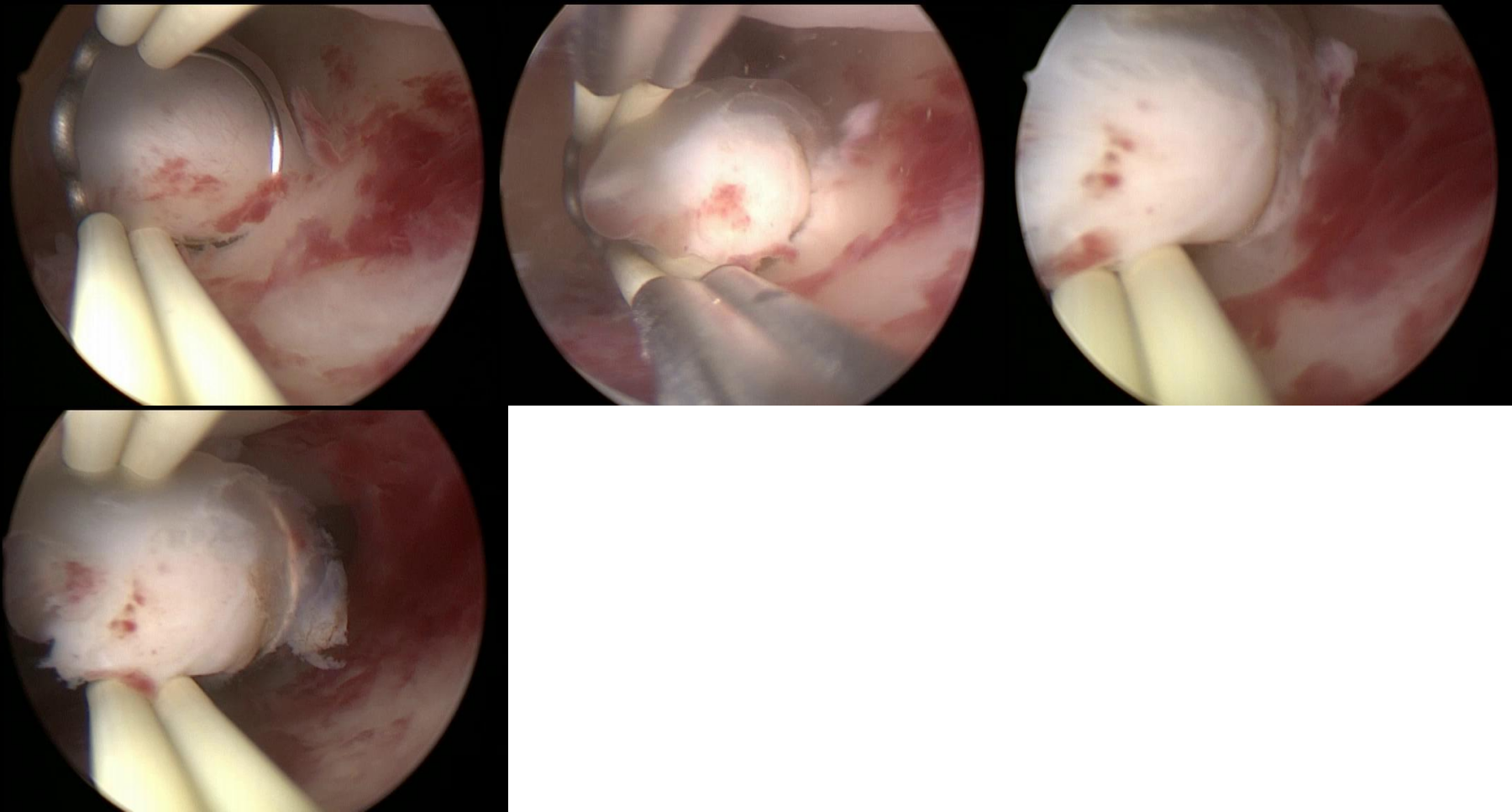}
  \label{C7_CASE1}
  \caption{Without series number}
\end{figure}

\begin{figure}[H]
  \centering
  \includegraphics[width=0.9\linewidth]{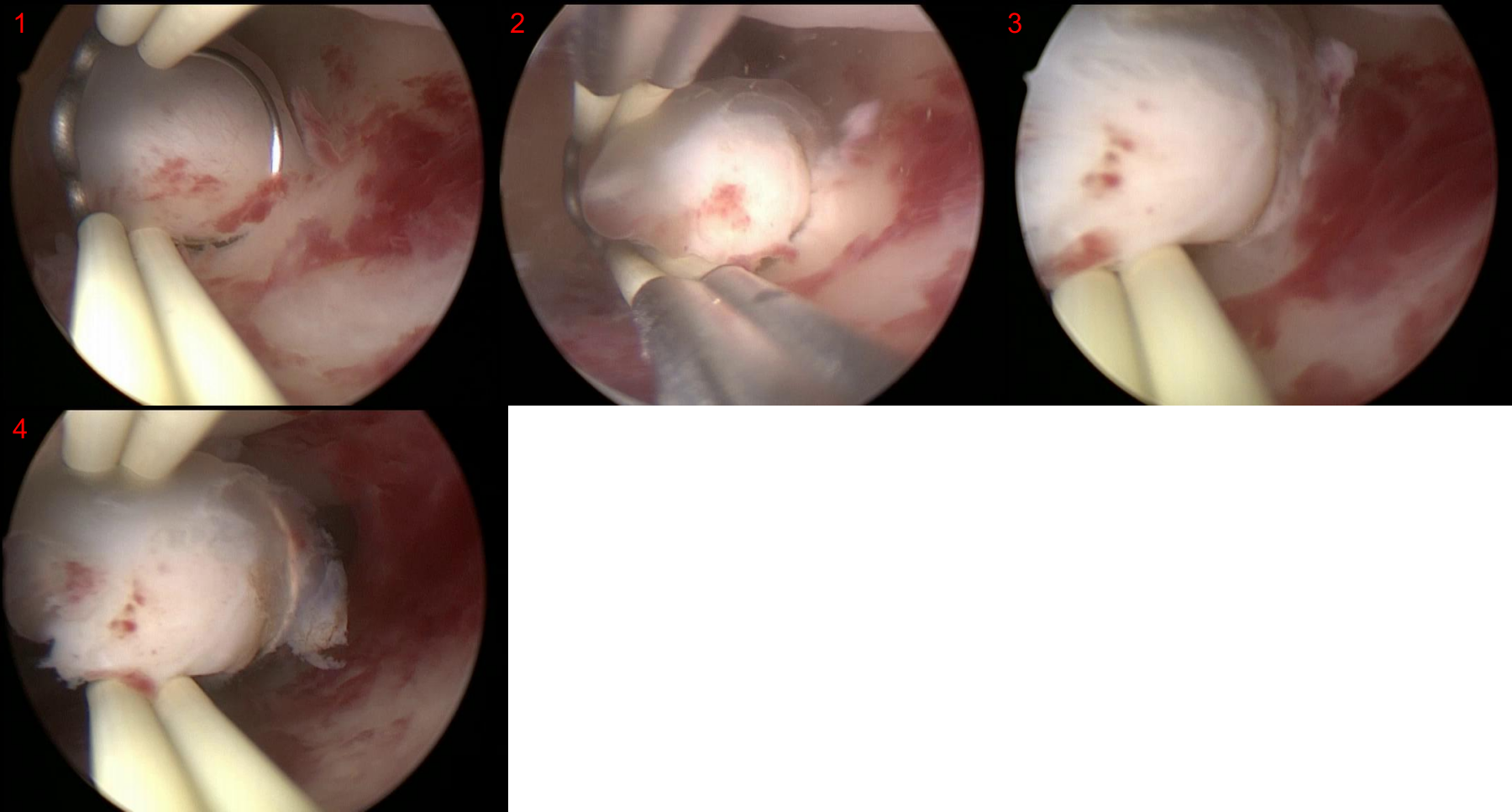}
  \label{C7_C8_CASE1}
  \caption{With series number}
\end{figure}

\vspace{1em}
\begin{flushleft}
\begin{itemize}

     \item \textbf{Picture modal:} Endoscopy
     \item \textbf{Tasks:} Hysteroscopic polypectomy
     \item \textbf{Reference Answer:} 
     \begin{itemize}
         \item \textbf{Pic.1: }Observe the surroundings, and locate the polyp.
         \item \textbf{Pic.2: }Excising the polyp.
         \item \textbf{Pic.3: }Further clearing of the polyp.
         \item \textbf{Pic.4: }Excision complete, removing the excised polyp.
     \end{itemize}

\end{itemize}     
\vspace{1em}
\end{flushleft}

\end{minipage}
\end{framed}
\end{center}

\begin{center}
\fbox{%
\begin{minipage}{0.9\textwidth}
\begin{center}
    {\relsize{2}\textbf{CASE 1}}
\end{center}

\begin{center}
    \textbf{Control Group (without contextual relationships)}
\end{center}

\vspace{1em}
\noindent
\begin{tcolorbox}[colback=gray!10,arc=0mm,auto outer arc]
\textbf{Prompt:} These four photos are from a hysteroscopic polypectomy. Please analyze the procedures in these four images.
\end{tcolorbox}
\centering
\vspace{1em}
\newcommand{\correct}[1]{\sethlcolor{green!20}\hl{#1}}
\newcommand{\wrong}[1]{\sethlcolor{red!20}\hl{#1}}

\begin{flushleft}
\textbf{GPT-4V Response:}
\begin{tcolorbox}[colback=cyan!10,arc=0mm,auto outer arc]
    \begin{itemize}
    \item \textbf{response 1:}
        \begin{itemize}
            \item \textbf{Picture 1:} The image shows a polyp within the uterine cavity. A medical instrument (possibly a suction device or curette) has been introduced into the cavity, \correct{poised to perform a procedure on the polyp.}
            \item \textbf{Picture 2:} The instrument is manipulating the polyp. \correct{It appears to be in the process of excision or curettage of the polyp.}
            \item \textbf{Picture 3:} \wrong{After the excision, portions of the polyp have been removed}, but remnants still persist within the uterine cavity.
            \item \textbf{Picture 4:} \wrong{There is continued clearing or treatment of the residuals inside the uterine cavity.}
        \end{itemize}
    \item \textbf{response 2:}
        \begin{itemize}
            \item \textbf{Photo 1:} \wrong{Assisted by hysteroscopy, the physician is examining the endometrial lining}. A distinct polyp is visible in this image, surrounded by some mild bleeding. Additionally, a surgical instrument (resembling forceps) is nearing the operational site.
            \item \textbf{Photo 2:} \wrong{The surgical instrument approaches the polyp, preparing for excision.} There's a slight increase in the bleeding surrounding the polyp.
            \item \textbf{Photo 3:} \correct{This is a pivotal moment in the excision procedure. The polyp, as evident from the image, has been grasped and is commencing detachment.}
            \item \textbf{Photo 4:} The polyp has been successfully excised, \correct{and the surgical instrument is extracting it from the uterine cavity. Hemorrhage within the cavity has been managed.}
        \end{itemize}

    \item \textbf{response 3:}
        \begin{itemize}
            \item \textbf{Photo 1:} This image offers a perspective from the hysteroscope, showcasing the endometrial lining and a polyp. \wrong{Hysteroscopic instruments are being employed to position and stabilize the polyp in preparation for its excision.}
            \item \textbf{Photo 2:} \correct{In this image, the tool has approached the polyp and commenced its excision}. There's a reddened region above the excision site, indicative of minor bleeding due to the surgical intervention.
            \item \textbf{Photo 3:} \correct{The image depicts a partially excised polyp}. A section of the polyp has been cut off, with the remaining still attached to the uterine wall.
            \item \textbf{Photo 4:} \wrong{This image presents the area post-polypectomy. There's a whitened zone on the endometrial lining, possibly a trace left after the polyp's removal.}
        \end{itemize}
        
    \end{itemize}
\vspace{1em}
\end{tcolorbox}
\vspace{1em}
\end{flushleft}

\end{minipage}
}
\end{center}
\begin{center}
\fbox{%
\begin{minipage}{0.9\textwidth}
\begin{center}
    {\relsize{2}\textbf{CASE 1}}
\end{center}

\begin{center}
    \textbf{With contextual relationships}
\end{center}

\vspace{1em}
\noindent
\begin{tcolorbox}[colback=gray!10,arc=0mm,auto outer arc]
\textbf{Prompt:} These four photos are from a hysteroscopic polypectomy. Please analyze the procedures in these four images.
\end{tcolorbox}
\centering
\vspace{1em}
\newcommand{\correct}[1]{\sethlcolor{green!20}\hl{#1}}
\newcommand{\wrong}[1]{\sethlcolor{red!20}\hl{#1}}

\begin{flushleft}
\textbf{GPT-4V Response:}
\begin{tcolorbox}[colback=cyan!10,arc=0mm,auto outer arc]
    \begin{itemize}
    \item \textbf{response 1:}
        \begin{itemize}
            \item \textbf{Picture 1:} \correct{Initial Observation}. At this stage, one can see the interior of the uterine cavity and a prominent polyp. At this point, the doctor might be looking for the best entry point and operational path.
            \item \textbf{Picture 2:} \correct{Beginning of Excision}. Using medical tools, the doctor starts excising the polyp from its base. This is a critical step that requires careful operation to ensure the polyp is completely removed without damaging surrounding tissues.
            \item \textbf{Picture 3:} \correct{Continued Excision}. At this stage, the polyp has been partially removed but still partially attached to the uterine wall. The doctor continues to operate carefully to ensure its complete removal.
            \item \textbf{Picture 4:} \sethlcolor{yellow!20}\hl{Polypectomy Complete. At this stage, one can observe that the polyp has been fully removed and the interior of the uterine cavity looks relatively clean.} Further cleaning or hemostasis might be performed.
        \end{itemize}
    \item \textbf{response 2:}
        \begin{itemize}
            \item \textbf{Picture 1:} The first picture displays the scene within the uterine cavity. \correct{One can observe part of the polyp and the inner walls of the uterus. Surgical instruments have already entered the view, preparing to operate on the polyp.}
            \item \textbf{Picture 2:} \correct{The second picture shows the process of the excision surgery}. The surgical instrument seems to be grasping and partially excising the polyp.
            \item \textbf{Picture 3:} \correct{The third picture further displays the steps during surgery. From the picture, one can observe that the polyp has already been partially excised.}
            \item \textbf{Picture 4:} The fourth picture showcases the situation inside the uterine cavity after the surgery. \correct{Most of the polyp seems to have been cleared, and surgical instruments might be conducting a final cleanup.}
        \end{itemize}
    \item \textbf{response 3:}
        \begin{itemize}
            \item \textbf{Picture 1:} \correct{The doctor might be using a hysteroscope to examine the condition of the endometrium and locate the polyp}. In the picture, one can observe a surgical instrument capturing or securing the polyp for better operation.
            \item \textbf{Picture 2:} \correct{The picture appears to show that the polyp is already clamped by the surgical instrument and is being excised.}
            \item \textbf{Picture 3:} \correct{The polyp has been partially excised}, and one can see the excised polyp tissue and the remaining part.
            \item \textbf{Picture 4:} This picture displays the state of the endometrium after complete excision. \wrong{The polyp has been fully removed, though there might be slight tissue remnants.}
        \end{itemize}
        
    \end{itemize}
\vspace{1em}
\end{tcolorbox}
\vspace{1em}
\end{flushleft}

\end{minipage}
}
\end{center}
\subsection{Tip 8: Splicing multiple images into one and providing a sequence order enhances task processing effectiveness compared to multiple simultaneous inputs.}


\subsubsection{Case 1: Endoscopic video sequence analysis}
\vspace{-2cm}
\begin{center}
\begin{framed}
\begin{minipage}{0.9\textwidth}
\begin{center}
    {\relsize{2}\textbf{CASE 1}}
\end{center}
\centering

\begin{figure}[H]
  \centering
  \includegraphics[width=0.25\linewidth]{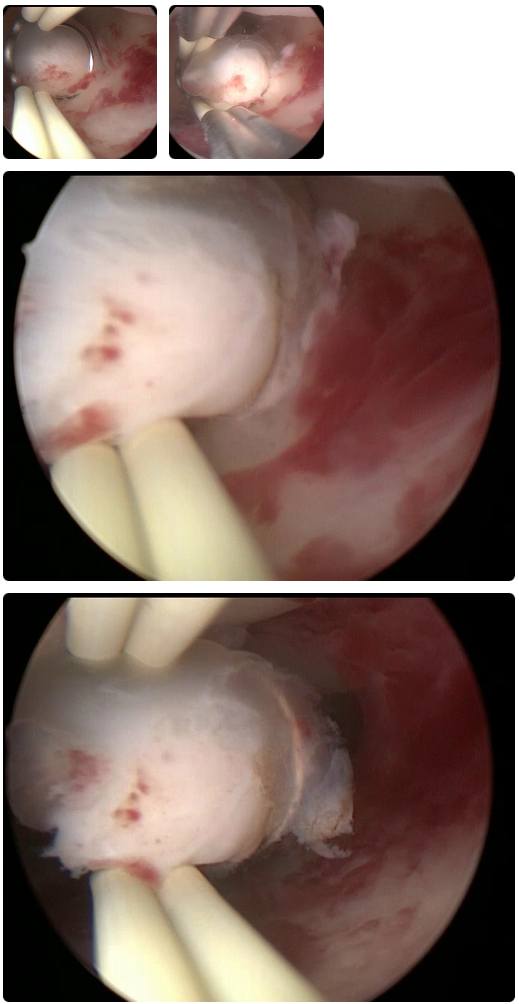}
  \label{C8_CASE1_systematic_upload}
  \caption{Uploaded one by one via GPT-4V website. The upload system of GPT-4V may cause different sizes to be displayed, but the actual sizes are not affected}
\end{figure}

\begin{figure}[H]
  \centering
  \includegraphics[width=0.9\linewidth]{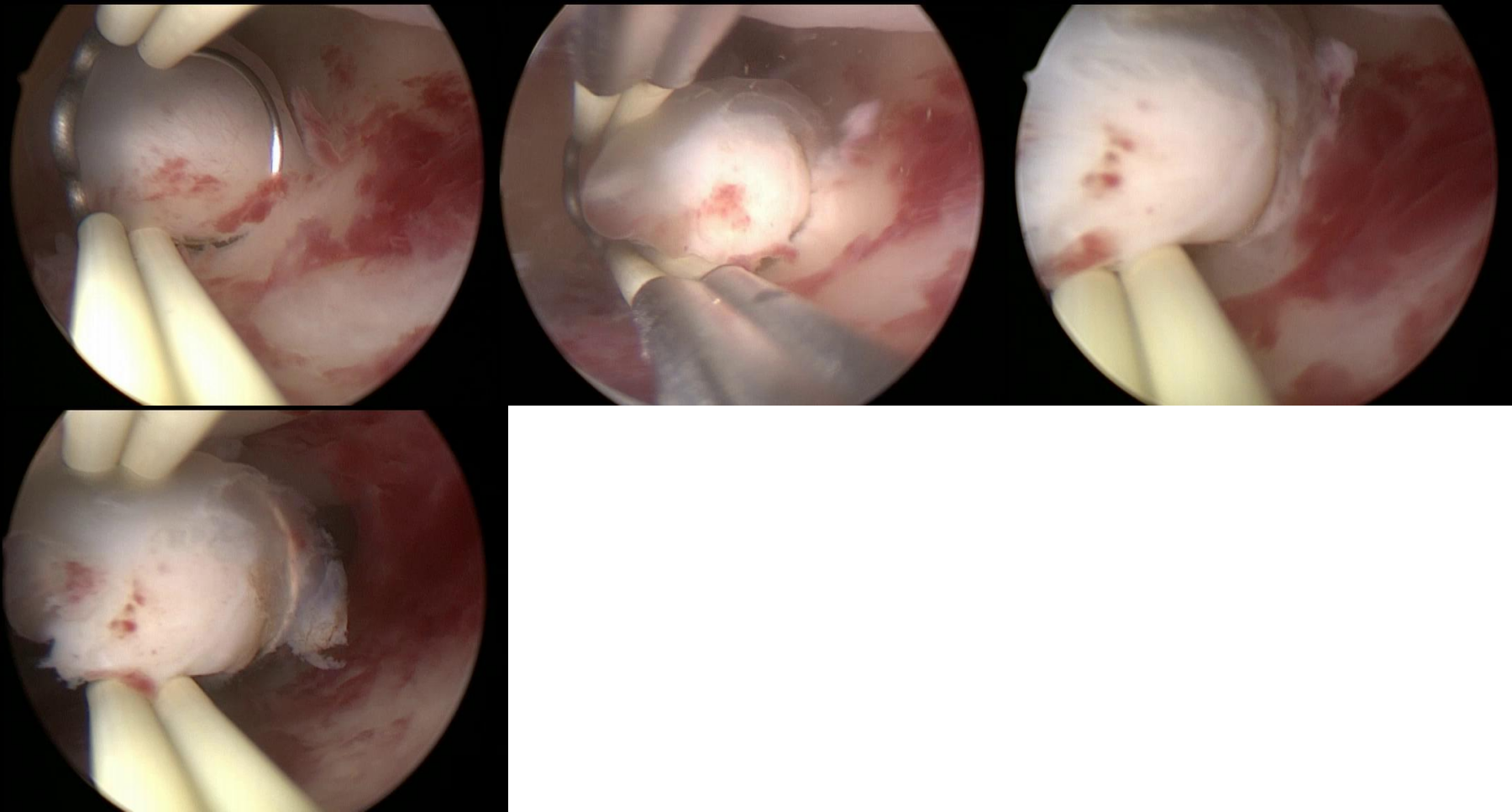}
  \label{C8_CASE1_combined}
  \caption{Uploaded after splicing}
\end{figure}

\begin{flushleft}
\begin{itemize}

     \item \textbf{Picture modal:} Endoscopy
     \item \textbf{Tasks:} Hysteroscopic polypectomy
     \item \textbf{Reference Answer:} 
     \begin{itemize}
         \item \textbf{Pic.1: }Observe the surroundings, and locate the polyp.
         \item \textbf{Pic.2: }Excising the polyp.
         \item \textbf{Pic.3: }Further clearing of the polyp.
         \item \textbf{Pic.4: }Excision complete, removing the excised polyp.
     \end{itemize}
     
\end{itemize}     
\end{flushleft}

\end{minipage}  
\end{framed}
\end{center}
\begin{center}
\fbox{%
\begin{minipage}{0.9\textwidth}
\begin{center}
    {\relsize{2}\textbf{CASE 1}}
\end{center}

\begin{center}
    \textbf{Uploaded one by one via GPT4-v website}
\end{center}

\vspace{1em}
\noindent
\begin{tcolorbox}[colback=gray!10,arc=0mm,auto outer arc]
\textbf{Prompt:} These four photos are from a hysteroscopic polypectomy. Please analyze the procedures in these four images.
\end{tcolorbox}
\centering
\vspace{1em}
\newcommand{\correct}[1]{\sethlcolor{green!20}\hl{#1}}
\newcommand{\wrong}[1]{\sethlcolor{red!20}\hl{#1}}

\begin{flushleft}
\textbf{GPT-4V Response:}
\begin{tcolorbox}[colback=cyan!10,arc=0mm,auto outer arc]
    \begin{itemize}
    \item \textbf{response 1:}
        \begin{itemize}
            \item \textbf{Picture 1:} \wrong{Showcases a slightly larger uterine polyp. One can observe the base of the polyp and a certain reddish area of bleeding.}
            \item \textbf{Picture 2:} \wrong{It seems to illustrate the doctor using a tool to grip or stabilize the polyp, preparing for excision.}
            \item \textbf{Picture 3:} \wrong{This image displays a close-up view of the polyp with signs of bleeding and potential inflammation on its surface.}
            \item \textbf{Picture 4:} \wrong{A tool approaches the polyp area, possibly in the process of excision or preparing for it.}
        \end{itemize}

    \item \textbf{response 2:}
        \begin{itemize}
            \item \textbf{Picture 1:} \wrong{Displays the red tissue inside the uterus, which might be a polyp or another intrauterine tissue with hysteroscopic tools visible on the side.}
            \item \textbf{Picture 2:} \correct{The tool seems to be in contact with or manipulating a certain tissue, possibly a step in the excision process.}
            \item \textbf{Picture 3:} \correct{This image too appears to show the tool in contact with or manipulating tissue.}
            \item \textbf{Picture 4:} \wrong{The tool is near the tissue, suggesting that a part of the excision procedure might have been completed.}
        \end{itemize}

    \item \textbf{response 3:}
        \begin{itemize}
            \item \textbf{Picture 1:} \wrong{Seems to display a polyp inside the uterine cavity.} Polyps are growths of the endometrium, usually benign, but sometimes require excision to avoid complications or for further pathological analysis.
            \item \textbf{Picture 2:} \correct{This image shows that the doctor might be using surgical tools like forceps or scissors to excise the polyp.}
            \item \textbf{Picture 3:} \wrong{The view in this image might be of the uterine cavity after the excision of the polyp}, showcasing certain bleeding or erythema, which is normal post-operatively.
            \item \textbf{Picture 4:} \wrong{This image too shows the surgical area, looking like the uterine cavity post-polyp excision.}
        \end{itemize}
        
    \end{itemize}
\vspace{1em}
\end{tcolorbox}
\vspace{1em}
\end{flushleft}

\end{minipage}
}
\end{center}

\subsection{Tip 9: Providing comparative analysis opportunities, especially with temporal patient data, deepens lesion or condition assessments.}
\subsubsection{Case 1: CT scan tracking of pulmonary nodule}
\begin{center}
\fbox{%
\begin{minipage}{0.9\textwidth}
\begin{center}
    {\relsize{2}\textbf{CASE 1}}
\end{center}
\centering
\vspace{1em}
\begin{figure}[H]
  \centering
  \includegraphics[width=0.4\linewidth]{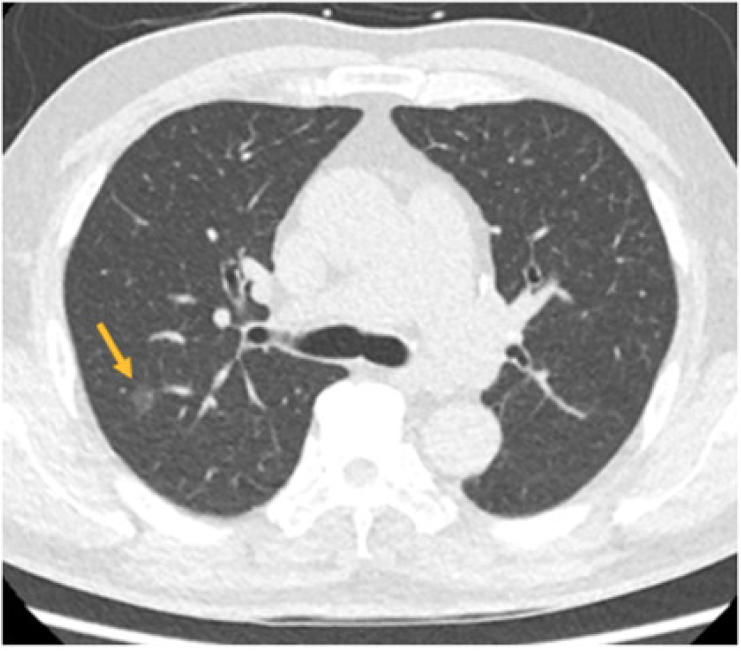}
  \label{C9_CASE1_single}
  \caption{Without comparison}
\end{figure}

\begin{figure}[H]
  \centering
  \includegraphics[width=0.9\linewidth]{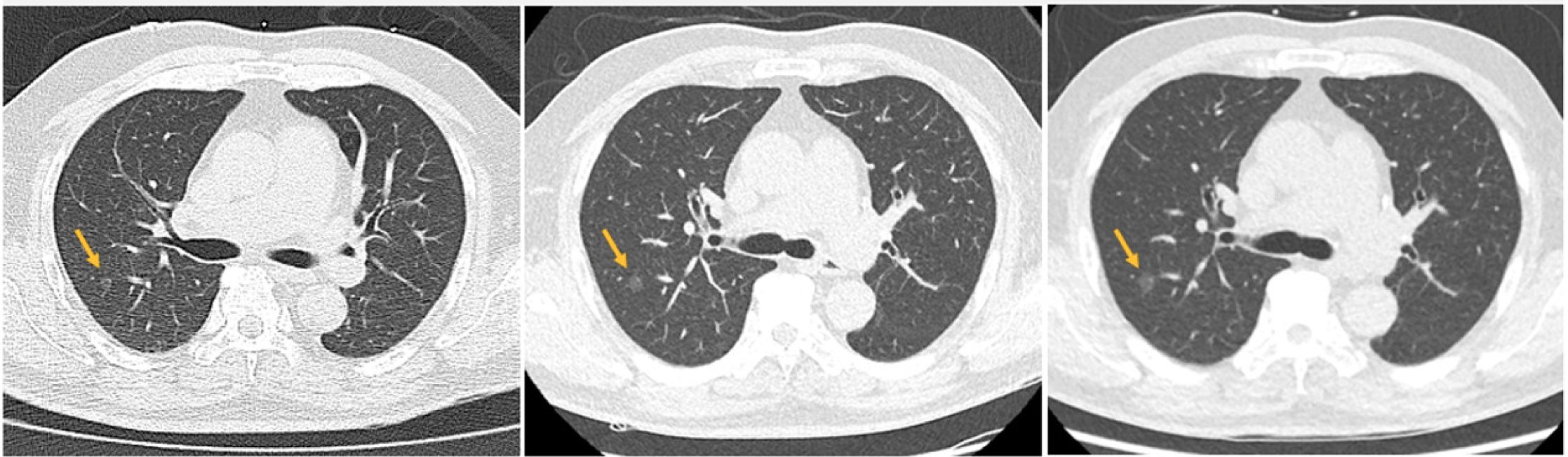}
  \label{C9_CASE1_comparision}
  \caption{With comparison}
\end{figure}

\vspace{1em}
\begin{flushleft}
\begin{itemize}

     \item \textbf{Picture modal:} CT scan
     \item \textbf{Tasks:} Pulmonary nodule growth tracking
     \item \textbf{Reference Answer:} adenocarcinoma 

\end{itemize}     
\vspace{1em}
\end{flushleft}

\end{minipage}
}
\end{center}
\vfill
\begin{center}
\fbox{%
\begin{minipage}{0.9\textwidth}
\begin{center}
    {\relsize{2}\textbf{CASE 1}}
\end{center}

\begin{center}
    \textbf{Singe picture(without comparison) \ref{C9_CASE1_single}}
\end{center}

\vspace{1em}
\noindent
\begin{tcolorbox}[colback=gray!10,arc=0mm,auto outer arc]
\textbf{Prompt:} It took 6 years for the patient's nodule to grow to the size indicated by the arrow in the image. What disease is the patient most likely to have 
\end{tcolorbox}
\centering
\vspace{1em}
\newcommand{\correct}[1]{\sethlcolor{green!20}\hl{#1}}
\newcommand{\wrong}[1]{\sethlcolor{red!20}\hl{#1}}
\begin{flushleft}
\textbf{GPT-4V Response:}
\begin{tcolorbox}[colback=cyan!10,arc=0mm,auto outer arc]
    \begin{itemize}
    \item \textbf{response 1:}

        Some benign conditions that can present with slow-growing lung nodules include:

        \textbf{Hamartoma}

        \textbf{Inflammatory nodules} (e.g., granulomatous infections like tuberculosis or fungal infections)

        \textbf{Rheumatoid nodules in patients with rheumatoid arthritis}

        \textbf{Scar tissue or old infections}

        \textbf{Some benign tumors} 

    \item \textbf{response 2:}

        Slow-growing nodules in the lungs are most commonly benign. Benign pulmonary nodules can arise from a number of causes, including:
      
        \textbf{Granulomas:} These are small areas of inflammation, often caused by past infections like tuberculosis or fungal infections.

        \textbf{Hamartomas:} These are benign growths made up of a mixture of tissue types commonly found in the lungs.

        \textbf{Other benign tumors or cysts.}

\correct{        However, it's also essential to consider the possibility of a slow-growing malignancy, like certain types of lung cancer (e.g., adenocarcinoma or carcinoid tumors).}

    \item \textbf{response 3:}

        Given the slow growth rate over 6 years, one of the possibilities is a benign lung nodule. Benign lung nodules can be due to various causes, including:
        
        \textbf{Hamartoma:} The most common type of benign lung tumor.

        \textbf{Granulomas:} Caused by inflammation in the lung.

        \textbf{Infectious causes:} Such as fungal infections or tuberculosis.

        \textbf{Other benign tumors or cysts.}

    \end{itemize}
\vspace{1em}
\end{tcolorbox}
\vspace{1em}
\end{flushleft}

\end{minipage}
}
\end{center}
\begin{center}
\fbox{%
\begin{minipage}{0.9\textwidth}
\begin{center}
    {\relsize{2}\textbf{CASE 1}}
\end{center}

\begin{center}
    \textbf{With comparison \ref{C9_CASE1_single}}
\end{center}

\vspace{1em}
\noindent
\begin{tcolorbox}[colback=gray!10,arc=0mm,auto outer arc]
\textbf{Prompt:} The patient's nodule has increased in size from left to right over 6 years, what kind of disease is the patient most likely to have 
\end{tcolorbox}
\centering
\vspace{1em}
\newcommand{\correct}[1]{\sethlcolor{green!20}\hl{#1}}
\newcommand{\wrong}[1]{\sethlcolor{red!20}\hl{#1}}

\begin{flushleft}
\textbf{GPT-4V Response:}
\begin{tcolorbox}[colback=cyan!10,arc=0mm,auto outer arc]
    \begin{itemize}
    \item \textbf{response 1:}
        \begin{itemize}
            \item \textbf{Primary lung cancer:} \correct{The slow growth might suggest a subtype like adenocarcinoma}, which can sometimes grow at a slower rate compared to other forms of lung cancer like small cell carcinoma. However, the speed of growth varies and cannot definitively diagnose the type.
            \item \textbf{Infectious causes:} Some infections can cause nodules that grow or change over time. Examples include fungal infections such as aspergillosis or coccidioidomycosis.
            \item \textbf{Benign tumors:} While benign tumors such as hamartomas or fibromas are typically slow-growing, they can still increase in size over time.
            \item \textbf{Inflammatory conditions:} Conditions such as rheumatoid nodules or granulomatous diseases can also present as lung nodules. The growth pattern might help differentiate between the different possibilities.
            \item \textbf{Metastatic cancers:} A lung nodule can be a metastasis from a primary tumor elsewhere in the body.    
        \end{itemize}
        
    \item \textbf{response 2:}
        \begin{itemize}
            \item \textbf{Lung Cancer:} Primary lung cancer is the most concerning possibility for a progressively enlarging nodule in the lung. \correct{It could be non-small cell lung cancer (e.g., adenocarcinoma, squamous cell carcinoma) or small cell lung cancer.}
            
            \item \textbf{Infectious Causes:} Some infections can cause nodules in the lung that can grow over time. This includes bacterial infections like tuberculosis, fungal infections like aspergillosis or histoplasmosis, or parasitic infections like echinococcosis.
            
            \item \textbf{Benign Tumors:} Some benign lung tumors, such as hamartomas or fibromas, can also grow, although at a slower rate than malignant ones.
            
            \item \textbf{Inflammatory Diseases:} Conditions like granulomatosis with polyangiitis, rheumatoid nodules in the context of rheumatoid arthritis, and sarcoidosis can cause lung nodules.
            
            \item \textbf{Metastases:} If the patient has a history of cancer elsewhere in the body, it's possible that the nodule represents metastatic disease.
        \end{itemize}   
        
    \item \textbf{response 3:}
    
        \begin{itemize}
            \item \textbf{Lung Cancer:} \correct{One of the most concerning possibilities for a slow-growing pulmonary nodule is lung cancer, like adenocarcinoma.} The risk is higher if the patient has a history of smoking, exposure to carcinogens, or a family history of lung cancer.
            
            \item \textbf{Infectious Causes:} Some infections, such as tuberculosis or fungal infections, can present as nodules that might change in size over time.
            
            \item \textbf{Benign Tumors:} Not all growing nodules are malignant. Some benign tumors can also grow over time.
            
            \item \textbf{Inflammatory Diseases:} Certain inflammatory conditions, like granulomas or rheumatoid nodules, can also appear as nodules in the lung.    
        \end{itemize} 
        
    \end{itemize}
\vspace{1em}
\end{tcolorbox}
\vspace{1em}
\end{flushleft}

\end{minipage}
}
\end{center}

\subsection{Tip 10: Directing GPT-4v's focus to the interested areas will facilitate more targeted and relevant output.}
\subsubsection{Case 1: MRI cerebral scan}
\begin{center}
\fbox{%
\begin{minipage}{0.9\textwidth}
\begin{center}
    {\relsize{2}\textbf{CASE 1}}
\end{center}
\centering
\vspace{1em}
\begin{figure}[H]
  \centering
  \includegraphics[width=0.9\linewidth]{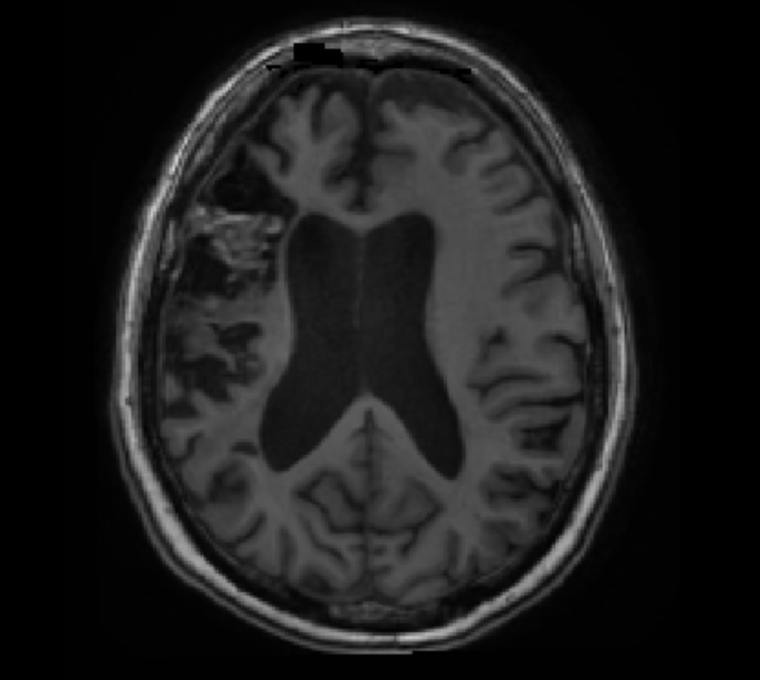}
  \label{C10_CASE1}
\end{figure}

\vspace{1em}
\begin{flushleft}
\begin{itemize}

     \item \textbf{Picture modal:} MRI
     \item \textbf{Tasks:} Cerebral lesions detection (after ischemic stroke)
     \item \textbf{Reference Answer:} A lesion is on the right frontal lobe (left part in image)

\end{itemize}     
\vspace{1em}
\end{flushleft}

\end{minipage}
}
\end{center}
\vfill
\begin{center}
\fbox{%
\begin{minipage}{0.9\textwidth}
\begin{center}
    {\relsize{2}\textbf{CASE 1}}
\end{center}

\begin{center}
    \textbf{Control Group}
\end{center}

\vspace{1em}
\noindent
\begin{tcolorbox}[colback=gray!10,arc=0mm,auto outer arc]
\textbf{Prompt:} Please examine this head MRI image and determine if it is normal (the information in this image is sufficient to make a judgment).
\end{tcolorbox}
\centering
\vspace{1em}
\newcommand{\correct}[1]{\sethlcolor{green!20}\hl{#1}}
\newcommand{\wrong}[1]{\sethlcolor{red!20}\hl{#1}}

\begin{flushleft}
\textbf{GPT-4V Response:}
\begin{tcolorbox}[colback=cyan!10,arc=0mm,auto outer arc]
    \begin{itemize}
    \item \textbf{response 1:}

        The image displays a T1-weighted magnetic resonance imaging (MRI) scan of the head. It's imperative to note that I can only provide general observations based on this image, which should not be construed as a substitute for a professional medical opinion. From the given image:

        \wrong{Most cerebral structures appear to be within normal parameters.}

        Midline structures, such as the boundary between the cerebral ventricles and the cerebral hemispheres, seem to be symmetrical.

        The dimensions of the third ventricle and lateral ventricles appear to be within the norm.

        The image's resolution and contrast might constrain a more intricate analysis.

    \item \textbf{response 2:}

        This represents a T1-weighted axial slice of a cranial MRI. From the scan:
        
        The majority of the cerebral structures can be discerned, including the cerebral cortex, the ventricular system, and the butterfly-shaped white matter region in the brain's center.

        \wrong{At the current resolution, there are no evident neoplasms, hemorrhagic regions, or other manifest pathologies.}

        The size and morphology of the ventricles also appear to be within standard limits.
        
    \item \textbf{response 3:}

        The image portrays an axial section of a cranial MRI. Upon initial examination:

        The cerebrum's symmetry seems to be devoid of notable anomalies.

        The ventricular system also appears unremarkable at first glance.

        \wrong{There should be no areas of abnormal intensity that could suggest the presence of lesions, tumors, or other abnormalities. The brain's gyri and sulci should follow a typical pattern without signs of displacement or compression. }
        
    \end{itemize}
\vspace{1em}
\end{tcolorbox}
\vspace{1em}
\end{flushleft}

\end{minipage}
}
\end{center}
\begin{center}
\fbox{%
\begin{minipage}{0.9\textwidth}
\begin{center}
    {\relsize{2}\textbf{CASE 1}}
\end{center}

\begin{center}
    \textbf{Directed to the interested area}
\end{center}

\vspace{1em}
\noindent
\begin{tcolorbox}[colback=gray!10,arc=0mm,auto outer arc]
\textbf{Prompt:} Please examine the upper left portion of this image and determine if it is normal (the information in this image is sufficient to make a judgment).
\end{tcolorbox}
\centering
\vspace{1em}
\newcommand{\correct}[1]{\sethlcolor{green!20}\hl{#1}}
\newcommand{\wrong}[1]{\sethlcolor{red!20}\hl{#1}}

\begin{flushleft}
\textbf{GPT-4V Response:}
\begin{tcolorbox}[colback=cyan!10,arc=0mm,auto outer arc]
    \begin{itemize}
    \item \textbf{response 1:}
        Based on the provided MRI image, a \correct{hyperintense region is evident in the upper left quadrant}, distinctly contrasting with the surrounding cerebral tissue. This region may signify a potential anomaly or lesion. However, the precise nature and clinical significance necessitate a more comprehensive medical evaluation for determination.

    \item \textbf{response 2:}
        In the MRI image you've presented, there appears to be a \correct{hyperintense area in the upper left section}, contrasting with the adjacent cerebral tissue. This could indicate the presence of a lesion or abnormality.

    \item \textbf{response 3:}
        On the MRI image you provided, a \correct{hyperintense region is discernible in the upper left quadrant.} Based on the image, this region seems to deviate from the typical cerebral tissue presentation. From this observation, the region may represent an aberration.
        
    \end{itemize}
\vspace{1em}
\end{tcolorbox}
\vspace{1em}
\end{flushleft}

\end{minipage}
}
\end{center}
\subsubsection{Case 2: PET scan for lung cancer detection}
\begin{center}
\fbox{%
\begin{minipage}{0.9\textwidth}
\begin{center}
    {\relsize{2}\textbf{CASE 2}}
\end{center}
\centering
\vspace{1em}
\begin{figure}[H]
  \centering
  \includegraphics[width=0.9\linewidth]{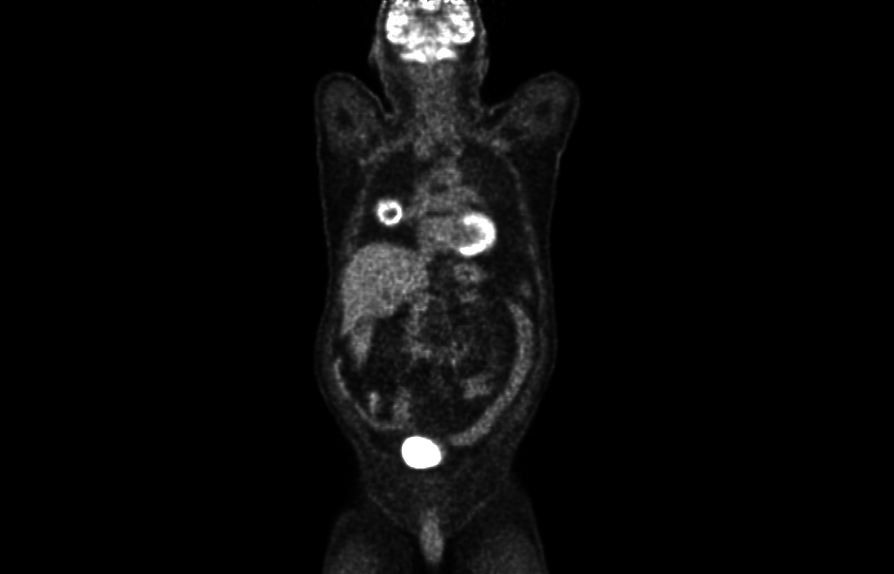}
  \label{C10_CASE2}
\end{figure}

\vspace{1em}
\begin{flushleft}
\begin{itemize}

     \item \textbf{Picture modal:} PET scan
     \item \textbf{Tasks:} Cancer detection
     \item \textbf{Reference Answer:} Lung cancer

\end{itemize}     
\vspace{1em}
\end{flushleft}

\end{minipage}
}
\end{center}
\vfill
\begin{center}
\fbox{%
\begin{minipage}{0.9\textwidth}
\begin{center}
    {\relsize{2}\textbf{CASE 2}}
\end{center}

\begin{center}
    \textbf{Control Group}
\end{center}

\vspace{1em}
\noindent
\begin{tcolorbox}[colback=gray!10,arc=0mm,auto outer arc]
\textbf{Prompt:} Please describe this PET scan and tell me what disease is most likely involved in this picture (only answer the most likely one)
\end{tcolorbox}
\centering
\vspace{1em}
\newcommand{\correct}[1]{\sethlcolor{green!20}\hl{#1}}
\newcommand{\wrong}[1]{\sethlcolor{red!20}\hl{#1}}

\begin{flushleft}
\textbf{GPT-4V Response:}
\begin{tcolorbox}[colback=cyan!10,arc=0mm,auto outer arc]
    \begin{itemize}
    \item \textbf{response 1:}
        The scan shows several areas of increased tracer uptake, which are the bright spots in the image. The areas of high uptake of the tracer suggest a higher level of metabolic activity, which is commonly associated with cancer cells that are more active than normal cells.
        
        \wrong{Given the multiple bright spots of varying sizes throughout the body, this pattern may be indicative of a widespread metastatic disease, where cancer cells from an original tumor have spread to other parts of the body.}

    \item \textbf{response 2:}
        The scan shows several areas of high uptake, which are visible as bright spots. These bright spots indicate areas where a large amount of the tracer has accumulated, which often signifies a high level of metabolic activity that is typical of rapidly dividing cells, such as cancer cells. \wrong{Given the multiple bright spots of varying sizes, which can be seen throughout the body, this pattern could suggest metastatic cancer, where the cancer has spread to multiple locations.}

    \item \textbf{response 3:}
        The bright areas in the head, chest, and abdomen could indicate areas of high metabolic activity, which is often associated with cancer, as cancer cells have a higher metabolic rate than normal cells. However, it's essential to consider the clinical context as these findings can also be seen in infection, inflammation, or benign tumors. Given the multiple areas of uptake, if this is indeed a pathological finding, it would suggest metastatic disease.
        
        \wrong{One of the most common diseases that could cause widespread increased uptake on a PET scan is metastatic cancer}, where the primary cancer has spread to multiple locations throughout the body.
        
    \end{itemize}
\vspace{1em}
\end{tcolorbox}
\vspace{1em}
\end{flushleft}

\end{minipage}
}
\end{center}
\begin{center}
\fbox{%
\begin{minipage}{0.9\textwidth}
\begin{center}
    {\relsize{2}\textbf{CASE 2}}
\end{center}

\begin{center}
    \textbf{Control Group}
\end{center}

\vspace{1em}
\noindent
\begin{tcolorbox}[colback=gray!10,arc=0mm,auto outer arc]
\textbf{Prompt:} Focusing on the chest, please describe this PET scan and tell me what disease is most likely involved in this picture (only answer the most likely one)
\end{tcolorbox}
\centering
\vspace{1em}
\newcommand{\correct}[1]{\sethlcolor{green!20}\hl{#1}}
\newcommand{\wrong}[1]{\sethlcolor{red!20}\hl{#1}}

\begin{flushleft}
\textbf{GPT-4V Response:}
\begin{tcolorbox}[colback=cyan!10,arc=0mm,auto outer arc]
    \begin{itemize}
    \item \textbf{response 1:}
        Focusing on the chest area, there are at least two distinct areas of increased tracer uptake. One is located in the region that would correspond to the mediastinum, which is the central part of the thorax. The other is in the region that might correspond to the lung or pleura.
        
        \correct{Given the presence of these bright spots in the chest, the most likely disease involved is lung cancer, possibly with mediastinal involvement which could suggest lymph node metastasis.} However, other possibilities like lymphoma or infections could also present with increased uptake in these areas, but these are less likely based on the singular bright spots seen in the image. It is important to note that a definitive diagnosis would require correlation with the patient's clinical history, physical examination, and other diagnostic studies. A biopsy would be necessary to confirm the type of cancer or disease.

    \item \textbf{response 2:}
        In the chest area of this scan, there are multiple areas that show increased tracer uptake (highlighted as bright spots). The presence of multiple bright spots can be associated with several conditions, but given that they are scattered throughout the chest, \correct{the most likely disease involved here is metastatic cancer}. Metastases in the chest can arise from primary cancers located in the chest (like lung cancer) or from cancers that originated in other parts of the body and have spread to the chest.

    \item \textbf{response 3:}
        In this particular PET scan image, there are multiple areas of increased radiotracer uptake (hypermetabolic activity) in the region of the chest. The most prominent hypermetabolic activity is noted centrally in the chest area, which could represent the mediastinum, the area between the lungs that contains the heart, great vessels, trachea, esophagus, and lymph nodes.

        Given that there is focal, intense uptake in the chest without clear visualization of the surrounding anatomical structures, \correct{the most likely diagnosis in this scenario would be a mediastinal tumor.} The most common type of mediastinal tumor that would show up on a PET scan is a lymphoma, especially if the hypermetabolic activity is associated with lymph nodes.
        
    \end{itemize}
\vspace{1em}
\end{tcolorbox}
\vspace{1em}
\end{flushleft}

\end{minipage}
}
\end{center}

\end{document}